\documentclass[10pt]{article} 
\usepackage[accepted]{tmlr}

\usepackage{amsmath,amsfonts,bm}

\def\eqref#1{equation~\ref{#1}}

\def\1{\bm{1}}

\DeclareMathAlphabet{\mathsfit}{\encodingdefault}{\sfdefault}{m}{sl}
\SetMathAlphabet{\mathsfit}{bold}{\encodingdefault}{\sfdefault}{bx}{n}

\usepackage{amsmath}
\usepackage{amsfonts}
\usepackage{graphicx}
\usepackage{booktabs}
\usepackage{multirow}
\usepackage{algorithm}
\usepackage{algpseudocode}
\usepackage{xcolor}
\usepackage{hyperref}
\hypersetup{pdftitle={AlphaWiSE: Adaptive Weight Interpolation for Continual Multimodal Representation Learning}, pdfauthor={Sarthak Jain, Qiran Hu, Zhen Zhu, Yaoyao Liu}}
\usepackage{url}
\usepackage{amssymb,amsthm,float}
\newtheorem{proposition}{Proposition}
\newtheorem{definition}{Definition}
\newtheorem*{remark}{Remark}

\title{AlphaWiSE: Adaptive Weight Interpolation \\for Continual Multimodal Representation Learning}

\author{%
  \name Sarthak Jain$^{1,\ast}$ \email sj84@illinois.edu \\
  \name Qiran Hu$^{1,\ast}$ \email qiranhu2@illinois.edu \\
  \name Zhen Zhu$^{1,2,\dagger}$ \email zhenzhucv@google.com \\
  \name Yaoyao Liu$^{1}$ \email lyy@illinois.edu \\[1em]
  \addr $^{1}$University of Illinois Urbana-Champaign \\
  \addr $^{2}$Google DeepMind \\[0.5em]
}

\def\month{11}
\def\year{2026}
\def\openreview{\url{https://openreview.net/forum?id=S7Wo2EGOHt}}

\begin{document}

\maketitle
\begingroup\renewcommand{\thefootnote}{$\ast$}\footnotetext{Equal contribution.}\renewcommand{\thefootnote}{$\dagger$}\footnotetext{Work performed during PhD at UIUC.}\endgroup

\begin{abstract}

Multimodal retrieval models align audio, images, and text in a shared embedding space, but sequential adaptation can degrade cross-modal relationships learned in earlier phases. Existing continual-learning procedures produce a single final checkpoint, so every retrieval direction shares the same stability-plasticity trade-off. We introduce AlphaWiSE, a post-hoc weight-space interpolation method that combines a sequential fine-tuning endpoint with a companion endpoint obtained with a stability-oriented continual-learning method. AlphaWiSE fits one scalar interpolation coefficient for each corresponding parameter tensor on an exemplar memory while both endpoints remain frozen. The interpolated tensors are then materialized into a single checkpoint, so the deployed model retains the architecture and inference cost of a single backbone. On continual audio-image-text retrieval, the fused model improves on the better of its two endpoints in 13 of 18 average retrieval measures and on the fixed blend of the same checkpoints in 25 of 36 measures. Its pairings achieve the strongest average audio-to-text and image-to-audio retrieval among all compared methods. Project page: \url{https://alphawise.cvmlgroup.web.illinois.edu/}.

\end{abstract}

\section{Introduction}

\begin{figure}[!htbp]
    \centering
    \includegraphics[width=1\linewidth]{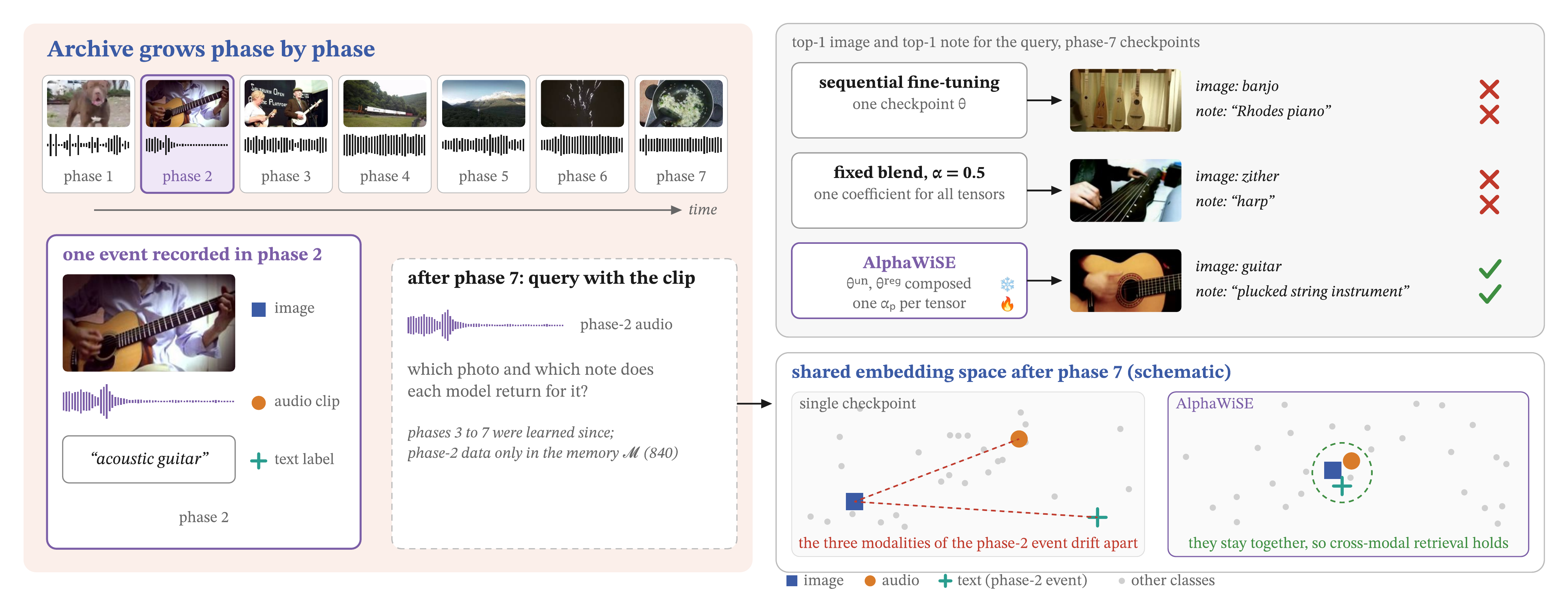}
    \caption{\textbf{Motivating use case.} After phase 7, the audio of a guitar clip recorded in phase 2 serves as the query. Sequential fine-tuning returns ``Rhodes piano'' and a banjo, and the fixed blend returns ``harp'' and a zither. AlphaWiSE returns ``plucked string instrument'' and an acoustic guitar, and this class name is one of the clip's labels. Appendix~\ref{app:figdetails} gives the details.}
    \label{fig:teaser}
\end{figure}

Multimodal representation models have become the foundation of modern retrieval systems by learning a shared embedding space across modalities such as images, text, and audio. Models such as CLIP~\citep{radford2021learningtransferablevisualmodels} and AudioCLIP~\citep{guzhov2021audioclipextendingclipimage} enable cross-modal retrieval and transfer learning by aligning heterogeneous inputs into a common representation. While these models are typically trained on large static datasets, many real-world applications require continual adaptation as new concepts, domains, or data distributions emerge over time. A practical multimodal retrieval system should therefore be able to incorporate new information without destroying the cross-modal relationships learned from previous data. Figure~\ref{fig:teaser} illustrates this motivation through a growing personal media archive, where an earlier recording should continue to retrieve the note and the photograph associated with the same event.

Continual adaptation is particularly challenging for multimodal representation learning because forgetting affects not only individual modality encoders but also the geometry of the shared embedding space~\citep{ni2023continualvisionlanguagerepresentationlearning, goodfellow2015empiricalinvestigationcatastrophicforgetting}. Small changes introduced while learning new tasks can alter the relative positions of representations across multiple modalities, degrading retrieval performance in different and often asymmetric ways. For example, adapting to improve audio-text retrieval may simultaneously weaken image-text or image-audio alignment. As a result, continual multimodal retrieval requires preserving a globally consistent embedding space while remaining sufficiently flexible to learn from new data.

Existing continual-learning methods address catastrophic forgetting through mechanisms such as regularization, knowledge distillation, or replay~\citep{de2019continualsurvey,douillard2020podnet,rebuffi2017icarl}. Although these approaches improve the balance between stability and plasticity, they still produce one final checkpoint. Since these approaches impose different constraints, their final checkpoints can preserve different parts of the multimodal embedding geometry. Selecting a single checkpoint therefore forces all retrieval directions to inherit one stability-plasticity trade-off. This motivates the central question of whether checkpoints obtained from different continual-learning strategies can be composed after training, with each parameter tensor assigned its own balance between retention and adaptation.

To address this question, we propose AlphaWiSE, a post-hoc weight-space interpolation method that composes two frozen endpoints into one fused model using tensor-wise coefficients. The sequential fine-tuning endpoint adapts newly arrived phases, while the companion continual-learning endpoint, trained with Elastic Weight Consolidation (EWC), Learning without Forgetting (LwF), or iCaRL, preserves the cross-modal alignment learned earlier. For each corresponding parameter tensor, AlphaWiSE fits one scalar interpolation coefficient shared by all tensor entries. This tensor-wise parameterization is more expressive than a fixed global interpolation coefficient while keeping coefficient fitting low-dimensional relative to the backbone. AlphaWiSE interpolates all parameter tensors of our AudioCLIP backbone, optimizing one scalar interpolation coefficient per tensor, 499 in total: 152 in the image encoder, 149 in the text encoder, 195 in the audio encoder, and three for the learnable per-pair temperature scalars. The two endpoints, each with approximately 182M parameters, remain frozen throughout coefficient fitting.

We evaluate AlphaWiSE in an AudioCLIP-based continual retrieval setting using the AudioSet dataset~\citep{7952261} under a constrained-memory regime with 840 exemplars. This setting tests whether post-hoc weight-space fusion can improve multimodal continual learning when only an exemplar set is available across sequential phases \citep{rebuffi2017icarl, liu2020mnemonics, chaudhry2019tinyepisodicmemoriescontinual}. Performance is measured on audio-image-text retrieval across continual-learning phases, with retrieval quality reported using recall at 1 (R@1) and mean average precision (mAP) \citep{guzhov2021audioclipextendingclipimage}.

Our contributions are threefold:

\begin{itemize}
\item We formulate continual multimodal checkpoint selection as post-hoc weight-space interpolation between a sequential fine-tuning endpoint and a companion continual-learning endpoint.

\item We introduce tensor-wise interpolation, which fits one scalar coefficient per corresponding parameter tensor on an exemplar memory while the two endpoints remain frozen.

\item We show that the fused model improves on the better of its two endpoints in 13 of 18 average retrieval measures and on the fixed blend of the same checkpoints in 25 of 36, while preserving the architecture, parameter count, and inference cost of a single backbone.

\end{itemize}

\section{Related Work}
\label{sec:related_work}

\textbf{Continual learning.}
Continual learning studies how to adapt models to sequentially arriving data while mitigating catastrophic forgetting, and it is closely related to online learning~\cite{li2019online,li2020online,li2021online,liu2023online}.  
Existing methods can be broadly categorized into regularization-based and replay-based approaches. Regularization-based methods~\citep{Tao2020topology,wang2022foster,simon2021learning,joseph2022energy,yu2020semantic,douillard2020podnet,liu2021adaptive}, including Elastic Weight Consolidation (EWC)~\citep{kirkpatrick2017overcoming} and Learning without Forgetting (LwF)~\citep{li2017learningforgetting}, preserve previously acquired knowledge by constraining parameter updates or distilling predictions from earlier models. Replay-based methods~\citep{rebuffi2017icarl,shin2017continual,liu2020mnemonics,prabhu12356gdumb,luo2023class,wu2018memory,YanHXHTL022,BangKY0C21,choi2021dual,duan2023prompt,fischer2024inemo,liu2024wakening,zhang2023continual,liu2023continual,luo2023class,liu2021rmm}, including iCaRL~\citep{rebuffi2017icarl} and experience replay, retain or revisit samples from previous tasks to stabilize sequential optimization. Despite their different optimization strategies, each method returns a single checkpoint, committing the deployed model to one particular stability-plasticity trade-off.

\textbf{Continual multimodal representation learning.}
Recent advances in multimodal representation learning have led to shared-embedding models such as CLIP and AudioCLIP \citep{guzhov2021audioclipextendingclipimage}, which align heterogeneous modalities within a common embedding space for cross-modal retrieval. Continual adaptation is particularly challenging in these models because forgetting affects not only unimodal representations but also the geometry of cross-modal alignment. Unlike classification, where forgetting primarily changes decision boundaries, retrieval depends on preserving globally consistent embedding neighborhoods across multiple modalities. Consequently, small representation shifts can alter nearest-neighbor rankings and degrade retrieval performance.

Recent methods therefore extend continual learning directly to multimodal or vision-language representation models. C-CLIP combines parameter-efficient adaptation with contrastive knowledge consolidation for continual vision-language learning \citep{liu2025cclip,hu2021loralowrankadaptationlarge}, while Dynamic Adapter Routing (DAR) introduces prototype-guided adapter routing for continual multimodal retrieval \citep{dobrzeniecka2026classificationdynamicadapterrouting,araujo2024learningroutedynamicadapter}. Other approaches similarly employ adapters, prompts, low-rank modules, or task-specific projections to reduce interference during continual adaptation \citep{yu2024boostingcontinuallearningvisionlanguage,poth2023adaptersunifiedlibraryparameterefficient,wang2022dualpromptcomplementarypromptingrehearsalfree,Zhou_2025, zhu2024continuallearningopenvocabularyclassification, zhu2024anytimecontinuallearningopen}. Despite their architectural differences, these methods improve continual learning by introducing additional trainable components or modifying the optimization process.

Our work takes a complementary perspective. Instead of proposing another continual-learning algorithm, we assume that multiple continual-learning solutions have already been obtained and investigate how to compose them into a stronger representation. AlphaWiSE performs post-hoc fusion of frozen continual-learning checkpoints, producing a single deployable model without adapters, routing mechanisms, or additional inference-time computation.

\textbf{Weight-space interpolation and model merging.}
Weight-space interpolation has recently emerged as an effective approach for combining neural networks without increasing inference cost. Stochastic Weight Averaging (SWA) demonstrated that averaging checkpoints along an optimization trajectory improves generalization by converging to wider optima \citep{izmailov2019averagingweightsleadswider,kozal2024continuallearningweightinterpolation,stojanovski2022momentumbasedweightinterpolationstrong}. Building on this idea, model soups, Fisher-weighted averaging, and task arithmetic showed that independently fine-tuned models can often be merged directly in parameter space while maintaining or improving downstream performance \citep{wortsman2022modelsoupsaveragingweights,matena2022mergingmodelsfisherweightedaveraging,ilharco2023editingmodelstaskarithmetic}. WiSE\mbox{-}FT further demonstrated that interpolating a pretrained model with its fine-tuned counterpart improves robustness while preserving zero-shot capability \citep{wortsman2022robustfinetuningzeroshotmodels}.

Existing interpolation methods primarily target transfer learning or robustness by combining a pretrained model with a fine-tuned model using one or a few global interpolation coefficients. In contrast, AlphaWiSE addresses continual multimodal retrieval by learning a tensor-wise interpolation rule between frozen continual-learning checkpoints using an exemplar memory. The coefficient vector follows an ordered set of checkpoint keys, and the scalar associated with each key is applied to every entry of the corresponding parameter tensor. This granularity keeps the post-hoc optimization compact while allowing different layers and modality-specific components to draw different proportions from the endpoints, producing a single fused model with unchanged inference cost.

Merging-based continual learners set their merging weights either per parameter or by one global coefficient. MagMax \citep{marczak2024magmax} and CoFiMA \citep{marouf2024cofima} operate per parameter, with merging weights set by data-free rules or by curvature, and the interpolation step of \citet{kozal2024continuallearningweightinterpolation} applies one fixed global coefficient between the previous and the newly trained model of one run. We discuss these methods in Appendix~\ref{sec:related} and apply merging rules and model soups to the N+EWC checkpoints in Appendix~\ref{app:rules}.

\section{Methods}
\label{sec:AlphaWiSE}

\paragraph{Continual multimodal retrieval.}
We consider a sequence of \(K\) training phases, indexed \(k=0,\ldots,K-1\); in our experiments \(K=8\), with phase \(0\) serving as a shared bootstrap phase and phases \(1\) to \(7\) forming the incremental phases. The data available at phase \(k\) consist of aligned audio-image-text samples.
\[
\mathcal{D}_k = \{(a_i, x_i, t_i)\}_{i=1}^{N_k}.
\]
The model consists of modality-specific audio, image, and text encoders that project each modality into a shared embedding space. The image and text encoders are initialized from CLIP, while the audio encoder is a dedicated ResNeXt-based network trained jointly to project audio into the same embedding space \citep{guzhov2021audioclipextendingclipimage}. The model is evaluated over a set of directed retrieval tasks \(\mathcal{P}_{\mathrm{eval}}\), consisting of audio-to-text, image-to-audio, and image-to-text retrieval \citep{guzhov2021audioclipextendingclipimage}. For each direction \((m,n) \in \mathcal{P}_{\mathrm{eval}}\), an observation from modality \(m\) is used as the query, and candidates from modality \(n\) are ranked according to their embedding similarity \citep{wang2021continuallearningcrossmodalretrieval, yan2025lowrankpromptinteractioncontinual}. This setting differs from static multimodal retrieval in three respects: (i) the training data arrive sequentially across phases rather than being available jointly; (ii) after each phase, the model must incorporate the newly observed data while retaining cross-modal alignment acquired during earlier phases \citep{wang2021continuallearningcrossmodalretrieval,ni2023continualvisionlanguagerepresentationlearning}; and (iii) access to previous-phase data is restricted to an exemplar memory \(\mathcal{M}\), preventing full replay of the training history \citep{rebuffi2017icarl,liu2020mnemonics,chaudhry2019tinyepisodicmemoriescontinual}.

Our method builds on the weight-space interpolation principle used in WiSE\mbox{-}FT~\citep{wortsman2022robustfinetuningzeroshotmodels}. AlphaWiSE is a post-hoc weight-space fusion procedure for two compatible checkpoints from the same architecture. During coefficient fitting, the checkpoint weights are fixed, and only the interpolation coefficients are updated. Our starting observation is that no single continual-learning method is uniformly best in multimodal retrieval: one strategy may better preserve certain modality relationships, while another may provide stronger adaptation to new phases. This suggests that useful information is distributed across distinct continual-learning solutions rather than concentrated in a single final checkpoint. We denote by \(\theta^{\mathrm{un}}\) the less constrained endpoint, such as a standard sequential fine-tuning checkpoint, and by \(\theta^{\mathrm{reg}}\) the more stability-preserving endpoint, such as an EWC, LwF, or iCaRL checkpoint.

\begin{figure}[!t]
    \centering
    \includegraphics[width=1\linewidth]{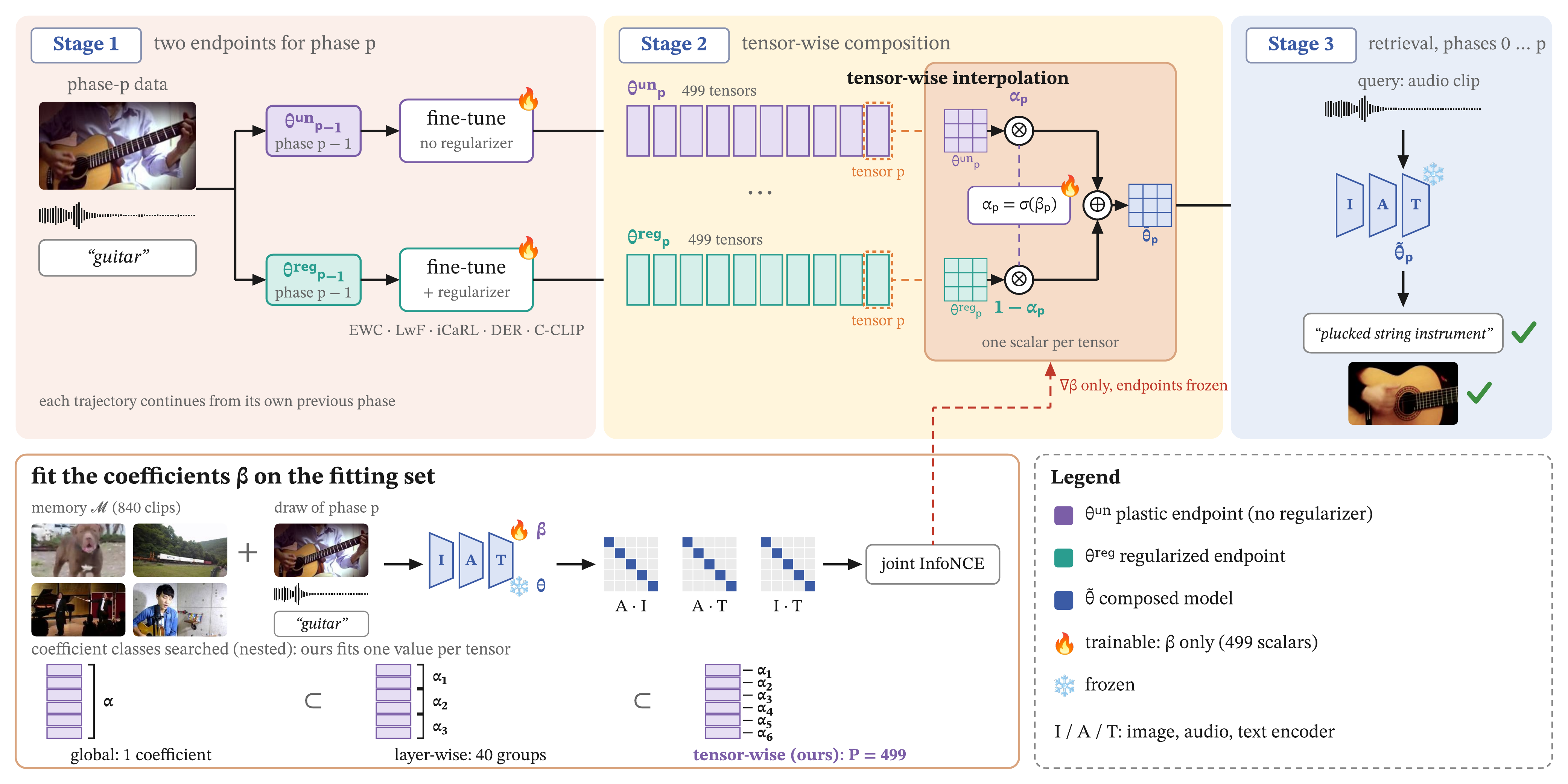}
    \caption{\textbf{AlphaWiSE performs post-hoc, tensor-wise fusion of two compatible continual-learning checkpoints.} In stage 1, each trajectory continues at phase $p$ from its own phase-$(p-1)$ checkpoint. In stage 2, $\tilde\theta=\alpha\odot\theta^{\mathrm{un}}+(1-\alpha)\odot\theta^{\mathrm{reg}}$ is formed tensor by tensor as in Eq.~\ref{eq:alphawise_fusion}. In stage 3, the fused model retrieves over all seen phases at the size and cost of either endpoint. Appendix~\ref{app:figdetails} gives the details.}
    \label{fig:architecture}
\end{figure}

Let \((\kappa_1,\ldots,\kappa_P)\) denote the ordered checkpoint keys selected
for interpolation. For each key \(\kappa_p\), let
\(\theta_p^{\mathrm{un}}\) and \(\theta_p^{\mathrm{reg}}\) denote the
corresponding endpoint tensors. The tensors under the same key must have
identical shapes and represent the same model component; we refer to such tensor pairs as corresponding tensors. AlphaWiSE learns one scalar interpolation coefficient shared by all entries of each corresponding parameter tensor.

\begin{equation}
\tilde{\theta}_p(\boldsymbol{\beta})
=
\alpha_p\theta_p^{\mathrm{un}}
+
(1-\alpha_p)\theta_p^{\mathrm{reg}},
\qquad
\alpha_p=\sigma(\beta_p),
\label{eq:alphawise_fusion}
\end{equation}

where \(\beta_p\in\mathbb{R}\) is unconstrained and the sigmoid keeps \(\alpha_p\in(0,1)\). We initialize \(\beta_p=0\) for all tensors, so every coefficient starts from \(\alpha_p=0.5\). Figure~\ref{fig:architecture} summarizes the resulting fusion procedure.

The coefficients are fitted on the exemplar memory \(\mathcal{M}\), extended by a class-balanced draw of the current phase to form the fitting set of Section~\ref{sec:implementation}. For a minibatch \(\mathcal{B}=\{(a_i,x_i,t_i)\}_{i=1}^{B}\), the fused model produces audio, image, and text embeddings in the shared retrieval space. For a directed modality pair \((m,n)\), let \(s_{ij}^{m,n}(\boldsymbol{\beta})\) denote the temperature-scaled similarity between the \(m\)-embedding of example \(i\) and the \(n\)-embedding of example \(j\) under the fused model \(\tilde{\theta}(\boldsymbol{\beta})\). The directed InfoNCE loss is

\begin{equation}
\mathcal{L}_{m\rightarrow n}(\mathcal{B};\boldsymbol{\beta})
=
-\frac{1}{B}\sum_{i=1}^{B}
\log
\frac{\exp(s_{ii}^{m,n}(\boldsymbol{\beta}))}
{\sum_{j=1}^{B}\exp(s_{ij}^{m,n}(\boldsymbol{\beta}))}.
\label{eq:directed_infonce}
\end{equation}

Given a set \(\mathcal{P}_{\mathrm{fit}}\subseteq\mathcal{P}_{\mathrm{eval}}\) of retrieval directions used for coefficient fitting, AlphaWiSE averages the directed losses of Eq.~\ref{eq:directed_infonce} and minimizes the empirical retrieval loss

\begin{equation}
\min_{\boldsymbol{\beta}\in\mathbb{R}^{P}}
\mathbb{E}_{\mathcal{B}\sim\mathcal{M}}
\left[
\mathcal{L}_{\mathrm{ret}}(\mathcal{B};\boldsymbol{\beta})
\right],
\qquad
\mathcal{L}_{\mathrm{ret}}(\mathcal{B};\boldsymbol{\beta})
=
\frac{1}{|\mathcal{P}_{\mathrm{fit}}|}
\sum_{(m,n)\in\mathcal{P}_{\mathrm{fit}}}
\mathcal{L}_{m\rightarrow n}(\mathcal{B};\boldsymbol{\beta}).
\label{eq:alphawise_objective}
\end{equation}

The set \(\mathcal{P}_{\mathrm{fit}}\) can contain one direction for a pair-specific objective or multiple directions for the joint objective studied in Section~\ref{sec:cross_objective}. Gradients are backpropagated through the fused model to \(\boldsymbol{\beta}\), while \(\theta^{\mathrm{un}}\) and \(\theta^{\mathrm{reg}}\) stay fixed. Algorithm~\ref{alg:alphawiseft} in Appendix~\ref{app:fitting} summarizes the complete coefficient-fitting procedure.

The coefficient gradient provides a local interpretation of the AlphaWiSE update. For tensor \(p\),
\begin{equation}
\frac{\partial\mathcal{L}_{\mathrm{ret}}}{\partial\beta_p}
=
\alpha_p(1-\alpha_p)
\left\langle
\nabla_{\tilde{\theta}_p}\mathcal{L}_{\mathrm{ret}},
\theta_p^{\mathrm{un}}-\theta_p^{\mathrm{reg}}
\right\rangle,
\label{eq:alphawise_beta_gradient}
\end{equation}
where the inner product is over all entries of tensor \(p\). Thus, each coefficient is updated according to the local directional derivative of the exemplar retrieval loss along the difference between the two endpoint tensors.

We fit one coefficient per tensor because the coefficient classes are nested. A coefficient class ties groups of tensors to one value, ranging from one global coefficient through layer-wise groups to one coefficient per tensor. This nesting implies that a finer class never has a higher minimum of the fitting loss. Under a quadratic model with tensor-diagonal curvature, Proposition~\ref{prop:gap} of Appendix~\ref{app:argument} gives the cost of tying as the curvature-weighted spread of the tensor-wise optima inside each tied group, and Section~\ref{sec:granularity} measures that cost in retrieval.

When a fit started at \(\alpha_p=0.5\) stays above the lower-loss endpoint in fitting loss, we restart it from that endpoint, which the tensor-wise class contains. This second fit, started at $\alpha_p=0.99$ or $0.01$, keeps the endpoint's retrieval performance and matches its fitting loss to within $0.02$ nats or lies below it. This is rule~(iv) of the four rules of use in Appendix~\ref{app:guidance}.

\section{Experiments}

We evaluate AlphaWiSE in the AudioCLIP-based continual retrieval setting using the AudioSet dataset~\citep{7952261} under a constrained-memory regime with 840 exemplars. This setting is designed to test whether post-hoc weight-space fusion can improve multimodal continual learning when only an exemplar set is available across sequential phases \citep{rebuffi2017icarl, liu2020mnemonics, chaudhry2019tinyepisodicmemoriescontinual}. Performance is measured on audio-image-text retrieval across the continual-learning phases, with retrieval quality reported using R@1 and mAP \citep{guzhov2021audioclipextendingclipimage}.

Our evaluation focuses on whether AlphaWiSE can improve retention and transfer relative to individual continual-learning trajectories, including standard fine-tuning, EWC, iCaRL, and LwF \citep{kirkpatrick2017overcoming, li2017learningforgetting, rebuffi2017icarl}. In this setting, each baseline represents a different stability-plasticity trade-off: standard fine-tuning provides stronger adaptation to new data, while regularization- and distillation-based methods are designed to preserve prior behavior. AlphaWiSE uses the 840-exemplar memory to learn tensor-wise interpolation coefficients between frozen checkpoints produced by these strategies, enabling the fused model to reuse complementary properties of different training trajectories without increasing inference-time model capacity.

This evaluation is intended to answer whether AlphaWiSE improves continual multimodal retrieval in the constrained-memory regime, and whether learned weight-space interpolation balances the preservation of prior cross-modal alignment against adaptation to later phases better than selecting any single continual-learning trajectory.

\subsection{Implementation details}
\label{sec:implementation}

\paragraph{Dataset.}

We conduct our experiments on \textbf{AudioSet}, a large-scale dataset of human-annotated audio events collected from YouTube videos. AudioSet contains 10-second clips annotated with 527 sound-event labels drawn from a hierarchical ontology. The classes cover a broad range of acoustic concepts, including human sounds, musical instruments, animals, vehicles, natural environments, and mechanical sounds. Individual clips may contain multiple labels, and the dataset is highly imbalanced, with substantially different numbers of examples across classes \citep{7952261}. To construct the continual-learning benchmark, we partition the classes used in our experiments into eight disjoint phases. The model is trained sequentially on these phases, introducing a new subset of sound-event classes at each phase without revisiting the complete training data from earlier phases. This class-incremental organization allows us to evaluate both adaptation to newly introduced concepts and retention of previously learned audio-image-text alignment.

\paragraph{Backbone architecture.}
Our implementation uses the AudioCLIP backbone with minor changes. The image and text encoders are inherited from CLIP ViT-B/32 \citep{radford2021learningtransferablevisualmodels}, with embedding dimension \(512\), patch size \(32\), transformer width \(768\), and \(12\) transformer layers, while the audio encoder uses an ESResNeXt-FBSP convolutional encoder \citep{guzhov2021esresnextfbsplearningrobusttimefrequency,guzhov2021audioclipextendingclipimage}. The model is trained and evaluated using three pairwise retrieval objectives: audio-image, audio-text, and image-text. Additionally,  because only the ESResNeXt-FBSP audio encoder contains BatchNorm layers, we replace each BatchNorm module in this encoder with GroupNorm, while preserving its learned affine weight and bias parameters; the CLIP image and text encoders already use LayerNorm and are unaffected. This replacement removes running mean and variance buffers, making normalization independent of batch composition and avoiding inconsistencies that can arise when interpolated affine parameters are paired with BatchNorm statistics estimated under different continual-learning checkpoints \citep{ioffe2015batchnormalizationacceleratingdeep, wu2018groupnormalization, ba2016layernormalization, pham2022continualnormalizationrethinkingbatch, cha2023rebalancingbatchnormalizationexemplarbased}.

\paragraph{Endpoints.}
At each phase, AlphaWiSE merges two checkpoints with identical architectures and parameter structures. The less constrained checkpoint, denoted by \(\theta^{\mathrm{un}}\), is obtained through standard sequential fine-tuning and serves as the more plastic endpoint. It is optimized using SGD with learning rate \(5\times10^{-5}\), momentum \(0.9\), weight decay \(5\times10^{-4}\), and Nesterov momentum. The learning rate follows an exponential decay schedule with \(\gamma=0.96\) per epoch. AlphaWiSE uses the epoch-\(30\) checkpoint as the endpoint \(\theta^{\mathrm{un}}\) \citep{guzhov2021audioclipextendingclipimage}.

The stability-preserving checkpoint, denoted by \(\theta^{\mathrm{reg}}\), is obtained using a continual-learning method such as EWC, LwF, or iCaRL. These methods preserve previous knowledge through parameter regularization, knowledge distillation, or exemplar replay, respectively, and are designed to provide a more constrained endpoint than standard sequential fine-tuning. Appendix~\ref{app:endpoints} gives the objectives and hyperparameters of these methods and the configuration of the two further baselines of Table~\ref{tab:continual_results_840}, DER \citep{yan2021dynamically} and C-CLIP \citep{liu2025cclip}. Both \(\theta^{\mathrm{un}}\) and \(\theta^{\mathrm{reg}}\) remain frozen during coefficient optimization, and AlphaWiSE updates only the interpolation variables. We fit the coefficients on the fitting set, which consists of the exemplar memory and a class-balanced draw of the current phase. The draw contains $840/p$ triplets, as many as the memory holds per phase, so the fitting set holds $n$ triplets, from $1{,}680$ at phase~1 to $960$ at phase~7. The fit uses Adam with learning rate $10^{-3}$ and batch size $64$ for $150$ epochs, starting from $\alpha_p=0.5$, with seeds $42$ and $0$ to $3$. Every AlphaWiSE row of Tables~\ref{tab:continual_results_840}, \ref{tab:cross_objective} and~\ref{tab:cl-transfer} is the mean over these five fitting seeds, with its standard deviation. Each WiSE\mbox{-}FT baseline is the fixed blend of the same two checkpoints, with one coefficient \(\alpha=0.5\) for every tensor.

\subsection{Continual-retrieval results}
\label{sec:results}

\begin{figure}[t]
    \centering
    \includegraphics[width=0.85\linewidth]{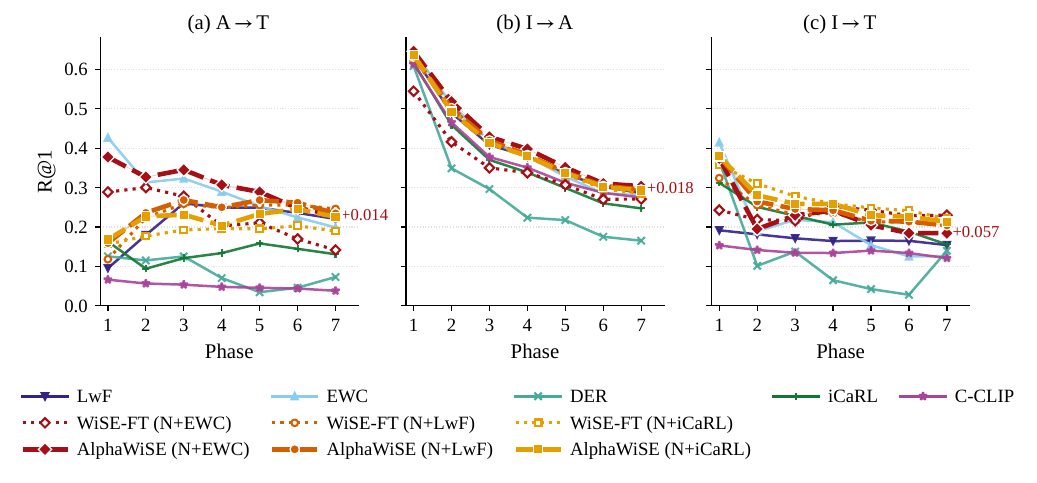}
    \caption{\textbf{Phase-wise R@1 for the listed continual-learning and AlphaWiSE configurations.} Each point is the R@1 of the phase-$p$ model averaged over the pools of phases 0 to $p$. Thin solid lines denote the five continual-learning baselines, dotted lines the three fixed blends, and thick dashed lines the three AlphaWiSE pairings, five-seed means with a one-standard-deviation band narrower than the line width.}
    \label{fig:retention}
\end{figure}

Table~\ref{tab:continual_results_840} reports the main 840-exemplar continual retrieval results across audio-to-text, image-to-audio, and image-to-text retrieval. Overall, AlphaWiSE improves over individual continual-learning baselines on most average retrieval metrics, indicating that post-hoc interpolation can recover a stronger multimodal representation than selecting a single continual-learning trajectory. Figure~\ref{fig:retention} plots the corresponding phase-wise R@1 trajectories over phases 1 to 7. At phase 7, the best AlphaWiSE pairing exceeds the strongest continual-learning baseline by $0.014$ R@1 for A$\rightarrow$T, $0.018$ for I$\rightarrow$A, and $0.057$ for I$\rightarrow$T.

\begin{table}[ht]
\centering
\scriptsize
\setlength{\tabcolsep}{2.5pt}
\caption{Performance comparison on A$\rightarrow$T, I$\rightarrow$A, and I$\rightarrow$T tasks with 840 exemplars. N denotes the normal sequential fine-tuning checkpoint (\(\theta^{\mathrm{un}}\)). Rows marked ``learned'' fit one global coefficient under the protocol of AlphaWiSE. The line labeled ``gain over the better endpoint'' is the AlphaWiSE mean minus the higher of N and the stability endpoint, per column. The best result in each column is shown in bold.}
\label{tab:continual_results_840}
\vspace{0.3cm}

\resizebox{\textwidth}{!}{
\begin{tabular}{lcccccccccccc}
\toprule
& \multicolumn{4}{c}{A$\rightarrow$T}
& \multicolumn{4}{c}{I$\rightarrow$A}
& \multicolumn{4}{c}{I$\rightarrow$T} \\
\cmidrule(lr){2-5}
\cmidrule(lr){6-9}
\cmidrule(lr){10-13}

& \multicolumn{2}{c}{Average}
& \multicolumn{2}{c}{Last Phase}
& \multicolumn{2}{c}{Average}
& \multicolumn{2}{c}{Last Phase}
& \multicolumn{2}{c}{Average}
& \multicolumn{2}{c}{Last Phase} \\
\cmidrule(lr){2-3}
\cmidrule(lr){4-5}
\cmidrule(lr){6-7}
\cmidrule(lr){8-9}
\cmidrule(lr){10-11}
\cmidrule(lr){12-13}

Method
& R@1 & mAP
& R@1 & mAP
& R@1 & mAP
& R@1 & mAP
& R@1 & mAP
& R@1 & mAP \\
\midrule

LwF
& 0.2127 & 0.3304
& 0.2204 & 0.3367
& 0.4046 & 0.3427
& 0.2856 & 0.2298
& 0.1704 & 0.2632
& 0.1540 & 0.2394 \\

EWC
& 0.2900 & 0.3792
& 0.1988 & 0.2901
& 0.4068 & 0.3408
& 0.2810 & 0.2233
& 0.2056 & 0.2784
& 0.1260 & 0.1973 \\

DER
& 0.0842 & 0.1793
& 0.0729 & 0.1625
& 0.2908 & 0.3001
& 0.1652 & 0.1784
& 0.1258 & 0.2012
& 0.1397 & 0.1943 \\

iCaRL
& 0.1350 & 0.2587
& 0.1308 & 0.2333
& 0.3703 & 0.3363
& 0.2473 & 0.2163
& 0.2218 & 0.3214
& 0.1557 & 0.2455 \\

C-CLIP
& 0.0502 & 0.1304
& 0.0380 & 0.1126
& 0.3830 & 0.3390
& 0.2751 & 0.2235
& 0.1368 & 0.2393
& 0.1210 & 0.2138 \\

N
& 0.2377 & 0.3602
& 0.2339 & 0.3473
& 0.4038 & 0.3487
& 0.2955 & 0.2363
& 0.2470 & 0.3435
& 0.1966 & 0.2871 \\

\midrule
Task arithmetic (N+EWC)
& 0.2648 & 0.3543
& 0.1560 & 0.2413
& 0.4003 & 0.3380
& 0.2952 & 0.2286
& 0.2274 & 0.3056
& 0.1768 & 0.2528 \\

TIES-Merging (N+EWC)
& 0.1987 & 0.3029
& 0.0961 & 0.1830
& 0.4015 & 0.3354
& 0.2768 & 0.2197
& 0.2075 & 0.2973
& 0.1842 & 0.2625 \\

Fisher-weighted averaging (N+EWC)
& 0.1715 & 0.2734
& 0.0820 & 0.1655
& 0.3434 & 0.3189
& 0.2183 & 0.1996
& 0.2629 & 0.3654
& 0.2203 & 0.3195 \\

MagMax (N+EWC)
& 0.2368 & 0.3265
& 0.1300 & 0.2143
& 0.3901 & 0.3320
& 0.2639 & 0.2163
& 0.2040 & 0.2777
& 0.1246 & 0.1972 \\

\midrule
WiSE\mbox{-}FT (N+EWC)
& 0.2273 & 0.3119
& 0.1415 & 0.2199
& 0.3566 & 0.3261
& 0.2709 & 0.2199
& 0.2330 & 0.3253
& 0.2298 & 0.3128 \\

WiSE\mbox{-}FT (N+LwF)
& 0.2295 & 0.3541
& \textbf{0.2457} & \textbf{0.3600}
& 0.4096 & 0.3491
& 0.2921 & 0.2359
& 0.2392 & 0.3427
& 0.2030 & 0.3002 \\

WiSE\mbox{-}FT (N+iCaRL)
& 0.1871 & 0.3151
& 0.1895 & 0.3002
& 0.4052 & 0.3464
& 0.2844 & 0.2294
& 0.2737 & 0.3790
& 0.2209 & 0.3223 \\

\midrule
WiSE\mbox{-}FT, learned $\alpha$ (N+EWC)
& 0.2334 & 0.3190
& 0.1471 & 0.2263
& 0.3618 & 0.3275
& 0.2730 & 0.2212
& 0.2338 & 0.3241
& \textbf{0.2303} & 0.3121 \\

WiSE\mbox{-}FT, learned $\alpha$ (N+LwF)
& 0.2298 & 0.3548
& 0.2455 & 0.3598
& 0.4096 & 0.3492
& 0.2914 & 0.2359
& 0.2411 & 0.3441
& 0.2037 & 0.3005 \\

WiSE\mbox{-}FT, learned $\alpha$ (N+iCaRL)
& 0.2015 & 0.3298
& 0.2018 & 0.3133
& 0.4071 & 0.3477
& 0.2894 & 0.2309
& \textbf{0.2756} & \textbf{0.3803}
& 0.2227 & \textbf{0.3245} \\

\midrule
AlphaWiSE (N+EWC)
& \textbf{0.3034} & \textbf{0.3944}
& 0.2311 & 0.3218
& \textbf{0.4224} & 0.3485
& \textbf{0.3035} & 0.2341
& 0.2305 & 0.3092
& 0.1842 & 0.2574 \\

{\scriptsize std}
& {\scriptsize$\pm$0.0003} & {\scriptsize$\pm$0.0001}
& {\scriptsize$\pm$0.0007} & {\scriptsize$\pm$0.0006}
& {\scriptsize$\pm$0.0002} & {\scriptsize$\pm$0.0001}
& {\scriptsize$\pm$0.0007} & {\scriptsize$\pm$0.0002}
& {\scriptsize$\pm$0.0003} & {\scriptsize$\pm$0.0002}
& {\scriptsize$\pm$0.0007} & {\scriptsize$\pm$0.0008} \\

{\scriptsize gain over the better endpoint}
& {\scriptsize$+0.0134$} & {\scriptsize$+0.0152$}
& {\scriptsize$-0.0028$} & {\scriptsize$-0.0255$}
& {\scriptsize$+0.0156$} & {\scriptsize$-0.0002$}
& {\scriptsize$+0.0080$} & {\scriptsize$-0.0022$}
& {\scriptsize$-0.0165$} & {\scriptsize$-0.0343$}
& {\scriptsize$-0.0124$} & {\scriptsize$-0.0297$} \\

\addlinespace[2pt]
AlphaWiSE (N+LwF)
& 0.2397 & 0.3657
& 0.2345 & 0.3523
& 0.4109 & \textbf{0.3498}
& 0.2952 & \textbf{0.2369}
& 0.2521 & 0.3511
& 0.2040 & 0.2967 \\

{\scriptsize std}
& {\scriptsize$\pm$0.0004} & {\scriptsize$\pm$0.0005}
& {\scriptsize$\pm$0.0008} & {\scriptsize$\pm$0.0004}
& {\scriptsize$\pm$0.0004} & {\scriptsize$\pm$0.0000}
& {\scriptsize$\pm$0.0011} & {\scriptsize$\pm$0.0001}
& {\scriptsize$\pm$0.0004} & {\scriptsize$\pm$0.0002}
& {\scriptsize$\pm$0.0006} & {\scriptsize$\pm$0.0003} \\

{\scriptsize gain over the better endpoint}
& {\scriptsize$+0.0020$} & {\scriptsize$+0.0055$}
& {\scriptsize$+0.0006$} & {\scriptsize$+0.0050$}
& {\scriptsize$+0.0063$} & {\scriptsize$+0.0011$}
& {\scriptsize$-0.0003$} & {\scriptsize$+0.0006$}
& {\scriptsize$+0.0051$} & {\scriptsize$+0.0076$}
& {\scriptsize$+0.0074$} & {\scriptsize$+0.0096$} \\

\addlinespace[2pt]
AlphaWiSE (N+iCaRL)
& 0.2191 & 0.3459
& 0.2256 & 0.3397
& 0.4077 & 0.3490
& 0.2927 & 0.2346
& 0.2638 & 0.3654
& 0.2131 & 0.3123 \\

{\scriptsize std}
& {\scriptsize$\pm$0.0004} & {\scriptsize$\pm$0.0002}
& {\scriptsize$\pm$0.0010} & {\scriptsize$\pm$0.0009}
& {\scriptsize$\pm$0.0005} & {\scriptsize$\pm$0.0000}
& {\scriptsize$\pm$0.0008} & {\scriptsize$\pm$0.0002}
& {\scriptsize$\pm$0.0003} & {\scriptsize$\pm$0.0002}
& {\scriptsize$\pm$0.0005} & {\scriptsize$\pm$0.0004} \\

{\scriptsize gain over the better endpoint}
& {\scriptsize$-0.0186$} & {\scriptsize$-0.0143$}
& {\scriptsize$-0.0083$} & {\scriptsize$-0.0076$}
& {\scriptsize$+0.0039$} & {\scriptsize$+0.0003$}
& {\scriptsize$-0.0028$} & {\scriptsize$-0.0017$}
& {\scriptsize$+0.0168$} & {\scriptsize$+0.0219$}
& {\scriptsize$+0.0165$} & {\scriptsize$+0.0252$} \\

\bottomrule
\end{tabular}

}

\end{table}

The strongest gains appear when AlphaWiSE combines the normal sequential checkpoint with a stability-preserving checkpoint. For audio-to-text retrieval, AlphaWiSE (N+EWC) achieves the best average performance overall, improving over EWC from 0.2900 to 0.3034 R@1 and from 0.3792 to 0.3944 mAP, and AlphaWiSE (N+LwF) achieves the strongest average performance among all LwF-based models, improving over LwF from 0.2127 to 0.2397 R@1 and from 0.3304 to 0.3657 mAP. For image-to-audio retrieval, AlphaWiSE (N+EWC) obtains the strongest average and last-phase R@1, while AlphaWiSE (N+LwF) obtains the strongest average mAP. Across the three pairings, the learned tensor-wise coefficients improve on the fixed blend of the same checkpoints in every average A$\rightarrow$T and I$\rightarrow$A cell, and the strongest source pairing depends on both the retrieval direction and the metric.

For image-to-text retrieval, the two WiSE\mbox{-}FT blends of the N+iCaRL pairing obtain the best average performance, and AlphaWiSE (N+iCaRL) has the next-highest average R@1. This indicates that the most effective checkpoint pairing can depend on the retrieval direction. While EWC provides strong audio-related retention, other trajectories can better preserve or recover image-text alignment. This supports the central motivation of AlphaWiSE: different continual-learning methods encode different stability-plasticity trade-offs, and these trade-offs are not uniformly optimal across modality pairs. Rather than treating each baseline checkpoint as a final solution, AlphaWiSE uses them as reusable components for learned weight-space composition.

In Table~\ref{tab:continual_results_840}, AlphaWiSE exceeds the better of its two endpoints in $13$ of the $18$ average cells, over three pairings, three directions, and two metrics. It exceeds the fixed blend of the same two checkpoints in $25$ of the $36$ average and last-phase cells, among them every A$\rightarrow$T and I$\rightarrow$A average cell. It exceeds its own stability endpoint, whether EWC, LwF, or iCaRL, in all $36$ average and last-phase cells, by at least $0.0063$. Over the five fitting seeds, the standard deviation of every AlphaWiSE cell of Tables~\ref{tab:continual_results_840}, \ref{tab:cross_objective} and~\ref{tab:cl-transfer} is at most $0.0014$, with a median of $0.0003$. Table~\ref{tab:continual_results_840} also lists four merging rules on the N and EWC checkpoints, task arithmetic \citep{ilharco2023editingmodelstaskarithmetic}, TIES-Merging \citep{yadav2023ties}, Fisher-weighted averaging \citep{matena2022mergingmodelsfisherweightedaveraging} and MagMax \citep{marczak2024magmax}. None of the eight merging rules of Appendix~\ref{app:rules} exceeds the EWC endpoint on average A$\rightarrow$T or I$\rightarrow$A R@1, whereas AlphaWiSE exceeds every one of them in all four average A$\rightarrow$T and I$\rightarrow$A cells. We list the cells that produce these counts in Appendices~\ref{app:wording}, \ref{app:r2-seeds}, \ref{app:rules} and~\ref{app:ablation}.

Figure~\ref{fig:tsne_it} in Appendix~\ref{app:moved} provides qualitative t-SNE visualizations of the learned embedding space.

In the image-to-text visualization, AlphaWiSE produces tighter or more separated neighborhoods for several selected classes than the corresponding baseline, suggesting improved organization of the shared representation space for those examples. These qualitative results support the interpretation that AlphaWiSE improves retrieval by shifting the fused model toward a region of weight space that better preserves class-level cross-modal alignment.

\subsection{Cross-objective effects across retrieval directions}
\label{sec:cross_objective}
\begin{table}[t]
\centering
\small
\setlength{\tabcolsep}{5pt}
\caption{Cross-objective analysis for AlphaWiSE (N+EWC). We optimize interpolation coefficients using either one modality-pair objective or the joint objective and evaluate the resulting fused model on all retrieval directions. The best result in each column is shown in bold.}
\label{tab:cross_objective}

\vspace{0.3cm}
\begin{tabular}{lcccccc}
\toprule
\multirow{2}{*}{Coefficient objective}
& \multicolumn{2}{c}{A$\rightarrow$T}
& \multicolumn{2}{c}{I$\rightarrow$A}
& \multicolumn{2}{c}{I$\rightarrow$T} \\
\cmidrule(lr){2-3}
\cmidrule(lr){4-5}
\cmidrule(lr){6-7}

& R@1 & mAP
& R@1 & mAP
& R@1 & mAP \\
\midrule

AlphaWiSE, optimize on A$\rightarrow$T
& 0.3019 & 0.3924
& 0.4171 & 0.3468
& $\mathbf{0.2383}$ & $\mathbf{0.3220}$ \\

{\scriptsize std}
& {\scriptsize$\pm$0.0005} & {\scriptsize$\pm$0.0002}
& {\scriptsize$\pm$0.0002} & {\scriptsize$\pm$0.0001}
& {\scriptsize$\pm$0.0005} & {\scriptsize$\pm$0.0003} \\

AlphaWiSE, optimize on I$\rightarrow$A
& 0.2873 & 0.3847
& 0.4195 & 0.3476
& 0.2258 & 0.3116 \\

{\scriptsize std}
& {\scriptsize$\pm$0.0003} & {\scriptsize$\pm$0.0002}
& {\scriptsize$\pm$0.0002} & {\scriptsize$\pm$0.0001}
& {\scriptsize$\pm$0.0001} & {\scriptsize$\pm$0.0001} \\

AlphaWiSE, optimize on I$\rightarrow$T
& 0.2217 & 0.3044
& 0.3587 & 0.3262
& 0.2337 & 0.3135 \\

{\scriptsize std}
& {\scriptsize$\pm$0.0003} & {\scriptsize$\pm$0.0002}
& {\scriptsize$\pm$0.0001} & {\scriptsize$\pm$0.0000}
& {\scriptsize$\pm$0.0002} & {\scriptsize$\pm$0.0002} \\

AlphaWiSE, optimize jointly
& $\mathbf{0.3034}$ & $\mathbf{0.3944}$
& $\mathbf{0.4224}$ & $\mathbf{0.3485}$
& 0.2305 & 0.3092 \\

{\scriptsize std}
& {\scriptsize$\pm$0.0003} & {\scriptsize$\pm$0.0001}
& {\scriptsize$\pm$0.0002} & {\scriptsize$\pm$0.0001}
& {\scriptsize$\pm$0.0003} & {\scriptsize$\pm$0.0002} \\

\bottomrule
\end{tabular}
\end{table}

Table~\ref{tab:cross_objective} studies whether the interpolation coefficients learned from one modality-pair objective affect other retrieval directions. In this ablation, the coefficients are fitted on one retrieval objective at a time, A$\rightarrow$T, I$\rightarrow$A, or I$\rightarrow$T, and the resulting fused model is evaluated on all retrieval directions. 

Joint optimization gives the best A$\rightarrow$T and I$\rightarrow$A results. In contrast, optimizing only A$\rightarrow$T gives the best I$\rightarrow$T result, reaching 0.2383 R@1 and 0.3220 mAP. Thus, the coefficient objective affects both the directly optimized direction and the other directions.

A gradient-descent step on the coefficients for one retrieval objective lowers another objective to first order when the two coefficient gradients have a positive inner product, as Appendix~\ref{app:crossobjective} derives. The same appendix discusses the pattern of Table~\ref{tab:cross_objective} further. The derivation is a local first-order interpretation, which means that Table~\ref{tab:cross_objective} does not directly measure gradient alignment or establish a causal transfer mechanism.

\subsection{Backward and forward transfer analysis}

Table~\ref{tab:cl-transfer} shows that AlphaWiSE primarily improves forward transfer (FWT), while its effect on backward transfer (BWT) depends on the underlying continual-learning method. Across the nine combinations of base method and retrieval direction, AlphaWiSE improves FWT in all nine cases, with the largest gains for EWC on I$\rightarrow$T retrieval ($+0.0400$) and for iCaRL on I$\rightarrow$A retrieval ($+0.0365$). For EWC, AlphaWiSE also improves BWT in all three retrieval directions, by $+0.0718$ on A$\rightarrow$T, $+0.0184$ on I$\rightarrow$A and $+0.0584$ on I$\rightarrow$T. For iCaRL, the improvements are concentrated in FWT, whereas the BWT changes are small (at most $0.03$ in magnitude). For LwF, AlphaWiSE improves FWT across all three directions but reduces BWT, indicating a stronger bias toward adaptation to unseen phases than toward preservation of earlier-phase performance. Overall, these results suggest that AlphaWiSE can improve transfer to unseen phases, while the stability-plasticity trade-off remains influenced by the trajectories being fused.

\begin{table*}[t]
  \centering
  \caption{Backward and forward transfer on all three retrieval directions, with
  840 exemplars. BWT is the change in earlier-phase performance by the end of training, and FWT is the performance on a phase not yet seen, relative to the shared phase-0 bootstrap. Both are on R@1 over phases 1 to 7. Higher is better for both. Bold marks the better value in each
  base/AlphaWiSE pair; $\Delta$ is AlphaWiSE minus its base.}
  \vspace{0.2cm}
  \label{tab:cl-transfer}
  \begin{tabular}{llcccccc}
    \toprule
    & & \multicolumn{2}{c}{A$\rightarrow$T} & \multicolumn{2}{c}{I$\rightarrow$A} & \multicolumn{2}{c}{I$\rightarrow$T} \\
    \cmidrule(lr){3-4} \cmidrule(lr){5-6} \cmidrule(lr){7-8}
    Base method & Variant & BWT $\uparrow$ & FWT $\uparrow$ & BWT $\uparrow$ & FWT $\uparrow$ & BWT $\uparrow$ & FWT $\uparrow$ \\
    \midrule
    \multirow{4}{*}{EWC}
      & Base        & $-0.2358$ & $+0.0318$ & $-0.0678$ & $+0.0039$ & $-0.2052$ & $-0.0419$ \\
      & + AlphaWiSE & $\mathbf{-0.1640}$ & $\mathbf{+0.0353}$ & $\mathbf{-0.0494}$ & $\mathbf{+0.0203}$ & $\mathbf{-0.1468}$ & $\mathbf{-0.0019}$ \\
      & {\scriptsize std} & {\scriptsize$\pm$0.0013} & {\scriptsize$\pm$0.0003} & {\scriptsize$\pm$0.0011} & {\scriptsize$\pm$0.0010} & {\scriptsize$\pm$0.0008} & {\scriptsize$\pm$0.0007} \\
      & $\Delta$    & $+0.0718$ & $+0.0035$ & $+0.0184$ & $+0.0164$ & $+0.0584$ & $+0.0400$ \\
    \midrule
    \multirow{4}{*}{iCaRL}
      & Base        & $\mathbf{+0.0106}$ & $+0.0167$ & $\mathbf{-0.0080}$ & $-0.0333$ & $\mathbf{-0.0520}$ & $-0.0046$ \\
      & + AlphaWiSE & $-0.0177$ & $\mathbf{+0.0248}$ & $-0.0220$ & $\mathbf{+0.0032}$ & $-0.0581$ & $\mathbf{-0.0004}$ \\
      & {\scriptsize std} & {\scriptsize$\pm$0.0010} & {\scriptsize$\pm$0.0003} & {\scriptsize$\pm$0.0013} & {\scriptsize$\pm$0.0009} & {\scriptsize$\pm$0.0011} & {\scriptsize$\pm$0.0004} \\
      & $\Delta$    & $-0.0283$ & $+0.0081$ & $-0.0140$ & $+0.0365$ & $-0.0061$ & $+0.0042$ \\
    \midrule
    \multirow{4}{*}{LwF}
      & Base        & $\mathbf{-0.0085}$ & $+0.0182$ & $\mathbf{-0.0139}$ & $-0.0040$ & $\mathbf{-0.0066}$ & $-0.0205$ \\
      & + AlphaWiSE & $-0.0309$ & $\mathbf{+0.0289}$ & $-0.0256$ & $\mathbf{-0.0006}$ & $-0.0610$ & $\mathbf{-0.0043}$ \\
      & {\scriptsize std} & {\scriptsize$\pm$0.0013} & {\scriptsize$\pm$0.0006} & {\scriptsize$\pm$0.0011} & {\scriptsize$\pm$0.0008} & {\scriptsize$\pm$0.0014} & {\scriptsize$\pm$0.0007} \\
      & $\Delta$    & $-0.0224$ & $+0.0107$ & $-0.0117$ & $+0.0034$ & $-0.0544$ & $+0.0162$ \\
    \bottomrule
  \end{tabular}
\end{table*}

\subsection{Analysis of learned interpolation}

Since AlphaWiSE learns one interpolation coefficient per corresponding parameter tensor, the converged $\alpha$ values reveal where in the network the fused model draws on \(\theta^{\mathrm{un}}\) and where it preserves the stability endpoint \(\theta^{\mathrm{reg}}\). Figure~\ref{fig:alpha_type_depth} in Appendix~\ref{app:moved} disaggregates the learned coefficients along two axes, relative block depth within each encoder and parameter type (attention weights, MLP/convolutional weights, normalization scales and biases, and remaining bias vectors).

In the image and text encoders, parameter type rather than depth separates the coefficients. The attention and MLP weights of an encoder move together toward one side of the fixed blend, while normalization scales and biases stay closer to it. Appendix~\ref{app:moved} gives the full analysis.

\subsection{\texorpdfstring{Granularity of the coefficients}{Granularity of the coefficients}}
\label{sec:granularity}

We refit the N+EWC coefficients at every granularity of Table~\ref{tab:granularity}, holding the fitting objective, data, and optimization schedule fixed and changing only the coefficient granularity. The learned global coefficient improves A$\rightarrow$T R@1 by only $0.006$ over the fixed blend, whereas the tensor-wise reference fit improves it by $0.076$.

\begin{table}[!htb]
\centering
\scriptsize
\setlength{\tabcolsep}{3pt}
\caption{Granularity ablation on the N+EWC pairing with $840$ exemplars in the forward order, R@1 / mAP averaged over phases 1 to 7. The learned rows differ only in the tying pattern, one coefficient per tensor type in the type-wise class and one per encoder block in the layer-wise class. $L$ at phase 7 is the fitting-set loss in nats on the batches of Table~\ref{tab:barrier}. The best R@1 and the best mAP of every column are in bold.}
\label{tab:granularity}
\vspace{0.2cm}
\renewcommand{\arraystretch}{0.95}
\begin{tabular}{llcccc}
\toprule
Model & \#coef. & A$\rightarrow$T & I$\rightarrow$A & I$\rightarrow$T & $L$ (phase 7) \\
\midrule
EWC endpoint, $\alpha\equiv0$, best global grid point & 0 & 0.2900 / 0.3792 & 0.4068 / 0.3408 & 0.2056 / 0.2784 & 0.69 \\
Fixed blend, $\alpha\equiv0.5$, WiSE-FT & 0 & 0.2273 / 0.3119 & 0.3566 / 0.3261 & 0.2330 / \textbf{0.3253} & 3.66 \\
Learned global coefficient & 1 & 0.2334 / 0.3190 & 0.3618 / 0.3275 & \textbf{0.2338} / 0.3241 & 3.49 \\
Learned type-wise & 6 & 0.2312 / 0.3185 & 0.3670 / 0.3296 & 0.2326 / 0.3149 & 3.11 \\
Learned layer-wise & 40 & \textbf{0.3082} / 0.3962 & 0.4202 / 0.3475 & 0.2309 / 0.3090 & 0.89 \\
Learned tensor-wise, AlphaWiSE seed-42 refit & 499 & 0.3038 / 0.3946 & 0.4222 / 0.3485 & 0.2304 / 0.3092 & 0.92 \\
Learned tensor-wise, AlphaWiSE reference fit & 499 & 0.3037 / 0.3946 & \textbf{0.4225} / 0.3485 & 0.2304 / 0.3090 & 0.93 \\
Learned element-wise & $1.8\times10^{8}$ & 0.3008 / \textbf{0.3969} & 0.4145 / \textbf{0.3501} & 0.2278 / 0.3165 & 0.49 \\
\bottomrule
\end{tabular}
\end{table}

The learned rows of Table~\ref{tab:granularity} fall into three groups, and the tensor-wise class is the finest class whose fit keeps the full R@1 gain. First, the fits of the two coarsest coefficient classes stay on the high-loss ridge between the endpoints. The global coefficient remains at the loss level of the segment interior, $3.49$ nats at phase 7 against $3.66$ for the fixed blend. The EWC endpoint lies lower, at $0.69$ nats. The six type-wise coefficients reach $3.11$ nats. Both fits stay close to the fixed blend in retrieval, with $0.2334$ and $0.2312$ A$\rightarrow$T R@1 for the global and the type-wise fit against $0.2273$ for the fixed blend. Second, the layer-wise and tensor-wise fits leave the ridge, with phase-7 losses of $0.89$ and $0.92$ nats. Both exceed the EWC endpoint in every average cell, and they agree with each other to within $0.005$ R@1 and $0.002$ mAP in every cell. Third, the element-wise class uses $1.8\times10^{8}$ coefficients for $960$ to $1{,}680$ triplets and reaches $0.49$ nats on the fitting set at phase~7, $0.2$ nats below the lower-loss endpoint. Its average R@1 nonetheless falls $0.003$ to $0.008$ below that of the tensor-wise fit in every direction, while its mAP is at most $0.008$ higher. Each deficit is at least eight standard deviations of its cell in Table~\ref{tab:continual_results_840}, consistent with an element-wise class that overfits the fitting set rather than learning which parameters to take from which endpoint.

The gain arises once coefficients vary across blocks, which the layer-wise and tensor-wise classes allow and the type-wise class does not. Tying by type keeps the fit on the ridge at $3.11$ nats, although six coefficients can express the mean coefficient of each type in Table~\ref{tab:alphatype}. The tensor-wise class is defined by the parameter tensors alone and applies unchanged to any backbone, whereas the layer-wise partition needs architecture-specific rules for each encoder, as Appendix~\ref{sec:groups} shows.

\subsection{\texorpdfstring{Further benchmarks}{Further benchmarks}}
\label{sec:benchmarks}

We add three standard class-incremental benchmarks under the protocol of Appendix~\ref{app:bench}, with our own runs in the upper block of Table~\ref{tab:morebench} and MagMax's published rows in the lower block. Within the upper block, AlphaWiSE is the best row in all six columns, and in the middle block, with CLIP ViT-L/14 as the image encoder, it is again the best row in all four columns. In the two UrbanSound8K columns, it exceeds the next-best row, the fixed blend, by $3.0$ points in Last and $1.9$ in Avg. The UrbanSound8K test set after the last task holds $1{,}653$ items, so this Last margin is nearly three binomial standard deviations, and the margin over every other row is at least five.

The lower block validates the recipe against MagMax's published numbers. The EWC row of the upper block lies within $0.3$ points of MagMax's published EWC on ImageNet-R and $2.9$ points above it on CIFAR-100. In Last accuracy, AlphaWiSE matches the published MagMax, at $80.51$ against $80.41$ on CIFAR-100 and $83.37$ against $83.33$ on ImageNet-R.

\begin{table}[!htb]
\centering
\footnotesize
\setlength{\tabcolsep}{4pt}
\caption{Class-incremental accuracy in percent on three further benchmarks. Last is the accuracy after the final task over all classes, and Avg is the mean over tasks of the seen-class accuracy. The upper block uses CLIP ViT-B/16 and AudioCLIP, the middle block CLIP ViT-L/14, and the lower block quotes \citet{marczak2024magmax}. The best value of every column in the upper and middle blocks is in bold.}
\label{tab:morebench}
\vspace{0.2cm}
\renewcommand{\arraystretch}{0.88}
\begin{tabular}{lcccccc}
\toprule
& \multicolumn{2}{c}{CIFAR-100} & \multicolumn{2}{c}{ImageNet-R} & \multicolumn{2}{c}{UrbanSound8K} \\
\cmidrule(lr){2-3}\cmidrule(lr){4-5}\cmidrule(lr){6-7}
Method & Last & Avg & Last & Avg & Last & Avg \\
\midrule
EWC & 79.12 & 86.28 & 82.68 & 88.24 & 17.00 & 43.83 \\
Fixed blend, $\alpha\equiv0.5$, WiSE-FT & 80.10 & 86.06 & 83.20 & 88.20 & 20.99 & 48.31 \\
MagMax selection on the N and EWC task vectors & 72.61 & 82.22 & 79.12 & 86.03 & 18.57 & 48.26 \\
AlphaWiSE (N+EWC) & \textbf{80.51} & \textbf{86.54} & \textbf{83.37} & \textbf{88.32} & \textbf{23.96} & \textbf{50.16} \\
\midrule
\multicolumn{7}{l}{\emph{The image benchmarks with CLIP ViT-L/14 as the image encoder}} \\
EWC & 84.55 & 90.60 & 91.32 & 93.79 &  &  \\
Fixed blend, $\alpha\equiv0.5$, WiSE-FT & 85.88 & 91.14 & 91.58 & 94.11 &  &  \\
MagMax selection on the N and EWC task vectors & 79.52 & 87.81 & 88.62 & 92.31 &  &  \\
AlphaWiSE (N+EWC) & \textbf{86.27} & \textbf{91.40} & \textbf{91.62} & \textbf{94.18} &  &  \\
\midrule
\multicolumn{7}{l}{\emph{As reported by MagMax at ECCV 2024, CLIP ViT-B/16, $10$ tasks, exemplar-free, Last only}} \\
Zero-shot CLIP & 66.91 &  & 77.73 &  &  &  \\
LwF & 73.45 &  & 82.97 &  &  &  \\
EWC & 76.24 &  & 82.42 &  &  &  \\
Task-vector averaging & 77.04 &  & 81.87 &  &  &  \\
TIES & 77.23 &  & 82.27 &  &  &  \\
MagMax & 80.41 &  & 83.33 &  &  &  \\
Joint training & 90.94 &  & 87.55 &  &  &  \\
\bottomrule
\end{tabular}
\end{table}

\subsection{\texorpdfstring{Memory budget of the coefficient fit}{Memory budget of the coefficient fit}}
\label{sec:memory}

We vary the exemplar memory of the fit from $84$ to $1{,}680$ exemplars, a factor of $20$ around the default of $840$. With the same number of optimizer updates as the fit on the default $840$-exemplar memory, a fit on $84$ exemplars reaches $0.3026$ A$\rightarrow$T and $0.4205$ I$\rightarrow$A average R@1, compared with $0.3038$ and $0.4222$ at $840$ (Table~\ref{tab:memory}). At a fixed $150$ epochs, a smaller memory receives fewer updates, and A$\rightarrow$T and I$\rightarrow$A retrieval then increases with the memory budget. On held-out training clips from the same classes, the tensor-wise fit has the lowest loss among the $840$-exemplar models in Table~\ref{tab:heldout}, at $3.77$ nats versus $4.31$ for EWC, and the matched-update fits at $84$, $210$, and $420$ exemplars reach $3.80$, $3.79$, and $3.79$.

\subsection{\texorpdfstring{Further analyses}{Further analyses}}
\label{sec:further}

Along the segment between the two endpoints, the fitting loss has a barrier in $18$ of the $21$ scans of Table~\ref{tab:barrier}. Table~\ref{tab:pairs} expands the evaluation to five stability methods, a pairing of two stability endpoints, and both class orders. In each of the six pairings that include N, one of its fits exceeds the stability endpoint in all six average retrieval metrics. The iCaRL and C-CLIP endpoints are trained on $3.7\%$ as many new triplets as N (Appendix~\ref{app:endpoints}), and their fusion with N still raises average I$\rightarrow$T R@1 by $0.017$ and $0.019$ over the better endpoint, so a stability endpoint trained on little data suffices for these gains. The coefficient fit, $1.84$ GPU hours or $14\%$ of the GPU hours of an N run, is the only compute we add to the two endpoint runs. Given the training compute of both endpoint runs and the coefficient fit, which amounts to $64$ epochs per phase, a single EWC run does not exceed its $30$-epoch counterpart and stays below AlphaWiSE in every average cell (Appendix~\ref{app:budget}). We report these analyses in Appendices~\ref{app:compat}, \ref{app:tensor}, \ref{app:robust} and~\ref{app:cost}.

\section{Conclusion}

We presented AlphaWiSE, a post-hoc weight-space interpolation method for continual multimodal representation learning. AlphaWiSE combines a frozen sequential fine-tuning checkpoint with a frozen companion continual-learning checkpoint by learning one scalar coefficient for each corresponding parameter tensor using a constrained exemplar memory. The interpolated tensors are then materialized into a single checkpoint with the same architecture, parameter count, and inference-time cost as the source models.
Experiments on continual audio-image-text retrieval show that AlphaWiSE generally improves over EWC, LwF, and iCaRL and often outperforms the fixed blend of WiSE\mbox{-}FT. The forward-transfer analysis further shows improvements in all nine evaluated combinations of base method and retrieval direction, indicating that the learned interpolation frequently improves performance on unseen phases. The cross-objective ablation demonstrates that the coefficient objective can affect both the directly optimized retrieval direction and the other directions.
In the image and text encoders, the learned coefficients also vary more strongly by parameter type than by network depth. Attention, MLP, or convolutional, normalization, and bias parameters often select different mixtures of the endpoints, supporting tensor-wise interpolation over a single global or layer-wise coefficient. Overall, AlphaWiSE shows that adaptive post-hoc interpolation can improve continual multimodal retrieval while retaining the deployment efficiency of a single backbone.


\textbf{Acknowledgments.} This research is supported by the National Artificial Intelligence Research Resource Pilot Awards NAIRR250199, NAIRR260019, and NAIRR260077, the AMD University Program's AI \& HPC Cluster, NVIDIA Academic Grant Program, and Lambda's Research Grant. Computational resources are also provided by Delta and DeltaAI at the National Center for Supercomputing Applications through ACCESS allocations CIS250012, CIS250816, and CIS251188. 

\clearpage
\bibliographystyle{tmlr} 
\bibliography{tmlr} 

@inproceedings{radford2021learningtransferablevisualmodels,
  title={Learning Transferable Visual Models From Natural Language Supervision},
  author={Radford, Alec and Kim, Jong Wook and Hallacy, Chris and Ramesh, Aditya and Goh, Gabriel and Agarwal, Sandhini and Sastry, Girish and Askell, Amanda and Mishkin, Pamela and Clark, Jack and Krueger, Gretchen and Sutskever, Ilya},
  booktitle={International Conference on Machine Learning (ICML)},
  volume={139},
  series={Proceedings of Machine Learning Research},
  pages={8748--8763},
  year={2021},
  publisher={PMLR}
}

@inproceedings{ni2023continualvisionlanguagerepresentationlearning,
  title={Continual Vision-Language Representation Learning with Off-Diagonal Information},
  author={Ni, Zixuan and Wei, Longhui and Tang, Siliang and Zhuang, Yueting and Tian, Qi},
  booktitle={International Conference on Machine Learning (ICML)},
  volume={202},
  series={Proceedings of Machine Learning Research},
  pages={26129--26149},
  year={2023},
  publisher={PMLR}
}

@inproceedings{goodfellow2015empiricalinvestigationcatastrophicforgetting,
  title={An Empirical Investigation of Catastrophic Forgetting in Gradient-Based Neural Networks},
  author={Goodfellow, Ian J. and Mirza, Mehdi and Xiao, Da and Courville, Aaron and Bengio, Yoshua},
  booktitle={International Conference on Learning Representations (ICLR)},
  year={2014}
}

@inproceedings{wortsman2022robustfinetuningzeroshotmodels,
  title={Robust Fine-Tuning of Zero-Shot Models},
  author={Wortsman, Mitchell and Ilharco, Gabriel and Kim, Jong Wook and Li, Mike and Kornblith, Simon and Roelofs, Rebecca and Gontijo Lopes, Raphael and Hajishirzi, Hannaneh and Farhadi, Ali and Namkoong, Hongseok and Schmidt, Ludwig},
  booktitle={IEEE/CVF Conference on Computer Vision and Pattern Recognition (CVPR)},
  pages={7959--7971},
  year={2022}
}

@article{li2017learningforgetting,
  title={Learning without Forgetting},
  author={Li, Zhizhong and Hoiem, Derek},
  journal={IEEE Transactions on Pattern Analysis and Machine Intelligence (TPAMI)},
  volume={40},
  number={12},
  pages={2935--2947},
  year={2018},
  doi={10.1109/TPAMI.2017.2773081}
}

@inproceedings{duan2023prompt,
  title={Prompt-Based Exemplar Super-Compression and Regeneration for Class-Incremental Learning},
  author={Duan, Ruxiao and Chen, Jieneng and Kortylewski, Adam and Yuille, Alan and Liu, Yaoyao},
  booktitle={BMVC},
  year={2025}
}

@inproceedings{fischer2024inemo,
  title={inemo: Incremental neural mesh models for robust class-incremental learning},
  author={Fischer, Tom and Liu, Yaoyao and Jesslen, Artur and Ahmed, Noor and Kaushik, Prakhar and Wang, Angtian and Yuille, Alan L and Kortylewski, Adam and Ilg, Eddy},
  booktitle={European Conference on Computer Vision (ECCV)},
  year={2024}
}

@inproceedings{liu2024wakening,
  title={Wakening past concepts without past data: Class-incremental learning from online placebos},
  author={Liu, Yaoyao and Li, Yingying and Schiele, Bernt and Sun, Qianru},
  booktitle={WACV},
  year={2024}
}

@inproceedings{zhang2023continual,
  title={Continual learning for abdominal multi-organ and tumor segmentation},
  author={Zhang, Yixiao and Li, Xinyi and Chen, Huimiao and Yuille, Alan L and Liu, Yaoyao and Zhou, Zongwei},
  booktitle={MICCAI},
  year={2023}
}

@inproceedings{luo2023class,
  title={Class-incremental exemplar compression for class-incremental learning},
  author={Luo, Zilin and Liu, Yaoyao and Schiele, Bernt and Sun, Qianru},
  booktitle={IEEE/CVF Conference on Computer Vision and Pattern Recognition (CVPR)},
  year={2023}
}

@inproceedings{li2021online,
  title={Online optimal control with affine constraints},
  author={Li, Yingying and Das, Subhro and Li, Na},
  booktitle={AAAI},
  year={2021}
}

@article{li2019online,
  title={Online optimal control with linear dynamics and predictions: Algorithms and regret analysis},
  author={Li, Yingying and Chen, Xin and Li, Na},
  journal={Advances in Neural Information Processing Systems (NeurIPS)},
  year={2019}
}

@article{li2020online,
  title={Online optimization with predictions and switching costs: Fast algorithms and the fundamental limit},
  author={Li, Yingying and Qu, Guannan and Li, Na},
  journal={IEEE TAC},
  year={2020}
}

@article{liu2021rmm,
  title={Rmm: Reinforced memory management for class-incremental learning},
  author={Liu, Yaoyao and Schiele, Bernt and Sun, Qianru},
  journal={Advances in Neural Information Processing Systems (NeurIPS)},
  year={2021}
}

@inproceedings{liu2021adaptive,
  title={Adaptive aggregation networks for class-incremental learning},
  author={Liu, Yaoyao and Schiele, Bernt and Sun, Qianru},
  booktitle={IEEE/CVF Conference on Computer Vision and Pattern Recognition (CVPR)},
  year={2021}
}

@inproceedings{liu2020mnemonics,
  title={Mnemonics training: Multi-class incremental learning without forgetting},
  author={Liu, Yaoyao and Su, Yuting and Liu, An-An and Schiele, Bernt and Sun, Qianru},
  booktitle={IEEE/CVF Conference on Computer Vision and Pattern Recognition (CVPR)},
  year={2020}
}

@inproceedings{douillard2020podnet,
  title={{PODNet}: Pooled Outputs Distillation for Small-Tasks Incremental Learning},
  author={Douillard, Arthur and Cord, Matthieu and Ollion, Charles and Robert, Thomas and Valle, Eduardo},
  booktitle={European Conference on Computer Vision (ECCV)},
  year={2020}
}

@inproceedings{guzhov2021audioclipextendingclipimage,
  title={{AudioCLIP}: Extending {CLIP} to Image, Text and Audio},
  author={Guzhov, Andrey and Raue, Federico and Hees, J{\"o}rn and Dengel, Andreas},
  booktitle={IEEE International Conference on Acoustics, Speech and Signal Processing (ICASSP)},
  pages={976--980},
  year={2022},
  publisher={IEEE},
  doi={10.1109/ICASSP43922.2022.9747631}
}

@inproceedings{liu2025cclip,
  title     = {{C-CLIP}: Multimodal Continual Learning for Vision-Language Model},
  author    = {Wenzhuo Liu and Fei Zhu and Longhui Wei and Qi Tian},
  booktitle = {International Conference on Learning Representations (ICLR)},
  year      = {2025},
  url       = {https://openreview.net/forum?id=sb7qHFYwBc}
}

@inproceedings{hu2021loralowrankadaptationlarge,
  title={{LoRA}: Low-Rank Adaptation of Large Language Models},
  author={Hu, Edward J. and Shen, Yelong and Wallis, Phillip and Allen-Zhu, Zeyuan and Li, Yuanzhi and Wang, Shean and Wang, Lu and Chen, Weizhu},
  booktitle={International Conference on Learning Representations (ICLR)},
  year={2022}
}

@article{dobrzeniecka2026classificationdynamicadapterrouting,
  title={Beyond Classification: Dynamic Adapter Routing for Continual Multimodal Retrieval},
  author={Alicja Dobrzeniecka and Filip Szatkowski and Sebastian Cygert and Szymon Lukasik and Bartlomiej Twardowski},
  journal={arXiv preprint arXiv:2605.31229},
  year={2026}
}

@inproceedings{araujo2024learningroutedynamicadapter,
  title={Learning to Route for Dynamic Adapter Composition in Continual Learning with Language Models},
  author={Araujo, Vladimir and Moens, Marie-Francine and Tuytelaars, Tinne},
  booktitle={Findings of the Association for Computational Linguistics: EMNLP},
  pages={687--696},
  year={2024},
  publisher={Association for Computational Linguistics},
  doi={10.18653/v1/2024.findings-emnlp.38}
}

@article{Zhou_2025,
  title={Learning Without Forgetting for Vision-Language Models},
  author={Zhou, Da-Wei and Zhang, Yuanhan and Wang, Yan and Ning, Jingyi and Ye, Han-Jia and Zhan, De-Chuan and Liu, Ziwei},
  journal={IEEE Transactions on Pattern Analysis and Machine Intelligence (TPAMI)},
  year={2025}
}

@inproceedings{yu2024boostingcontinuallearningvisionlanguage,
  title={Boosting Continual Learning of Vision-Language Models via Mixture-of-Experts Adapters},
  author={Yu, Jiazuo and Zhuge, Yunzhi and Zhang, Lu and Hu, Ping and Wang, Dong and Lu, Huchuan and He, You},
  booktitle={IEEE/CVF Conference on Computer Vision and Pattern Recognition (CVPR)},
  pages={23219--23230},
  year={2024}
}

@inproceedings{poth2023adaptersunifiedlibraryparameterefficient,
  title={Adapters: A Unified Library for Parameter-Efficient and Modular Transfer Learning},
  author={Poth, Clifton and Sterz, Hannah and Paul, Indraneil and Purkayastha, Sukannya and Engl{\"a}nder, Leon and Imhof, Timo and Vuli{\'c}, Ivan and Ruder, Sebastian and Gurevych, Iryna and Pfeiffer, Jonas},
  booktitle={Conference on Empirical Methods in Natural Language Processing: System Demonstrations (EMNLP)},
  pages={149--160},
  year={2023},
  publisher={Association for Computational Linguistics},
  doi={10.18653/v1/2023.emnlp-demo.13}
}

@inproceedings{wang2022dualpromptcomplementarypromptingrehearsalfree,
  title={{DualPrompt}: Complementary Prompting for Rehearsal-Free Continual Learning},
  author={Wang, Zifeng and Zhang, Zizhao and Ebrahimi, Sayna and Sun, Ruoxi and Zhang, Han and Lee, Chen-Yu and Ren, Xiaoqi and Su, Guolong and Perot, Vincent and Dy, Jennifer and Pfister, Tomas},
  booktitle={European Conference on Computer Vision (ECCV)},
  pages={631--648},
  year={2022},
  publisher={Springer Nature Switzerland},
  doi={10.1007/978-3-031-19809-0_36}
}

@inproceedings{izmailov2019averagingweightsleadswider,
  title={Averaging Weights Leads to Wider Optima and Better Generalization},
  author={Izmailov, Pavel and Podoprikhin, Dmitrii and Garipov, Timur and Vetrov, Dmitry and Wilson, Andrew Gordon},
  booktitle={Conference on Uncertainty in Artificial Intelligence (UAI)},
  year={2018}
}

@inproceedings{wortsman2022modelsoupsaveragingweights,
  title={Model soups: averaging weights of multiple fine-tuned models improves accuracy without increasing inference time},
  author={Wortsman, Mitchell and Ilharco, Gabriel and Gadre, Samir Ya and Roelofs, Rebecca and Gontijo-Lopes, Raphael and Morcos, Ari S. and Namkoong, Hongseok and Farhadi, Ali and Carmon, Yair and Kornblith, Simon and Schmidt, Ludwig},
  booktitle={International Conference on Machine Learning (ICML)},
  volume={162},
  series={Proceedings of Machine Learning Research},
  pages={23965--23998},
  year={2022},
  publisher={PMLR}
}

@inproceedings{matena2022mergingmodelsfisherweightedaveraging,
  title={Merging Models with {Fisher}-Weighted Averaging},
  author={Matena, Michael S and Raffel, Colin},
  booktitle={Advances in Neural Information Processing Systems (NeurIPS)},
  volume={35},
  pages={17703--17716},
  year={2022},
  doi={10.52202/068431-1287}
}

@inproceedings{ilharco2023editingmodelstaskarithmetic,
  title={Editing models with task arithmetic},
  author={Ilharco, Gabriel and Ribeiro, Marco Tulio and Wortsman, Mitchell and Gururangan, Suchin and Schmidt, Ludwig and Hajishirzi, Hannaneh and Farhadi, Ali},
  booktitle={International Conference on Learning Representations (ICLR)},
  year={2023}
}

@inproceedings{wang2021continuallearningcrossmodalretrieval,
  title={Continual Learning in Cross-Modal Retrieval},
  author={Wang, Kai and Herranz, Luis and van de Weijer, Joost},
  booktitle={IEEE/CVF Conference on Computer Vision and Pattern Recognition Workshops (CVPRW)},
  pages={3628--3638},
  year={2021}
}

@inproceedings{kozal2024continuallearningweightinterpolation,
  title={Continual Learning with Weight Interpolation},
  author={Kozal, J{\k{e}}drzej and Wasilewski, Jan and Krawczyk, Bartosz and Wo{\'z}niak, Micha{\l}},
  booktitle={IEEE/CVF Conference on Computer Vision and Pattern Recognition Workshops (CVPRW)},
  pages={4187--4195},
  year={2024}
}

@article{stojanovski2022momentumbasedweightinterpolationstrong,
  title={Momentum-based Weight Interpolation of Strong Zero-Shot Models for Continual Learning},
  author={Zafir Stojanovski and Karsten Roth and Zeynep Akata},
  journal={arXiv preprint arXiv:2211.03186},
  year={2022}
}

@inproceedings{7952261,
  title={{Audio Set}: An ontology and human-labeled dataset for audio events},
  author={Gemmeke, Jort F. and Ellis, Daniel P. W. and Freedman, Dylan and Jansen, Aren and Lawrence, Wade and Moore, R. Channing and Plakal, Manoj and Ritter, Marvin},
  booktitle={IEEE International Conference on Acoustics, Speech and Signal Processing (ICASSP)},
  year={2017}
}

@article{chaudhry2019tinyepisodicmemoriescontinual,
  title={On Tiny Episodic Memories in Continual Learning},
  author={Arslan Chaudhry and Marcus Rohrbach and Mohamed Elhoseiny and Thalaiyasingam Ajanthan and Puneet K. Dokania and Philip H. S. Torr and Marc'Aurelio Ranzato},
  journal={arXiv preprint arXiv:1902.10486},
  year={2019}
}

@inproceedings{girdhar2023imagebindembeddingspacebind,
  title={{ImageBind}: One Embedding Space To Bind Them All},
  author={Girdhar, Rohit and El-Nouby, Alaaeldin and Liu, Zhuang and Singh, Mannat and Alwala, Kalyan Vasudev and Joulin, Armand and Misra, Ishan},
  booktitle={IEEE/CVF Conference on Computer Vision and Pattern Recognition (CVPR)},
  pages={15180--15190},
  year={2023}
}

@inproceedings{wu2022wav2cliplearningrobustaudio,
  title={{Wav2CLIP}: Learning Robust Audio Representations From {CLIP}},
  author={Wu, Ho-Hsiang and Seetharaman, Prem and Kumar, Kundan and Bello, Juan Pablo},
  booktitle={IEEE International Conference on Acoustics, Speech and Signal Processing (ICASSP)},
  pages={4563--4567},
  year={2022},
  publisher={IEEE},
  doi={10.1109/ICASSP43922.2022.9747669}
}

@article{de2019continualsurvey,
  title={A Continual Learning Survey: Defying Forgetting in Classification Tasks},
  author={De Lange, Matthias and Aljundi, Rahaf and Masana, Marc and Parisot, Sarah and Jia, Xu and Leonardis, Ale{\v{s}} and Slabaugh, Gregory and Tuytelaars, Tinne},
  journal={IEEE Transactions on Pattern Analysis and Machine Intelligence (TPAMI)},
  volume={44},
  number={7},
  pages={3366--3385},
  year={2022},
  doi={10.1109/TPAMI.2021.3057446}
}

@article{kirkpatrick2017overcoming,
  title={Overcoming catastrophic forgetting in neural networks},
  author={Kirkpatrick, James and Pascanu, Razvan and Rabinowitz, Neil and Veness, Joel and Desjardins, Guillaume and Rusu, Andrei A. and Milan, Kieran and Quan, John and Ramalho, Tiago and Grabska-Barwinska, Agnieszka and Hassabis, Demis and Clopath, Claudia and Kumaran, Dharshan and Hadsell, Raia},
  journal={Proceedings of the National Academy of Sciences (PNAS)},
  volume={114},
  number={13},
  pages={3521--3526},
  year={2017},
  doi={10.1073/pnas.1611835114}
}

@inproceedings{BangKY0C21,
  title={Rainbow Memory: Continual Learning With a Memory of Diverse Samples},
  author={Jihwan Bang and Heesu Kim and Youngjoon Yoo and Jung{-}Woo Ha and Jonghyun Choi},
  booktitle={IEEE/CVF Conference on Computer Vision and Pattern Recognition (CVPR)},
  year={2021}
}

@inproceedings{YanHXHTL022,
  title={Generative Negative Text Replay for Continual Vision-Language Pretraining},
  author={Shipeng Yan and Lanqing Hong and Hang Xu and Jianhua Han and Tinne Tuytelaars and Zhenguo Li and Xuming He},
  booktitle={European Conference on Computer Vision (ECCV)},
  year={2022}
}

@inproceedings{choi2021dual,
  title={Dual-teacher class-incremental learning with data-free generative replay},
  author={Choi, Yoojin and El-Khamy, Mostafa and Lee, Jungwon},
  booktitle={IEEE/CVF Conference on Computer Vision and Pattern Recognition (CVPR)},
  year={2021}
}

@inproceedings{yu2020semantic,
  title={Semantic drift compensation for class-incremental learning},
  author={Yu, Lu and Twardowski, Bartlomiej and Liu, Xialei and Herranz, Luis and Wang, Kai and Cheng, Yongmei and Jui, Shangling and Weijer, Joost van de},
  booktitle={IEEE/CVF Conference on Computer Vision and Pattern Recognition (CVPR)},
  year={2020}
}

@inproceedings{wu2018memory,
  title={Memory Replay {GANs}: Learning to Generate New Categories without Forgetting},
  author={Wu, Chenshen and Herranz, Luis and Liu, Xialei and Wang, Yaxing and van de Weijer, Joost and Raducanu, Bogdan},
  booktitle={Advances in Neural Information Processing Systems (NeurIPS)},
  volume={31},
  year={2018}
}

@inproceedings{prabhu12356gdumb,
  title={{GDumb}: A Simple Approach that Questions Our Progress in Continual Learning},
  author={Prabhu, Ameya and Torr, Philip HS and Dokania, Puneet K},
  booktitle={European Conference on Computer Vision (ECCV)},
  year={2020}
}

@inproceedings{rebuffi2017icarl,
  title={{iCaRL}: Incremental classifier and representation learning},
  author={Rebuffi, Sylvestre-Alvise and Kolesnikov, Alexander and Sperl, Georg and Lampert, Christoph H},
  booktitle={IEEE/CVF Conference on Computer Vision and Pattern Recognition (CVPR)},
  year={2017}
}

@inproceedings{shin2017continual,
  title={Continual learning with deep generative replay},
  author={Shin, Hanul and Lee, Jung Kwon and Kim, Jaehong and Kim, Jiwon},
  booktitle={Advances in Neural Information Processing Systems (NeurIPS)},
  year={2017}
}

@inproceedings{joseph2022energy,
  title={Energy-based Latent Aligner for Incremental Learning},
  author={Joseph, KJ and Khan, Salman and Khan, Fahad Shahbaz and Anwer, Rao Muhammad and Balasubramanian, Vineeth N},
  booktitle={IEEE/CVF Conference on Computer Vision and Pattern Recognition (CVPR)},
  year={2022}
}

@inproceedings{Tao2020topology,
  title={Topology-Preserving Class-Incremental Learning},
  author={Tao, Xiaoyu and Chang, Xinyuan and Hong, Xiaopeng and Wei, Xing and Gong, Yihong},
  booktitle={European Conference on Computer Vision (ECCV)},
  year={2020}
}

@inproceedings{Liu2023Continual,
  title={Continual Detection Transformer for Incremental Object Detection},
  author={Yaoyao Liu and Bernt Schiele and Andrea Vedaldi and Christian Rupprecht},
  booktitle={IEEE/CVF Conference on Computer Vision and Pattern Recognition (CVPR)},
  year={2023}
}

@inproceedings{Liu2023Online,
  title={Online Hyperparameter Optimization for Class-Incremental Learning},
  author={Yaoyao Liu and Yingying Li and Bernt Schiele and Qianru Sun},
  booktitle={AAAI},
  year={2023}
}

@inproceedings{yan2021dynamically,
  title={{DER}: Dynamically Expandable Representation for Class Incremental Learning},
  author={Yan, Shipeng and Xie, Jiangwei and He, Xuming},
  booktitle={IEEE/CVF Conference on Computer Vision and Pattern Recognition (CVPR)},
  year={2021}
}

@inproceedings{simon2021learning,
  title={On learning the geodesic path for incremental learning},
  author={Simon, Christian and Koniusz, Piotr and Harandi, Mehrtash},
  booktitle={IEEE/CVF Conference on Computer Vision and Pattern Recognition (CVPR)},
  year={2021}
}

@inproceedings{wang2022foster,
  title={{FOSTER}: Feature Boosting and Compression for Class-Incremental Learning},
  author={Wang, Fu-Yun and Zhou, Da-Wei and Ye, Han-Jia and Zhan, De-Chuan},
  booktitle={European Conference on Computer Vision (ECCV)},
  year={2022}
}

@article{zhu2024continuallearningopenvocabularyclassification,
  title={Continual Learning in Open-vocabulary Classification with Complementary Memory Systems},
  author={Zhu, Zhen and Lyu, Weijie and Xiao, Yao and Hoiem, Derek},
  journal={Transactions on Machine Learning Research (TMLR)},
  year={2024},
  url={https://openreview.net/forum?id=6j5M75iK3a}
}

@inproceedings{zhu2024anytimecontinuallearningopen,
  title={Anytime Continual Learning for Open Vocabulary Classification},
  author={Zhu, Zhen and Gong, Yiming and Hoiem, Derek},
  booktitle={European Conference on Computer Vision (ECCV)},
  pages={269--285},
  year={2024},
  publisher={Springer Nature Switzerland},
  doi={10.1007/978-3-031-72658-3_16}
}

@inproceedings{yan2025lowrankpromptinteractioncontinual,
  title={Low-rank Prompt Interaction for Continual Vision-Language Retrieval},
  author={Yan, Weicai and Wang, Ye and Lin, Wang and Guo, Zirun and Zhao, Zhou and Jin, Tao},
  booktitle={ACM International Conference on Multimedia (ACM MM)},
  pages={8257--8266},
  year={2024},
  publisher={ACM},
  doi={10.1145/3664647.3681264}
}

@inproceedings{guzhov2021esresnextfbsplearningrobusttimefrequency,
  title={{ESResNe(X)t-fbsp}: Learning Robust Time-Frequency Transformation of Audio},
  author={Guzhov, Andrey and Raue, Federico and Hees, J{\"o}rn and Dengel, Andreas},
  booktitle={International Joint Conference on Neural Networks (IJCNN)},
  pages={1--8},
  year={2021},
  publisher={IEEE},
  doi={10.1109/IJCNN52387.2021.9533654}
}

@inproceedings{ioffe2015batchnormalizationacceleratingdeep,
  title={Batch Normalization: Accelerating Deep Network Training by Reducing Internal Covariate Shift},
  author={Ioffe, Sergey and Szegedy, Christian},
  booktitle={International Conference on Machine Learning (ICML)},
  volume={37},
  series={Proceedings of Machine Learning Research},
  pages={448--456},
  year={2015},
  publisher={PMLR}
}

@inproceedings{wu2018groupnormalization,
  title={Group Normalization},
  author={Wu, Yuxin and He, Kaiming},
  booktitle={European Conference on Computer Vision (ECCV)},
  pages={3--19},
  year={2018}
}

@article{ba2016layernormalization,
  title={Layer Normalization},
  author={Jimmy Lei Ba and Jamie Ryan Kiros and Geoffrey E. Hinton},
  journal={arXiv preprint arXiv:1607.06450},
  year={2016}
}

@inproceedings{pham2022continualnormalizationrethinkingbatch,
  title={Continual Normalization: Rethinking Batch Normalization for Online Continual Learning},
  author={Pham, Quang and Liu, Chenghao and Hoi, Steven},
  booktitle={International Conference on Learning Representations (ICLR)},
  year={2022}
}

@inproceedings{cha2023rebalancingbatchnormalizationexemplarbased,
  title={Rebalancing Batch Normalization for Exemplar-Based Class-Incremental Learning},
  author={Cha, Sungmin and Cho, Sungjun and Hwang, Dasol and Hong, Sunwon and Lee, Moontae and Moon, Taesup},
  booktitle={IEEE/CVF Conference on Computer Vision and Pattern Recognition (CVPR)},
  pages={20127--20136},
  year={2023}
}

@inproceedings{benzaken2022bitfitsimpleparameterefficientfinetuning,
  title={{BitFit}: Simple Parameter-efficient Fine-tuning for Transformer-based Masked Language-models},
  author={Ben Zaken, Elad and Goldberg, Yoav and Ravfogel, Shauli},
  booktitle={Annual Meeting of the Association for Computational Linguistics (ACL)},
  pages={1--9},
  year={2022},
  publisher={Association for Computational Linguistics},
  doi={10.18653/v1/2022.acl-short.1}
}

@inproceedings{frankle2021trainingbatchnormbatchnormexpressive,
  title={Training {BatchNorm} and Only {BatchNorm}: On the Expressive Power of Random Features in {CNNs}},
  author={Frankle, Jonathan and Schwab, David J. and Morcos, Ari S.},
  booktitle={International Conference on Learning Representations (ICLR)},
  year={2021}
}

@inproceedings{adilova2024layerwiselmc,
  title={Layer-wise Linear Mode Connectivity},
  author={Adilova, Linara and Andriushchenko, Maksym and Kamp, Michael and Fischer, Asja and Jaggi, Martin},
  booktitle={International Conference on Learning Representations (ICLR)},
  year={2024}
}

@inproceedings{ainsworth2023rebasin,
  title     = {Git {R}e-{B}asin: merging models modulo permutation symmetries},
  author    = {Ainsworth, Samuel K. and Hayase, Jonathan and Srinivasa, Siddhartha},
  booktitle = {International Conference on Learning Representations (ICLR)},
  year      = {2023}
}

@inproceedings{dziadzio2025time,
  title={How to Merge Your Multimodal Models Over Time?},
  author={Dziadzio, Sebastian and Udandarao, Vishaal and Roth, Karsten and Prabhu, Ameya and Akata, Zeynep and Albanie, Samuel and Bethge, Matthias},
  booktitle={IEEE/CVF Conference on Computer Vision and Pattern Recognition (CVPR)},
  year={2025}
}

@inproceedings{entezari2022permutation,
  title     = {The role of permutation invariance in linear mode connectivity of neural networks},
  author    = {Entezari, Rahim and Sedghi, Hanie and Saukh, Olga and Neyshabur, Behnam},
  booktitle = {International Conference on Learning Representations (ICLR)},
  year      = {2022}
}

@inproceedings{frankle2020lmc,
  title     = {Linear mode connectivity and the lottery ticket hypothesis},
  author    = {Frankle, Jonathan and Dziugaite, Gintare Karolina and Roy, Daniel M. and Carbin, Michael},
  booktitle = {International Conference on Machine Learning (ICML)},
  year      = {2020}
}

@inproceedings{garg2024ticclip,
  title={{TiC-CLIP}: Continual Training of {CLIP} Models},
  author={Garg, Saurabh and Farajtabar, Mehrdad and Pouransari, Hadi and Vemulapalli, Raviteja and Mehta, Sachin and Tuzel, Oncel and Shankar, Vaishaal and Faghri, Fartash},
  booktitle={International Conference on Learning Representations (ICLR)},
  year={2024}
}

@article{hess2025forgetting,
  title={Forgetting of task-specific knowledge in model merging-based continual learning},
  author={Hess, Timm and van de Ven, Gido M. and Tuytelaars, Tinne},
  journal={arXiv preprint arXiv:2507.23311},
  year={2025}
}

@inproceedings{ilharco2022paint,
  title     = {Patching open-vocabulary models by interpolating weights},
  author    = {Ilharco, Gabriel and Wortsman, Mitchell and Gadre, Samir Yitzhak and Song, Shuran and Hajishirzi, Hannaneh and Kornblith, Simon and Farhadi, Ali and Schmidt, Ludwig},
  booktitle = {Advances in Neural Information Processing Systems (NeurIPS)},
  year      = {2022}
}

@inproceedings{jang2024modelstock,
  title     = {Model {S}tock: all we need is just a few fine-tuned models},
  author    = {Jang, Dong-Hwan and Yun, Sangdoo and Han, Dongyoon},
  booktitle = {European Conference on Computer Vision (ECCV)},
  year      = {2024}
}

@inproceedings{jin2023regmean,
  title     = {Dataless knowledge fusion by merging weights of language models},
  author    = {Jin, Xisen and Ren, Xiang and Preotiuc-Pietro, Daniel and Cheng, Pengxiang},
  booktitle = {International Conference on Learning Representations (ICLR)},
  year      = {2023}
}

@inproceedings{jin2025attention,
  title={Fine-Tuning Attention Modules Only: Enhancing Weight Disentanglement in Task Arithmetic},
  author={Jin, Ruochen and Hou, Bojian and Xiao, Jiancong and Su, Weijie and Shen, Li},
  booktitle={International Conference on Learning Representations (ICLR)},
  year={2025}
}

@inproceedings{juneja2023linearconnectivity,
  title     = {Linear connectivity reveals generalization strategies},
  author    = {Juneja, Jeevesh and Bansal, Rachit and Cho, Kyunghyun and Sedoc, Jo\~ao and Saphra, Naomi},
  booktitle = {International Conference on Learning Representations (ICLR)},
  year      = {2023}
}

@inproceedings{lee2017imm,
  title     = {Overcoming catastrophic forgetting by incremental moment matching},
  author    = {Lee, Sang-Woo and Kim, Jin-Hwa and Jun, Jaehyun and Ha, Jung-Woo and Zhang, Byoung-Tak},
  booktitle = {Advances in Neural Information Processing Systems (NeurIPS)},
  year      = {2017}
}

@inproceedings{lee2023surgical,
  title={Surgical Fine-Tuning Improves Adaptation to Distribution Shifts},
  author={Lee, Yoonho and Chen, Annie S. and Tajwar, Fahim and Kumar, Ananya and Yao, Huaxiu and Liang, Percy and Finn, Chelsea},
  booktitle={International Conference on Learning Representations (ICLR)},
  year={2023}
}

@inproceedings{lee2024stella,
  title={{STELLA}: Continual Audio-Video Pre-training with Spatio-Temporal Localized Alignment},
  author={Lee, Jaewoo and Yoon, Jaehong and Kim, Wonjae and Kim, Yunji and Hwang, Sung Ju},
  booktitle={International Conference on Machine Learning (ICML)},
  year={2024}
}

@inproceedings{li2025became,
  title={{BECAME}: {Bayesian} Continual Learning with Adaptive Model Merging},
  author={Li, Mei and Lu, Yuxiang and Dai, Qinyan and Huang, Suizhi and Ding, Yue and Lu, Hongtao},
  booktitle={International Conference on Machine Learning (ICML)},
  year={2025}
}

@inproceedings{li2026pace,
  title={{PACE}: Pretrained Audio Continual Learning},
  author={Li, Chang and Zhou, Kanglei and Wang, Liyuan},
  booktitle={International Conference on Learning Representations (ICLR)},
  year={2026}
}

@inproceedings{liu2025cmcl,
  title={Continual Multimodal Contrastive Learning},
  author={Liu, Xiaohao and Xia, Xiaobo and Ng, See-Kiong and Chua, Tat-Seng},
  booktitle={Advances in Neural Information Processing Systems (NeurIPS)},
  year={2025}
}

@inproceedings{marczak2024magmax,
  title     = {{MagMax}: leveraging model merging for seamless continual learning},
  author    = {Marczak, Daniel and Twardowski, Bart{\l}omiej and Trzci\'nski, Tomasz and Cygert, Sebastian},
  booktitle = {European Conference on Computer Vision (ECCV)},
  year      = {2024}
}

@inproceedings{marouf2024cofima,
  title     = {Weighted ensemble models are strong continual learners},
  author    = {Marouf, Imad Eddine and Roy, Subhankar and Tartaglione, Enzo and Lathuili\`ere, St\'ephane},
  booktitle = {European Conference on Computer Vision (ECCV)},
  year      = {2024}
}

@inproceedings{mirzadeh2021lmccl,
  title     = {Linear mode connectivity in multitask and continual learning},
  author    = {Mirzadeh, Seyed Iman and Farajtabar, Mehrdad and Gorur, Dilan and Pascanu, Razvan and Ghasemzadeh, Hassan},
  booktitle = {International Conference on Learning Representations (ICLR)},
  year      = {2021}
}

@inproceedings{neyshabur2020transfer,
  title     = {What is being transferred in transfer learning?},
  author    = {Neyshabur, Behnam and Sedghi, Hanie and Zhang, Chiyuan},
  booktitle = {Advances in Neural Information Processing Systems (NeurIPS)},
  year      = {2020}
}

@inproceedings{pian2023avcil,
  title={Audio-Visual Class-Incremental Learning},
  author={Pian, Weiguo and Mo, Shentong and Guo, Yunhui and Tian, Yapeng},
  booktitle={IEEE/CVF International Conference on Computer Vision (ICCV)},
  year={2023}
}

@inproceedings{qiu2025mingle,
  title={{MINGLE}: Mixture of Null-Space Gated Low-Rank Experts for Test-Time Continual Model Merging},
  author={Qiu, Zihuan and Xu, Yi and He, Chiyuan and Meng, Fanman and Xu, Linfeng and Wu, Qingbo and Li, Hongliang},
  booktitle={Advances in Neural Information Processing Systems (NeurIPS)},
  year={2025}
}

@inproceedings{qiu2025perturbmerge,
  title={Train with Perturbation, Infer after Merging: A Two-Stage Framework for Continual Learning},
  author={Qiu, Haomiao and Zhang, Miao and Qiao, Ziyue and Nie, Liqiang},
  booktitle={Advances in Neural Information Processing Systems (NeurIPS)},
  year={2025}
}

@inproceedings{rame2022diwa,
  title     = {Diverse weight averaging for out-of-distribution generalization},
  author    = {Ram\'e, Alexandre and Kirchmeyer, Matthieu and Rahier, Thibaud and Rakotomamonjy, Alain and Gallinari, Patrick and Cord, Matthieu},
  booktitle = {Advances in Neural Information Processing Systems (NeurIPS)},
  year      = {2022}
}

@inproceedings{rame2023ratatouille,
  title     = {Model {R}atatouille: recycling diverse models for out-of-distribution generalization},
  author    = {Ram\'e, Alexandre and Ahuja, Kartik and Zhang, Jianyu and Cord, Matthieu and Bottou, L\'eon and Lopez-Paz, David},
  booktitle = {International Conference on Machine Learning (ICML)},
  year      = {2023}
}

@inproceedings{sun2025tail,
  title={{TAIL}: Text-Audio Incremental Learning},
  author={Sun, Yingfei and Gu, Xu and Ji, Wei and Zhao, Hanbin and Yin, Yifang and Zimmermann, Roger},
  booktitle={ACM International Conference on Multimedia (ACM MM)},
  year={2025}
}

@inproceedings{tang2024diki,
  title={Mind the Interference: Retaining Pre-trained Knowledge in Parameter Efficient Continual Learning of Vision-Language Models},
  author={Tang, Longxiang and Tian, Zhuotao and Li, Kai and He, Chunming and Zhou, Hantao and Zhao, Hengshuang and Li, Xiu and Jia, Jiaya},
  booktitle={European Conference on Computer Vision (ECCV)},
  year={2024}
}

@inproceedings{tang2025opcm,
  title={Merging Models on the Fly Without Retraining: A Sequential Approach to Scalable Continual Model Merging},
  author={Tang, Anke and Yang, Enneng and Shen, Li and Luo, Yong and Hu, Han and Du, Bo and Tao, Dacheng},
  booktitle={Advances in Neural Information Processing Systems (NeurIPS)},
  year={2025}
}

@inproceedings{wang2025lines,
  title     = {{LiNeS}: post-training layer scaling prevents forgetting and enhances model merging},
  author    = {Wang, Ke and Dimitriadis, Nikolaos and Favero, Alessandro and Ortiz-Jimenez, Guillermo and Fleuret, Fran\c{c}ois and Frossard, Pascal},
  booktitle = {International Conference on Learning Representations (ICLR)},
  year      = {2025}
}

@inproceedings{wen2023modeconnectivitycil,
  title={Optimizing Mode Connectivity for Class Incremental Learning},
  author={Wen, Haitao and Cheng, Haoyang and Qiu, Heqian and Wang, Lanxiao and Pan, Lili and Li, Hongliang},
  booktitle={International Conference on Machine Learning (ICML)},
  year={2023}
}

@article{wortsman2023lofi,
  title     = {{lo-fi}: distributed fine-tuning without communication},
  author    = {Wortsman, Mitchell and Gururangan, Suchin and Li, Shen and Farhadi, Ali and Schmidt, Ludwig and Rabbat, Michael and Morcos, Ari S.},
  journal   = {Transactions on Machine Learning Research (TMLR)},
  year      = {2023}
}

@inproceedings{xu2024rail,
  title={Advancing Cross-domain Discriminability in Continual Learning of Vision-Language Models},
  author={Xu, Yicheng and Chen, Yuxin and Nie, Jiahao and Wang, Yusong and Zhuang, Huiping and Okumura, Manabu},
  booktitle={Advances in Neural Information Processing Systems (NeurIPS)},
  year={2024}
}

@inproceedings{xu2025prodistill,
  title={Scalable Model Merging with Progressive Layer-wise Distillation},
  author={Xu, Jing and Li, Jiazheng and Zhang, Jingzhao},
  booktitle={International Conference on Machine Learning (ICML)},
  year={2025}
}

@inproceedings{yadav2023ties,
  title     = {{TIES}-{M}erging: resolving interference when merging models},
  author    = {Yadav, Prateek and Tam, Derek and Choshen, Leshem and Raffel, Colin and Bansal, Mohit},
  booktitle = {Advances in Neural Information Processing Systems (NeurIPS)},
  year      = {2023}
}

@article{yadav2024mergingatscale,
  title={What Matters for Model Merging at Scale?},
  author={Yadav, Prateek and Vu, Tu and Lai, Jonathan and Chronopoulou, Alexandra and Faruqui, Manaal and Bansal, Mohit and Munkhdalai, Tsendsuren},
  journal={Transactions on Machine Learning Research (TMLR)},
  year={2025},
  url={https://openreview.net/forum?id=9sbetmvNpW}
}

@inproceedings{yang2024adamerging,
  title     = {{AdaMerging}: adaptive model merging for multi-task learning},
  author    = {Yang, Enneng and Wang, Zhenyi and Shen, Li and Liu, Shiwei and Guo, Guibing and Wang, Xingwei and Tao, Dacheng},
  booktitle = {International Conference on Learning Representations (ICLR)},
  year      = {2024}
}

@inproceedings{zhang2024atlas,
  title     = {Knowledge composition using task vectors with learned anisotropic scaling},
  author    = {Zhang, Frederic Z. and Albert, Paul and Rodriguez-Opazo, Cristian and van den Hengel, Anton and Abbasnejad, Ehsan},
  booktitle = {Advances in Neural Information Processing Systems (NeurIPS)},
  year      = {2024}
}

@inproceedings{zhou2023llfc,
  title={Going Beyond Linear Mode Connectivity: The Layerwise Linear Feature Connectivity},
  author={Zhou, Zhanpeng and Yang, Yongyi and Yang, Xiaojiang and Yan, Junchi and Hu, Wei},
  booktitle={Advances in Neural Information Processing Systems (NeurIPS)},
  year={2023}
}

@inproceedings{zhou2024crosstask,
  title={On the Emergence of Cross-Task Linearity in the Pretraining-Finetuning Paradigm},
  author={Zhou, Zhanpeng and Chen, Zijun and Chen, Yilan and Zhang, Bo and Yan, Junchi},
  booktitle={International Conference on Machine Learning (ICML)},
  year={2024}
}

\clearpage
\appendix 
\section{Pairing-Specific Interpretations}

Appendices~\ref{app:normal_ewc} and~\ref{app:normal_lwf} instantiate the two endpoints of Section~\ref{sec:AlphaWiSE} for the Normal+EWC and Normal+LwF pairings. Appendix~\ref{app:crossobjective} treats cross-objective transfer, and Eq.~\ref{eq:alphawise_beta_gradient} gives the coefficient gradient in general. In both pairings, the normal sequential fine-tuning checkpoint $\theta^{N}$ takes the role of the less constrained endpoint $\theta^{\mathrm{un}}$, and the companion continual-learning checkpoint, written $\theta^{E}$ for EWC and $\theta^{\mathrm{LwF}}$ for LwF, takes the role of the stability-preserving endpoint $\theta^{\mathrm{reg}}$. Throughout, $\mathcal{L}_{\mathrm{new}}$ denotes the current-phase training objective, written $\mathcal{L}_{\mathrm{CLIP}}$ in Appendix~\ref{app:endpoints}; $J_E$ and $J_{\mathrm{LwF}}$ correspond to $\mathcal{L}_{\mathrm{EWC}}$ and $\mathcal{L}_{\mathrm{LwF}}$, with $\lambda_{\mathrm{ewc}}\Omega_E$ the quadratic penalty of that appendix and $D_{\mathrm{old}}$ its distillation term. The Normal+iCaRL pairing follows the same fusion rule with the iCaRL checkpoint in place of $\theta^{E}$ or $\theta^{\mathrm{LwF}}$.

\subsection{Normal + EWC}
\label{app:normal_ewc}

Let $\theta^{N}$ denote the checkpoint obtained by normal sequential training,
and let $\theta^{E}$ denote the checkpoint obtained by EWC. At a given phase,
normal training approximately optimizes the current-phase objective
\[
J_N(\theta)
=
\mathcal{L}_{\mathrm{new}}(\theta),
\]
whereas EWC approximately optimizes
\[
J_E(\theta)
=
\mathcal{L}_{\mathrm{new}}(\theta)
+
\lambda_{\mathrm{ewc}}
\Omega_E(\theta),
\]
where $\Omega_E$ penalizes movement in parameters to which the Fisher estimate of Appendix~\ref{app:endpoints} assigns high importance. In this pairing, $\theta^{E}$ is the more regularized
checkpoint, while $\theta^{N}$ is the less constrained checkpoint.

AlphaWiSE performs interpolation at the level of individual corresponding parameter tensors, each identified by its checkpoint key. Let $p$ index such a tensor. For
Normal+EWC, AlphaWiSE defines the fused tensor as
\[
\tilde{\theta}_{p}
=
\alpha_{p}\theta_{p}^{N}
+
(1-\alpha_{p})\theta_{p}^{E}.
\]
Equivalently,
\[
\tilde{\theta}_{p}
=
\theta_{p}^{E}
+
\alpha_{p}
\left(
\theta_{p}^{N}
-
\theta_{p}^{E}
\right).
\]
Since \(\alpha_p=\sigma(\beta_p)\in(0,1)\) for every finite \(\beta_p\), the implemented parameterization does not attain either endpoint exactly. Instead,
\[
\lim_{\beta_p\to-\infty}\tilde{\theta}_p=\theta_p^{E},
\qquad
\lim_{\beta_p\to+\infty}\tilde{\theta}_p=\theta_p^{N}.
\]
Algebraically extending \(\alpha_p\) to the closed interval \([0,1]\) would recover the two endpoints at \(\alpha_p=0\) and \(\alpha_p=1\).

Thus,
$\alpha_p$ controls movement from the regularized EWC solution toward the more
plastic normal sequential solution.

For the exemplar retrieval loss $\mathcal{L}_{\mathrm{ret}}$ of Eq.~\ref{eq:alphawise_objective}, the coefficient gradient is
\[
\frac{\partial \mathcal{L}_{\mathrm{ret}}}{\partial \alpha_p}
=
\left\langle
\nabla_{\tilde{\theta}_p}\mathcal{L}_{\mathrm{ret}},
\theta_p^{N}-\theta_p^{E}
\right\rangle_F,
\qquad
\langle A,B\rangle_F
=
\sum_{\mathbf{i}} A_{\mathbf{i}}B_{\mathbf{i}}.
\]
Under gradient descent, $\alpha_p$ increases when the inner product above is negative, meaning that movement from the EWC checkpoint toward the normal sequential checkpoint is a local descent direction. It decreases when the inner product is positive. In this sense, AlphaWiSE learns a tensor-wise effective EWC preference after continual training, where smaller values of $\alpha_p$ retain more of the
EWC-stabilized solution, while larger values move toward the more plastic
normal sequential checkpoint.

A tensor-wise EWC bound requires the quadratic penalty to decompose across the
same tensor partition. Vectorizing each parameter tensor, suppose
\[
\Omega_E(\theta)
=
\sum_{p=1}^{P}\Omega_{E,p}(\theta_p),
\qquad
\Omega_{E,p}(\theta_p)
=
(\theta_p-\theta_{\mathrm{old},p})^{\top}
F_p
(\theta_p-\theta_{\mathrm{old},p}),
\quad F_p\succeq 0.
\]
Then each $\Omega_{E,p}$ is convex. For the tensor-wise interpolation above, writing $\tilde{\theta}(\boldsymbol{\alpha})$ for the fused model as a function of the coefficients $\alpha_p=\sigma(\beta_p)$,
\[
\Omega_E\bigl(\tilde{\theta}(\boldsymbol{\alpha})\bigr)
\leq
\sum_{p=1}^{P}
\left[
\alpha_p\Omega_{E,p}(\theta_p^{N})
+
(1-\alpha_p)\Omega_{E,p}(\theta_p^{E})
\right].
\]
When it holds, the fused model inherits a tensor-wise stability guarantee from its endpoints as its EWC penalty is at most the coefficient-weighted tensor-wise mixture of the endpoint penalties. This inequality applies when the EWC matrix has no cross-tensor blocks with respect to the chosen tensor partition. If cross-tensor couplings are present, the bound does not follow from convexity because the tensor-wise mixture is not a single convex combination of the two complete checkpoints.

\subsection{Normal + LwF}
\label{app:normal_lwf}

Let $\theta^{N}$ denote the normal sequential checkpoint, and let
$\theta^{\mathrm{LwF}}$ denote the checkpoint obtained by Learning without Forgetting
(LwF). Normal training approximately optimizes
\[
J_N(\theta)
=
\mathcal{L}_{\mathrm{new}}(\theta),
\]
whereas LwF approximately optimizes
\[
J_{\mathrm{LwF}}(\theta)
=
\mathcal{L}_{\mathrm{new}}(\theta)
+
\lambda_{\mathrm{LwF}}
D_{\mathrm{old}}(\theta),
\]
where $D_{\mathrm{old}}$ is a distillation loss that encourages the current
model to preserve the predictions or similarities of an earlier model on inputs available during the current phase. In this pairing, $\theta^{\mathrm{LwF}}$ is the more regularized checkpoint, while
$\theta^{N}$ is the less constrained checkpoint.

For Normal+LwF, AlphaWiSE defines the fused parameter tensor as
\[
\tilde{\theta}_{p}
=
\alpha_{p}\theta_{p}^{N}
+
(1-\alpha_{p})\theta_{p}^{\mathrm{LwF}}.
\]
Equivalently,
\[
\tilde{\theta}_{p}
=
\theta_{p}^{\mathrm{LwF}}
+
\alpha_{p}
\left(
\theta_{p}^{N}
-
\theta_{p}^{\mathrm{LwF}}
\right).
\]

As in Appendix~\ref{app:normal_ewc}, $\alpha_p=\sigma(\beta_p)\in(0,1)$ for every finite $\beta_p$, so the implemented parameterization does not attain either endpoint. Instead,

\[
\lim_{\beta_p\to-\infty}\tilde{\theta}_p=\theta_p^{\mathrm{LwF}},
\qquad
\lim_{\beta_p\to+\infty}\tilde{\theta}_p=\theta_p^{N},
\]

and algebraically extending $\alpha_p$ to the closed interval $[0,1]$ recovers the LwF checkpoint at $\alpha_p=0$ and the normal sequential checkpoint at $\alpha_p=1$.

For the exemplar retrieval loss $\mathcal{L}_{\mathrm{ret}}$ of Eq.~\ref{eq:alphawise_objective}, the coefficient gradient is
\[
\frac{\partial \mathcal{L}_{\mathrm{ret}}}{\partial \alpha_{p}}
=
\left\langle
\nabla_{\tilde{\theta}_{p}}\mathcal{L}_{\mathrm{ret}},
\theta_{p}^{N}
-
\theta_{p}^{\mathrm{LwF}}
\right\rangle_F,
\]
using the Frobenius inner product defined in Appendix~\ref{app:normal_ewc}.
Under gradient descent, $\alpha_p$ increases when the inner product above is negative, meaning that movement from the LwF checkpoint toward the normal sequential checkpoint is a local descent direction. It decreases when the inner product is positive.

This gives a direct interpretation of Normal+LwF AlphaWiSE: rather than
choosing between pure plasticity and distillation-based preservation, the method
learns how much of the distillation-biased checkpoint to retain for each corresponding parameter tensor. Smaller values of $\alpha_p$ preserve more of the LwF
solution, while larger values move toward the normal sequential checkpoint.

\section{\texorpdfstring{Figures and discussion of Sections 4.2 and 4.5}{Figures and discussion of Sections 4.2 and 4.5}}
\label{app:moved}

\begin{figure}[H]
    \centering
    \includegraphics[width=1\linewidth,trim=8pt 16pt 8pt 22pt,clip]{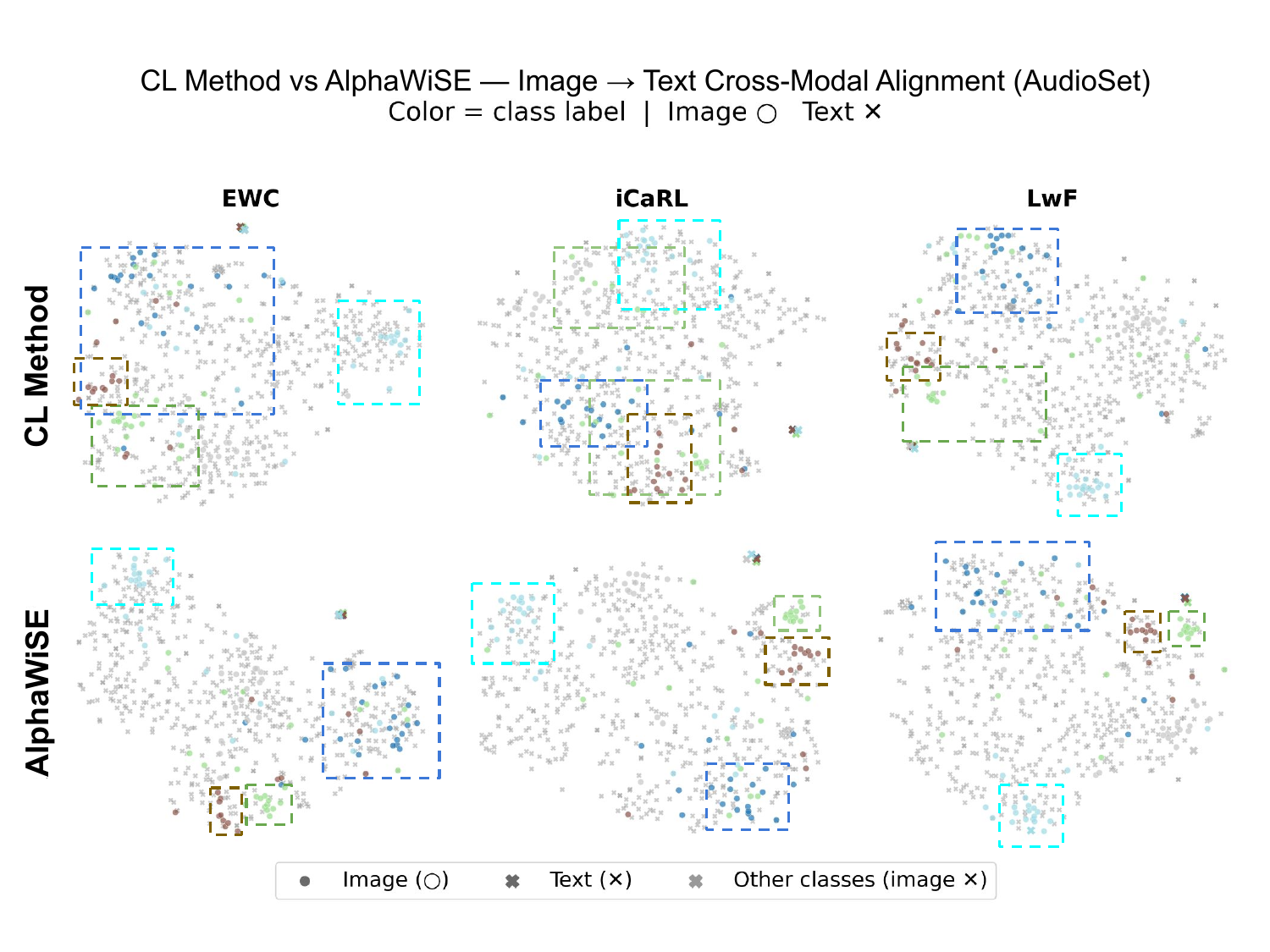}
    \caption{\textbf{Qualitative image-to-text (I$\rightarrow$T) t-SNE projections.} Rows compare a continual-learning checkpoint with its AlphaWiSE fusion, columns EWC, iCaRL, and LwF. Colors denote five selected classes with up to 25 images each, circles image embeddings, and crosses text embeddings. Gray crosses are image embeddings of the remaining classes, one per class. Dashed boxes mark selected neighborhoods.}
    \label{fig:tsne_it}
\end{figure}
\clearpage

\begin{figure}[H]
    \centering
    \includegraphics[width=1\linewidth]{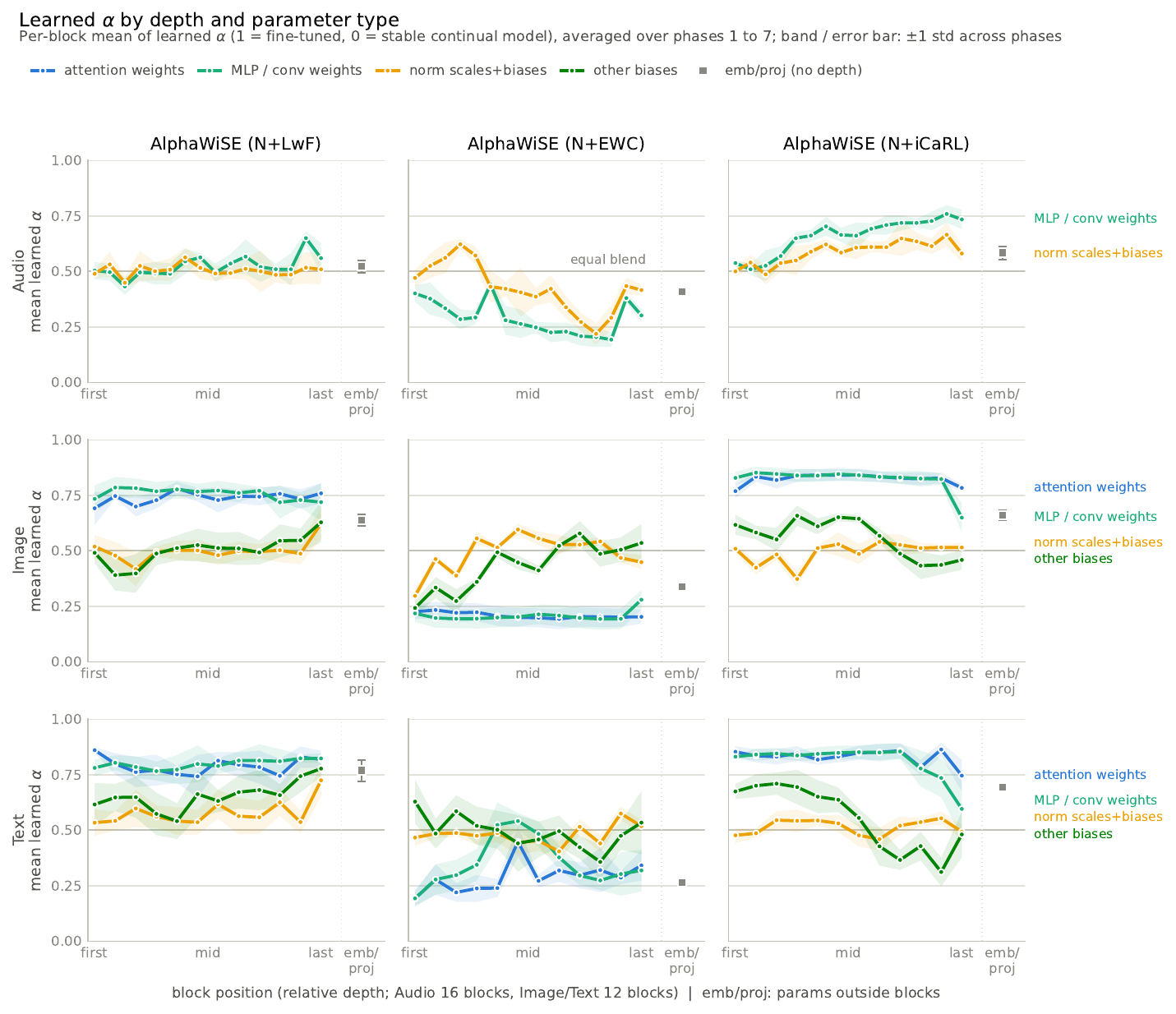}
    \caption{\textbf{Learned interpolation coefficients by depth and parameter type.} Per-block mean of the learned coefficient $\alpha$, the weight on the plastic endpoint N, averaged over phases 1 to 7, with bands and error bars of one standard deviation across phases. Rows correspond to encoder (audio, image,
  text) and columns to the stable continual learner (LwF, EWC, iCaRL).}
    \label{fig:alpha_type_depth}
\end{figure}
\clearpage

Two observations emerge. First, in the image and text encoders, parameter type, not depth, is the dominant axis of variation. The attention and MLP weights of an encoder move together toward one side of the fixed blend while normalization scales and biases stay closer to it, and the separation between the two groups (up to $\Delta\alpha \approx 0.3$ to $0.55$ at individual blocks) is at least as large as the variation of either group across depth, supporting the choice of tensor-wise coefficients rather than layer-wise or global coefficients. Second, normalization and bias parameters absorb a disproportionate share of adaptation whenever the weights remain conservative. For example, in Figure~\ref{fig:alpha_type_depth}, when the stable model is trained with EWC, the image-encoder attention and MLP weights converge to $\alpha \approx 0.2$ in every block and are therefore taken mostly from the stable model, while LayerNorm scales and biases rise from $\alpha \approx 0.3$ in the first blocks to $0.6$ in the middle of the encoder, a range of roughly $0.3$ across depth.

With LwF and iCaRL, the roles are mirrored. The weight matrices are drawn mostly from the fine-tuned model ($\alpha \approx 0.75$ to $0.85$) while the normalization and bias parameters again stay closer to the fixed blend than the weights, and in the audio encoder the convolution weights and normalization parameters move together, toward the stable model for EWC and toward the fine-tuned model for iCaRL. This mirrors prior evidence that norms and biases constitute a cheap, high-leverage channel for domain adaptation \citep{benzaken2022bitfitsimpleparameterefficientfinetuning,frankle2021trainingbatchnormbatchnormexpressive}. Additionally, the results in Table~\ref{tab:continual_results_840} support a view of AlphaWiSE as a mechanism for post-hoc trajectory composition. Standard fine-tuning, EWC, LwF, and replay-based methods each produce checkpoints with different retrieval strengths. EWC is particularly useful for audio-related retrieval, while other pairings are more effective for image-text retrieval. AlphaWiSE exploits this complementarity by learning tensor-wise interpolation coefficients on an exemplar memory.

The cross-objective ablation further suggests that the interpolation coefficients are sensitive to the optimization objective. Joint optimization gives the best A$\rightarrow$T and I$\rightarrow$A results, whereas A$\rightarrow$T-only optimization gives the best I$\rightarrow$T result. This indicates that different modality-pair objectives favor different tensor-wise mixtures, and it motivates future work on retrieval-conditioned or objective-aware interpolation strategies.

Overall, the results indicate that AlphaWiSE is most useful when the endpoints provide complementary solutions. The method does not require a single continual-learning baseline to dominate across all retrieval directions. Instead, it benefits from the fact that different baselines preserve different parts of the multimodal embedding geometry.

\section{\texorpdfstring{Cross-objective transfer in the coefficients}{Cross-objective transfer in the coefficients}}
\label{app:crossobjective}

Since AlphaWiSE updates $\boldsymbol{\beta}$ rather than $\boldsymbol{\alpha}$
directly, define

\[
L_i(\boldsymbol{\beta})
=
\mathcal{L}_i\!\left(
\tilde{\theta}(\boldsymbol{\beta})
\right),
\qquad
\tilde{\theta}_p(\boldsymbol{\beta})
=
\theta_p^{\mathrm{reg}}
+
\sigma(\beta_p)
\left(
\theta_p^{\mathrm{un}}-\theta_p^{\mathrm{reg}}
\right)
\]
for retrieval objective $i$ (here $i$ and $j$ index retrieval objectives, not batch samples). A gradient-descent step on objective $i$ is
\[
\boldsymbol{\beta}'
=
\boldsymbol{\beta}
-
\eta\nabla_{\boldsymbol{\beta}}L_i(\boldsymbol{\beta}).
\]
A first-order Taylor expansion of objective $j$ gives
\[
L_j(\boldsymbol{\beta}')
\approx
L_j(\boldsymbol{\beta})
-
\eta
\left\langle
\nabla_{\boldsymbol{\beta}}L_j(\boldsymbol{\beta}),
\nabla_{\boldsymbol{\beta}}L_i(\boldsymbol{\beta})
\right\rangle.
\]
For a sufficiently small step size, objective $j$ decreases locally
when the two gradients have a positive inner product.

Training on A$\rightarrow$T gives the best I$\rightarrow$T result in this ablation and keeps I$\rightarrow$A within $0.006$ R@1 of the best result. Training on I$\rightarrow$A gives strong I$\rightarrow$A performance and remains competitive on A$\rightarrow$T. Overall, joint optimization gives the strongest audio-related performance, achieving the best A$\rightarrow$T and I$\rightarrow$A results in Table~\ref{tab:cross_objective}.

This pattern suggests that a coefficient objective can benefit retrieval directions beyond its own modality pair in this setting. Optimizing I$\rightarrow$A directly improves image-audio retrieval and remains competitive on A$\rightarrow$T, whereas optimizing A$\rightarrow$T primarily benefits audio-text and image-text retrieval. This behavior is consistent with prior work showing that audio can be aligned with pretrained image-text representation spaces and subsequently support retrieval across all three modalities \citep{guzhov2021audioclipextendingclipimage, wu2022wav2cliplearningrobustaudio}. More directly, ImageBind demonstrates that aligning multiple modalities through images can induce emergent alignment between modality pairs that are not observed together during training \citep{girdhar2023imagebindembeddingspacebind}. Although AlphaWiSE differs substantially from ImageBind in both training regime and scale, our ablation suggests a related phenomenon: the learned interpolation can improve retrieval directions beyond the modality pair directly optimized, as the A$\rightarrow$T objective does for I$\rightarrow$T.

\section{\texorpdfstring{Fitting procedure}{Fitting procedure}}
\label{app:fitting}

In the following appendices, N denotes the sequential fine-tuning checkpoint \(\theta^{\mathrm{un}}\), the plastic endpoint, and \(\theta^{\mathrm{reg}}\) is the stability endpoint. A pairing is named after its two endpoints, as in N+EWC, and the three main pairings are N+EWC, N+LwF, and N+iCaRL, the pairings of Table~\ref{tab:continual_results_840}. The fixed blend is the interpolation with \(\alpha_p=0.5\) for every tensor, as in WiSE\mbox{-}FT, and a global coefficient \(a\) sets \(\alpha_p=a\) for every tensor, so that the blends \(\tilde{\theta}(a\mathbf{1})\) with \(a\in[0,1]\) form the segment between the two endpoints. The fitting set is the data on which the coefficients are fitted and from which the minibatches of Eq.~\ref{eq:alphawise_objective} and Algorithm~\ref{alg:alphawiseft} are drawn, and the fitting loss is the objective of Eq.~\ref{eq:alphawise_objective} on that set, reported as the average of its six directed terms.

\begin{algorithm}[H]
\caption{AlphaWiSE: tensor-wise fusion of two frozen continual-learning checkpoints.}
\label{alg:alphawiseft}
\begin{algorithmic}[1]
\Require Compatible frozen checkpoints \(\theta^{\mathrm{un}}=\{\theta_p^{\mathrm{un}}\}_{p=1}^{P}\) and \(\theta^{\mathrm{reg}}=\{\theta_p^{\mathrm{reg}}\}_{p=1}^{P}\); exemplar memory \(\mathcal{M}\) with the current-phase draw (the fitting set); retrieval-direction set \(\mathcal{P}_{\mathrm{fit}}\); batch size \(B\); optimizer \(\mathrm{Opt}\); fitting steps \(T_{\mathrm{fit}}\).
\Ensure Fused model \(\tilde{\theta}^{\star}\).
\State \(\beta_p\gets0\) for all \(p\in\{1,\ldots,P\}\) \Comment{initializes \(\alpha_p=0.5\)}
\For{\(u=1\) \textbf{to} \(T_{\mathrm{fit}}\)}
    \State Sample \(\mathcal{B}=\{(a_i,x_i,t_i)\}_{i=1}^{B}\) from the fitting set
    \For{\(p=1\) \textbf{to} \(P\)}
        \State \(\alpha_p\gets\sigma(\beta_p)\)
        \State \(\tilde{\theta}_p\gets\alpha_p\theta_p^{\mathrm{un}}+(1-\alpha_p)\theta_p^{\mathrm{reg}}\)
    \EndFor
    \State Compute \(\mathcal{L}_{\mathrm{ret}}(\mathcal{B};\boldsymbol{\beta})\) using Eq.~\ref{eq:alphawise_objective}
    \State \(\boldsymbol{\beta}\gets\mathrm{OptStep}(\mathrm{Opt},\boldsymbol{\beta},\nabla_{\boldsymbol{\beta}}\mathcal{L}_{\mathrm{ret}})\)
\EndFor
\State \(\boldsymbol{\alpha}^{\star}\gets\sigma(\boldsymbol{\beta})\)
\State \(\tilde{\theta}^{\star}\gets\{\alpha_p^{\star}\theta_p^{\mathrm{un}}+(1-\alpha_p^{\star})\theta_p^{\mathrm{reg}}\}_{p=1}^{P}\)
\State \Return \(\tilde{\theta}^{\star}\)
\end{algorithmic}
\end{algorithm}

The coefficients are fitted on the fitting set, which consists of the exemplar memory and a class-balanced draw of the current phase. The draw contains $840/p$ triplets, as many as the memory holds per phase, so the fitting set holds $n$ triplets, from $1{,}680$ at phase~1 to $960$ at phase~7. The fit uses Adam with learning rate $10^{-3}$ and batch size $64$ for $150$ epochs, starting from $\alpha_p=0.5$, with seeds $42$ and $0$ to $3$. The ratio of free coefficients to triplets in the fitting set is about $10^{-3}$ for the global class, $0.02$ to $0.04$ for the layer-wise class, $0.3$ to $0.5$ for the tensor-wise class, and about $10^{5}$ for the element-wise class. Each tensor-wise coefficient is a bounded scalar initialized at $0.5$, so the fit perturbs a fixed blend rather than solving an unconstrained regression. The seed refits, the element-wise row of Table~\ref{tab:granularity}, and the held-out loss of Appendix~\ref{app:heldout} indicate that the fitting set supports the tensor-wise class, at $499/n\approx0.3$ to $0.5$ coefficients per triplet. In every average R@1 and mAP cell, the standard deviation over the five seeds of Appendix~\ref{app:r2-seeds} is at most $0.0005$. The element-wise fit of Appendix~\ref{app:ablation} reaches a fitting loss $0.2$ nats below the lower-loss endpoint, yet it lies $0.003$ to $0.008$ average R@1 below the tensor-wise fit in every direction. On held-out training clips, the tensor-wise fit has the lowest loss of the models at the $840$-exemplar budget. Each retrieval direction of \(\mathcal{P}_{\mathrm{fit}}\) enters Eq.~\ref{eq:alphawise_objective} with both directed terms of its modality pair, so the joint objective averages six directed terms, which equals Eq.~\ref{eq:alphawise_objective} up to a constant factor of two, and a pair-specific objective sums its two. As in the released implementation of AudioCLIP \citep{guzhov2021audioclipextendingclipimage}, an objective that involves audio also carries a penalty on the kernel norms of the learnable time-frequency transform of the audio encoder. This penalty stays between $0.10$ and $0.12$ along every segment scanned in Appendix~\ref{sec:barrier} and is excluded from the reported fitting losses. Ablations on N+EWC that need a single model use one reference fit of AlphaWiSE (N+EWC), obtained with the original implementation.

\section{\texorpdfstring{Endpoint objectives and baseline configuration}{Endpoint objectives and baseline configuration}}
\label{app:endpoints}

For the EWC instantiation, the model is optimized using AdamW at $5\times10^{-5}$ with weight decay $0.01$ and the exponential decay of N, with a learning rate of $10^{-6}$ for the logit-scale parameters. The diagonal Fisher information is estimated at the parameters of the preceding phase from at most \(64\) minibatches per GPU of the training data of the current phase, lower-bounded by \(\epsilon=10^{-3}\), and upper-bounded by \(10^{4}\). The corresponding objective is
\[
\mathcal{L}_{\mathrm{EWC}}(\theta)
=
\mathcal{L}_{\mathrm{CLIP}}(\theta)
+
\lambda_{\mathrm{ewc}}
\sum_d
F_d
\left(
\theta_d-\theta_d^{(k-1)}
\right)^2,
\]
where \(\mathcal{L}_{\mathrm{CLIP}}\) is the training loss of AudioCLIP, the mean of the six directed terms of Eq.~\ref{eq:directed_infonce} with the penalty of Appendix~\ref{app:fitting}, \(F_d\) is the estimated diagonal Fisher importance of parameter \(d\), \(\theta^{(k-1)}\) denotes the parameters retained from the preceding phase, and \(\lambda_{\mathrm{ewc}}=0.8\) \citep{kirkpatrick2017overcoming}.

For LwF, a frozen copy of the preceding-phase checkpoint provides temperature-softened targets for the audio-image, audio-text, and image-text objectives. Let \(\mathcal{R}_{\mathrm{dist}}\) denote the set of directed retrieval pairs used for distillation during training. The loss is
\[
\mathcal{L}_{\mathrm{LwF}}
=
\mathcal{L}_{\mathrm{CLIP}}
+
\lambda_{\mathrm{LwF}}T^2
\sum_{r\in\mathcal{R}_{\mathrm{dist}}}
\operatorname{CE}\!\left(q_r^{\mathrm{old}}(T),q_r(T)\right),
\]
where \(q_r(T)\) and \(q_r^{\mathrm{old}}(T)\) are the similarity distributions of direction \(r\) at temperature \(T\) under the current and the frozen model, CE is the cross-entropy, \(T=2\), \(\lambda_{\mathrm{LwF}}=0.1\), and \(\mathcal{R}_{\mathrm{dist}}\) holds one direction of each modality pair, audio-to-image, audio-to-text, and image-to-text \citep{li2017learningforgetting}. LwF uses the optimizer and schedule of N. For iCaRL, we use the optimizer and schedule of N and replay a fixed memory of \(840\) exemplars \citep{rebuffi2017icarl}, with \(\mathcal{L}_{\mathrm{CLIP}}\) as the training loss.

The iCaRL run and the DER and C-CLIP runs of Appendix~\ref{app:r2-baselines} use a smaller training set than the N, EWC, and LwF runs. With the replayed memory, their phases 1 to 7 hold $7{,}962$, $2{,}007$, $1{,}845$, $2{,}453$, $1{,}324$, $1{,}105$, and $1{,}735$ training triplets, $18{,}431$ in all against the $342{,}111$ of Appendix~\ref{app:compute}. Without the memory, these are $12{,}551$ new triplets against $336{,}231$, or $3.7\%$. At phase 1, DER instead holds $13{,}838$ triplets, $6{,}716$ of them replayed from phase 0, which brings its seven phases to $24{,}307$ in all.

\subsection{\texorpdfstring{Configuration of DER and C-CLIP}{Configuration of DER and C-CLIP}}
\label{app:r2-baselines}

DER and C-CLIP, the two further baselines of Table~\ref{tab:continual_results_840}, were run once in the forward class order, from the shared phase-0 checkpoint, and on the same class partition as the other baselines. Each ran for $30$ epochs per phase, and no hyperparameter was tuned per retrieval direction. Every baseline of Table~\ref{tab:continual_results_840} replays its $840$-exemplar memory during training, except that DER replays $6{,}716$ phase-0 triplets at phase 1. The shared phase-0 checkpoint is phase 0 of the N run, which trains the pretrained encoders of Section~\ref{sec:implementation} on the $188{,}645$ triplets of the phase-0 classes for $30$ epochs. DER follows the dynamically expandable representation of \citet{yan2021dynamically}, adapted to the contrastive backbone. At phases 2 to 7, the frozen phase-$(p-1)$ model serves as a second feature extractor. Its audio, image, and text embeddings are concatenated with those of the current model, and a trainable head projects them back to the embedding size. The three-pair contrastive loss is computed on these projected embeddings. The training objective adds an auxiliary contrastive loss on the current-phase samples with weight $1$ and an L1 sparsity term on the audio encoder with weight $10^{-5}$. At phase 1, the backbone is fine-tuned with SGD, without the frozen phase-0 model as a second feature extractor. At phases 2 to 7, the entire backbone is trained with AdamW at $5\times10^{-5}$ with weight decay $0.01$, and the temperature parameters use a learning rate of $10^{-6}$. The epoch-$30$ checkpoint of each phase is both the evaluated model and the initialization of the next phase. The evaluated model is the backbone alone, without the fusion head. Its representation therefore never grows, and its checkpoint keeps the same $499$ tensors as every other method, so DER can serve as an endpoint.

C-CLIP follows \citet{liu2025cclip}. Within a phase, the backbone is frozen except for the learnable filterbank of the audio front end and the three temperatures. Rank-$16$ LoRA adapters with scale $32$ and dropout $0.1$ adapt the two MLP projections of every image and text block and the output layer of the audio encoder. The adapted layers number $49$ in all, and the attention projections are left unadapted. The adapters are trained together with a $512$-unit projector for the contrastive knowledge consolidation loss against the frozen previous-phase model. That loss has weight $1$ and temperature $0.07$. At the phase boundary, the adapters are merged into the backbone with coefficient $0.5$ and discarded, so the checkpoint keeps the same $499$ tensors as every other method. C-CLIP uses the optimizer and schedule of the sequential fine-tuning endpoint, SGD at $5\times10^{-5}$ with Nesterov momentum $0.9$ and weight decay $5\times10^{-4}$. Rank, scale, dropout, merge coefficient, and temperature are the implementation's defaults.

Every count of Appendix~\ref{app:wording} is made within a pairing, and Table~\ref{tab:pairs} uses both methods as endpoints. For N+DER, the refit of rule~(iv) of Appendix~\ref{app:guidance} exceeds DER in every cell, and the N+C-CLIP fit exceeds both endpoints on I$\rightarrow$A and I$\rightarrow$T, as Appendix~\ref{sec:sensitivity} reports.

\section{\texorpdfstring{Why tensor-wise coefficients}{Why tensor-wise coefficients}}
\label{app:tensor}

\subsection{\texorpdfstring{The nested classes and Proposition~\ref{prop:gap}}{The nested classes and Proposition}}
\label{app:argument}

The coefficient classes are nested, $\mathcal{A}_{\mathrm{glob}}=\{a\mathbf{1}\}\subset\mathcal{A}_{\mathrm{layer}}\subset\mathcal{A}_{\mathrm{tensor}}=[0,1]^P\subset\mathcal{A}_{\mathrm{elem}}$, where the layer-wise class uses the $40$ layer groups of Appendix~\ref{sec:groups}. The type-wise class of six groups lies between the global and the tensor-wise class, and every one of these classes contains both endpoints. Let $\Delta_p=\theta^{\mathrm{un}}_p-\theta^{\mathrm{reg}}_p$, so that the blend of Eq.~\ref{eq:alphawise_fusion} is $\tilde\theta(\boldsymbol{\alpha})=\theta^{\mathrm{reg}}+\sum_p\alpha_p\Delta_p$. A second-order expansion of the objective around the stability endpoint, under the Gauss-Newton or Fisher curvature $H\succeq0$ that also underlies EWC, gives
{\begin{equation}
Q(\boldsymbol{\alpha})=L(\theta^{\mathrm{reg}})+b^{\top}\boldsymbol{\alpha}+\tfrac12\boldsymbol{\alpha}^{\top}A\boldsymbol{\alpha},\qquad b_p=\langle g_p,\Delta_p\rangle,\quad A_{pq}=\Delta_p^{\top}H_{pq}\Delta_q,
\label{eq:quad}
\end{equation}}
where $g_p$ is the gradient of the objective with respect to tensor $p$ at the stability endpoint, so that $b_p$ is the inner product of Eq.~\ref{eq:alphawise_beta_gradient} evaluated there.

{\begin{proposition}[Cost of tying coefficients]
\label{prop:gap}
Let $\boldsymbol{\alpha}^{\star}=-A^{-1}b$ minimize Eq.~\ref{eq:quad} with $A\succ0$, and let $\boldsymbol{\alpha}^{\mathcal{S}}$ minimize Eq.~\ref{eq:quad} among vectors constant on each group of a tying pattern $\mathcal{S}$. Then $Q(\boldsymbol{\alpha}^{\mathcal{S}})-Q(\boldsymbol{\alpha}^{\star})=\tfrac12(\boldsymbol{\alpha}^{\mathcal{S}}-\boldsymbol{\alpha}^{\star})^{\top}A(\boldsymbol{\alpha}^{\mathcal{S}}-\boldsymbol{\alpha}^{\star})\ge0$. For $A=\operatorname{diag}(c_1,\ldots,c_P)$, the tied value of group $k$ is the curvature-weighted mean $\gamma^{\mathcal{S}}_k=\sum_{p\in k}c_p\alpha^{\star}_p/\sum_{p\in k}c_p$ of the tensor-wise optima $\alpha^{\star}_p=-b_p/c_p$ and
\begin{equation}
Q(\boldsymbol{\alpha}^{\mathcal{S}})-Q(\boldsymbol{\alpha}^{\star})=\tfrac12\sum_{k}\sum_{p\in k}c_p\bigl(\alpha^{\star}_p-\gamma^{\mathcal{S}}_k\bigr)^2 .
\label{eq:gap-diag}
\end{equation}
\end{proposition}}

Appendix~\ref{app:proof} gives the general form of the proposition and its proof.

The size of the endpoint difference depends on the tensor type. At phase~7 of N+EWC, the median relative difference is $0.22$ for MLP and convolution weights, $0.16$ for attention weights, $0.04$ for normalization parameters, and $0.01$ for biases. The curvature weights $c_p$ grow quadratically with $\Delta_p$, so the weight tensors, which differ most, dominate the curvature-weighted mean that a tied coefficient takes. In the N+EWC column of Table~\ref{tab:alphatype}, the weight tensors take $0.25$ to $0.31$, whereas the normalization and bias tensors take $0.44$ to $0.48$. A tied coefficient therefore pulls the normalization and bias tensors toward the values of the weight tensors.

\subsection{\texorpdfstring{General form of Proposition~\ref{prop:gap}, proof and remarks}{General form of Proposition, proof and remarks}}
\label{app:proof}

Proposition~\ref{prop:gap} extends to any tying pattern encoded by an indicator matrix.

{\begin{proposition}[Cost of tying coefficients, general form]
\label{prop:gap-general}
Assume $A\succ0$ and let $\boldsymbol{\alpha}^{\star}=-A^{-1}b$ be the tensor-wise minimizer of Eq.~\ref{eq:quad}. For a tying pattern $\mathcal{S}$ with $G$ non-empty groups, encoded by the $P\times G$ indicator matrix $M$ with $M_{pk}=1$ if tensor $p$ belongs to group $k$ and $0$ otherwise, one nonzero per row, so that $\boldsymbol{\alpha}=M\gamma$ and $M^{\top}AM\succ0$, let $\boldsymbol{\alpha}^{\mathcal{S}}=M\gamma^{\mathcal{S}}$ with $\gamma^{\mathcal{S}}=-(M^{\top}AM)^{-1}M^{\top}b$ be the tied minimizer. Then
\begin{equation}
Q(\boldsymbol{\alpha}^{\mathcal{S}})-Q(\boldsymbol{\alpha}^{\star})=\tfrac12\,(\boldsymbol{\alpha}^{\mathcal{S}}-\boldsymbol{\alpha}^{\star})^{\top}A\,(\boldsymbol{\alpha}^{\mathcal{S}}-\boldsymbol{\alpha}^{\star})\;\ge0 .
\label{eq:gap-general}
\end{equation}
If $H$ has no cross-tensor blocks, then $A=\operatorname{diag}(c_1,\ldots,c_P)$ with $c_p=\Delta_p^{\top}H_{pp}\Delta_p$, the tensor-wise optima are $\alpha^{\star}_p=-b_p/c_p$, the tied optimum of every group $k$ is the curvature-weighted mean $\gamma^{\mathcal{S}}_k=\sum_{p\in k}c_p\alpha^{\star}_p/\sum_{p\in k}c_p$, and the gap is the curvature-weighted spread of Eq.~\ref{eq:gap-diag}. For the global class, where $G=1$, the right-hand side of Eq.~\ref{eq:gap-diag} equals $\tfrac12\bigl(\sum_pc_p\bigr)\operatorname{Var}_c(\boldsymbol{\alpha}^{\star})$, where $\operatorname{Var}_c$ denotes the variance of the entries of $\boldsymbol{\alpha}^{\star}$ under the weights $c_p/\sum_qc_q$.
\end{proposition}}
{\begin{proof}
Since $\nabla Q(\boldsymbol{\alpha}^{\star})=b+A\boldsymbol{\alpha}^{\star}=0$, the second-order Taylor expansion of the quadratic $Q$ around $\boldsymbol{\alpha}^{\star}$ is exact and has no linear term, which gives Eq.~\ref{eq:gap-general} for any $\boldsymbol{\alpha}$, in particular for $\boldsymbol{\alpha}^{\mathcal{S}}$, and nonnegativity follows from $A\succeq0$. The tied minimizer follows by substituting $\boldsymbol{\alpha}=M\gamma$ into Eq.~\ref{eq:quad} and setting the gradient in $\gamma$ to zero. When $A$ is diagonal, $Q$ separates as $\sum_p\bigl(b_p\alpha_p+\tfrac12c_p\alpha_p^2\bigr)$, so $\alpha^{\star}_p=-b_p/c_p$. Minimizing $\sum_{p\in k}\bigl(b_p\gamma+\tfrac12c_p\gamma^2\bigr)$ over $\gamma$ gives $\gamma^{\mathcal{S}}_k=-\sum_{p\in k}b_p/\sum_{p\in k}c_p=\sum_{p\in k}c_p\alpha^{\star}_p/\sum_{p\in k}c_p$, and Eq.~\ref{eq:gap-diag} is Eq.~\ref{eq:gap-general} written out for a diagonal $A$.
\end{proof}}

{\begin{remark}
The proposition is stated for the unconstrained quadratic. If $A$ is only positive semidefinite and $b$ lies in the range of $A$, the pseudo-inverse replaces $A^{-1}$ and Eq.~\ref{eq:gap-general} is unchanged. A script in the code release checks every identity of the proposition numerically on $50$ random instances, with block-diagonal and with full curvature matrices. The same script checks the gradient identity of Eq.~\ref{eq:alphawise_beta_gradient} by automatic differentiation. The method itself searches $[0,1]^P$ through the sigmoid. The box constraint preserves the nesting of the classes and with it the inequality between their minima. For diagonal $A$, the tied value of group $k$ becomes $\operatorname{clip}(\gamma^{\mathcal{S}}_k,0,1)$, where $\gamma^{\mathcal{S}}_k$ is the curvature-weighted mean of the unconstrained tensor-wise optima. Equation~\ref{eq:gap-diag} therefore holds unchanged for every group whose tensor-wise optima and tied value are all interior, and only clipped coefficients alter it. The constraint is inactive in the seed-42 fits of Appendix~\ref{app:r2-seeds}, since none of the $499$ coefficients of N+EWC, N+LwF, or N+iCaRL lies within $0.01$ of a boundary at any phase. Under the quadratic model, the curvature-weighted spread of Eq.~\ref{eq:gap-diag} then describes these fits for every coefficient. When the segment carries a barrier, as for N+EWC in Table~\ref{tab:barrier}, the measured scan of Appendix~\ref{sec:barrier} gives the conclusion of the proposition without the quadratic model. At the three scanned phases, the tensor-wise refit of rule~(iv) of Appendix~\ref{app:guidance} reaches $2.71$, $1.00$, and $0.68$ nats, against $2.73$, $1.02$, and $0.69$ for the best grid point, and so lies below every grid point.
\end{remark}}

\subsection{\texorpdfstring{Variance decomposition and coefficient by type}{Variance decomposition and coefficient by type}}
\label{sec:decomp}

\begin{table}[!htb]
\centering
\footnotesize
\caption{Variance decomposition of the learned coefficients of the seed-42 fits of Appendix~\ref{app:r2-seeds}, as the mean over phases 1 to 7. The between-group fraction is the share of the variance of the $499$ coefficients that the group means explain under each tying pattern. The within-layer columns give the standard deviation of the coefficients inside one block, stage, or module and the median within-group range over the $40$ layer groups.}
\label{tab:decomp}
\vspace{0.2cm}
\renewcommand{\arraystretch}{0.95}
\begin{tabular}{lcccccc}
\toprule
& & \multicolumn{3}{c}{Between-group fraction} & \multicolumn{2}{c}{Within layer} \\
\cmidrule(lr){3-5}\cmidrule(lr){6-7}
Pairing & std of $\alpha$ & encoder (4) & layer (40) & type (6) & std & median range \\
\midrule
N+EWC   & 0.183 & 0.02 & 0.15 & 0.19 & 0.169 & 0.49 \\
N+LwF   & 0.157 & 0.19 & 0.29 & 0.22 & 0.133 & 0.42 \\
N+iCaRL & 0.179 & 0.02 & 0.11 & 0.32 & 0.169 & 0.49 \\
\bottomrule
\end{tabular}
\end{table}
\begin{table}[!htb]
\centering
\footnotesize
\caption{Learned coefficient $\alpha$ by tensor type, where $\alpha$ is the weight on the plastic endpoint N. Each entry is the mean over the tensors of one type and over phases 1 to 7 of the seed-42 fits of the three main pairings.}
\label{tab:alphatype}
\vspace{0.2cm}
\begin{tabular}{lcccc}
\toprule
Tensor type & Tensors & N+EWC & N+LwF & N+iCaRL \\
\midrule
Attention weights & 48 & 0.25 & 0.76 & 0.83 \\
MLP, convolution, projection and embedding weights & 126 & 0.31 & 0.63 & 0.72 \\
Normalization weights and biases & 210 & 0.45 & 0.53 & 0.55 \\
Other biases & 64 & 0.44 & 0.56 & 0.54 \\
Attention biases & 48 & 0.48 & 0.58 & 0.58 \\
Temperatures & 3 & 0.47 & 0.53 & 0.41 \\
\bottomrule
\end{tabular}
\end{table}

Most of the variation that the tensor-wise fit learns lies inside single blocks, as Table~\ref{tab:decomp} shows. The encoder groups of that table are the three encoders and the temperature scalars, the layer groups are the $40$ groups of Appendix~\ref{sec:groups}, and the type groups are the six parameter types of Table~\ref{tab:alphatype}. A global coefficient explains none of the variance by definition. Layer-wise coefficients explain $11$ to $29\%$ of that variation. The remaining $71$ to $89\%$ lies between the attention, MLP, normalization, and bias tensors of the same block, stage, or module. Within such a group, the coefficients span a median range of $0.42$ to $0.49$ on a $[0,1]$ scale. The fitting seeds of Appendix~\ref{app:r2-seeds} reproduce this within-block variation tensor by tensor. Neither of the other partitions accounts for most of the variation. A coefficient per encoder explains $2$ to $19\%$, and the type partition of Table~\ref{tab:alphatype} explains $19$ to $32\%$. The decomposition describes the learned coefficients, whereas Appendix~\ref{app:ablation} measures how much of the retrieval gain the fit of each coarser class keeps. There, the layer-wise fit keeps the gain under a hand-designed partition, and the type-wise and global fits do not.

Table~\ref{tab:alphatype} gives the learned coefficient by tensor type. In every pairing, the attention and other weight matrices move furthest from the fixed blend, toward the stability endpoint for N+EWC and toward N for N+LwF and N+iCaRL. The normalization and bias tensors stay between $0.44$ and $0.58$, and the three temperatures between $0.41$ and $0.53$. A global coefficient cannot express this assignment.

The tensor-wise derivatives disagree in sign at the shared initialization $\alpha_p=0.5$. At phase~7, $37\%$ of the $499$ tensors of N+EWC have a derivative of Eq.~\ref{eq:alphawise_beta_gradient} whose sign is opposite to that of the summed derivative, which a global coefficient would follow. The share is $35$ to $45\%$ for N+iCaRL, N+LwF, and N+DER at the same phase. The opposite-sign tensors of N+EWC carry $34\%$ of the summed gradient magnitude, whose tensor-wise values span more than three orders of magnitude, from a median of $9\times10^{-4}$ to a maximum of $2.2$ nats per unit of $\beta$. A global update therefore moves a third of the tensors, which carry a third of the summed gradient magnitude, against their own descent direction.

\subsection{\texorpdfstring{Granularity ablation}{Granularity ablation}}
\label{app:ablation}

Section~\ref{sec:granularity} reports the granularity ablation of Table~\ref{tab:granularity}. This appendix adds the learned global coefficient, the grid of fixed global coefficients, and the other two main pairings.

The row ``Learned global coefficient'' of Table~\ref{tab:granularity} is WiSE-FT with a learned coefficient. It is produced by the same code as the tensor-wise fit, with the tying pattern set to one group, so that one logit $\beta$ with $\alpha=\sigma(\beta)$ applies to all $499$ tensors. It follows the fitting protocol of Appendix~\ref{app:fitting} with seed $42$, and the other learned rows differ from it only in the tying pattern. The seed-42 refit is the reference fit rerun with the release version of the code under that protocol, and the loss column is read against $0.69$ nats, the best grid point on its batches (Table~\ref{tab:barrier}). Table~\ref{tab:granularity} therefore separates the two components of the second contribution, namely learning a coefficient and learning it per tensor. The learned global coefficient differs from the fixed blend by $0.006$ A$\rightarrow$T, $0.005$ I$\rightarrow$A, and $0.001$ I$\rightarrow$T R@1 and by $0.007$, $0.001$, and $-0.001$ mAP, so learning a single coefficient adds little. On I$\rightarrow$T, the fixed blend and the learned global coefficient hold the highest values of the table. Under the joint objective, the tensor-wise fit moves the weights toward EWC for its audio-related retrieval and exceeds both rows on A$\rightarrow$T and I$\rightarrow$A (Appendix~\ref{sec:sensitivity}).

A fitted global coefficient can only reach points of the segment between the endpoints, so the segment bounds the global class. Table~\ref{tab:grid} evaluates both endpoints and every interior grid point of the segment in retrieval. The blends use the N+EWC checkpoints at $840$ exemplars in the forward order, and the loss column scores each of them at phase 7 on the batches of the scans in Table~\ref{tab:barrier}. For comparison, the bottom block repeats the learned global coefficient and the AlphaWiSE fit of Table~\ref{tab:granularity}. Retrieval along the segment peaks in two regions. Near the EWC end, $a=0.2$ is the best grid point on A$\rightarrow$T, at $0.3090$ / $0.3976$. Near the N end, $a=0.9$ is the best on I$\rightarrow$A and I$\rightarrow$T, at $0.4229$ / $0.3536$ and $0.2728$ / $0.3739$, but on A$\rightarrow$T it falls to $0.2149$ R@1, below both endpoints and $0.089$ below AlphaWiSE. No grid point exceeds AlphaWiSE on both A$\rightarrow$T and I$\rightarrow$A. The coefficient closest to AlphaWiSE, $a=0.2$, has the highest mean R@1 of the grid over the three directions and is within $0.006$ of AlphaWiSE in every average cell. It lies above AlphaWiSE by $0.005$ / $0.003$ on A$\rightarrow$T and below it by $0.002$ / $0.002$ on I$\rightarrow$A and by $0.001$ / $0.003$ on I$\rightarrow$T. The tensor-wise fit reaches the same neighborhood from the initialization at $0.5$, by descent on the fitting set alone. The fitting loss does not select that coefficient, and selecting it would require a search over the test data.

The fitting loss was scanned along the segment at phases 1, 4, and 7. At each of these phases, the EWC endpoint, $a=0$, is the best grid point, at $2.728$, $1.018$, and $0.690$ nats. The point $a=0.1$ is the next-best grid point, at $2.806$, $1.036$, and $0.693$ nats, and $a=0.2$ reaches $0.84$ nats at phase 7. AlphaWiSE exceeds $a=0$ in every average cell. It also exceeds $a=0.1$ on I$\rightarrow$A and I$\rightarrow$T, by $0.011$ R@1 and $0.015$ mAP on I$\rightarrow$T, and stays within $0.003$ of it on A$\rightarrow$T.

Started at $0.5$, the learned global coefficient lies within $0.008$ R@1 or mAP and $0.2$ nats of the fixed blend in Table~\ref{tab:grid}. Under the same protocol, the tensor-wise fit improves on the learned global coefficient of Table~\ref{tab:granularity} by $0.070$ A$\rightarrow$T and $0.060$ I$\rightarrow$A R@1, and by $0.042$ mean R@1 over the three directions. Each tensor-wise coefficient follows its own derivative, whereas one tied coefficient follows their sum. At the shared initialization at phase~7, that sum has the opposite sign to the derivatives of $37\%$ of the tensors, as Appendix~\ref{sec:decomp} shows.

\begin{table}[!htb]
\centering
\scriptsize
\setlength{\tabcolsep}{4pt}
\caption{Both endpoints and every interior fixed global coefficient of the grid on the N+EWC pairing, $\tilde\theta=a\,\theta^{\mathrm{un}}+(1-a)\,\theta^{\mathrm{reg}}$ for $a\in\{0,0.1,\ldots,1\}$, with the fitting-set loss in nats at phase 7 and the R@1 / mAP averaged over phases 1 to 7. The bottom block repeats two rows of Table~\ref{tab:granularity}. Within each block, the best R@1 and the best mAP of every column are in bold.}
\label{tab:grid}
\vspace{0.2cm}
\renewcommand{\arraystretch}{0.92}
\begin{tabular}{lcccc}
\toprule
Model & $L$ (phase 7) & A$\rightarrow$T & I$\rightarrow$A & I$\rightarrow$T \\
\midrule
$a=0$, EWC endpoint & 0.69 & 0.2900 / 0.3792 & 0.4068 / 0.3408 & 0.2056 / 0.2784 \\
$a=0.1$ & 0.69 & 0.3065 / 0.3947 & 0.4191 / 0.3456 & 0.2191 / 0.2936 \\
$a=0.2$ & 0.84 & \textbf{0.3090} / \textbf{0.3976} & 0.4208 / 0.3463 & 0.2298 / 0.3065 \\
$a=0.3$ & 1.30 & 0.2913 / 0.3811 & 0.4087 / 0.3421 & 0.2362 / 0.3162 \\
$a=0.4$ & 2.35 & 0.2571 / 0.3471 & 0.3871 / 0.3342 & 0.2364 / 0.3215 \\
$a=0.5$, fixed blend, WiSE-FT & 3.66 & 0.2273 / 0.3119 & 0.3566 / 0.3261 & 0.2330 / 0.3253 \\
$a=0.6$ & 4.61 & 0.2113 / 0.2893 & 0.3479 / 0.3232 & 0.2313 / 0.3285 \\
$a=0.7$ & 4.99 & 0.2078 / 0.2879 & 0.3712 / 0.3306 & 0.2396 / 0.3382 \\
$a=0.8$ & 4.64 & 0.2197 / 0.3122 & 0.4042 / 0.3448 & 0.2553 / 0.3554 \\
$a=0.9$ & 2.88 & 0.2149 / 0.3344 & \textbf{0.4229} / \textbf{0.3536} & \textbf{0.2728} / \textbf{0.3739} \\
$a=1$, N endpoint & 1.09 & 0.2377 / 0.3602 & 0.4038 / 0.3487 & 0.2470 / 0.3435 \\
\midrule
Learned global, started at $0.5$ & 3.49 & 0.2334 / 0.3190 & 0.3618 / 0.3275 & \textbf{0.2338} / \textbf{0.3241} \\
Learned tensor-wise, AlphaWiSE reference fit & 0.93 & \textbf{0.3037} / \textbf{0.3946} & \textbf{0.4225} / \textbf{0.3485} & 0.2304 / 0.3090 \\
\bottomrule
\end{tabular}
\end{table}

The other two main pairings show the same ordering under the same protocol, with the tensor-wise fit above the learned global coefficient on A$\rightarrow$T and I$\rightarrow$A. Table~\ref{tab:learnedwiseft} lists the fixed blend, the learned global coefficient, and AlphaWiSE for the three pairings. The learned coefficient is one global coefficient fitted from $0.5$ with the loss, fitting set, and schedule of AlphaWiSE. AlphaWiSE exceeds the learned coefficient in $14$ of the $18$ average cells, among them all $12$ A$\rightarrow$T and I$\rightarrow$A cells. On N+LwF, the global coefficient stays within $0.002$ R@1 of the fixed blend in every direction. On A$\rightarrow$T, it reaches $0.2298$ / $0.3548$, against $0.2295$ / $0.3541$ for the fixed blend. On N+iCaRL, it gains $0.014$ A$\rightarrow$T R@1 over the fixed blend, at $0.2015$ against $0.1871$. The tensor-wise fits of these two pairings reach $0.2397$ and $0.2191$ A$\rightarrow$T R@1, $0.010$ and $0.018$ above the global coefficient.

The tensor-wise coefficients are reproducible. They correlate at $0.95$ to $0.99$ across the five fitting seeds and move in one direction with the memory budget, as Appendices~\ref{app:r2-seeds} and~\ref{app:r2-memory} show.

\begin{table}[!htb]
\centering
\footnotesize
\setlength{\tabcolsep}{5pt}
\caption{WiSE-FT with its fixed coefficient and with a learned coefficient on the three main pairings, at $840$ exemplars in the forward order, R@1 / mAP averaged over phases 1 to 7. The AlphaWiSE rows are the five-seed means of Table~\ref{tab:continual_results_840}. Within each pairing, the best R@1 and the best mAP of every column are in bold.}
\label{tab:learnedwiseft}
\vspace{0.2cm}
\renewcommand{\arraystretch}{0.92}
\begin{tabular}{llccc}
\toprule
Pairing & Model & A$\rightarrow$T & I$\rightarrow$A & I$\rightarrow$T \\
\midrule
\multirow{3}{*}{N + EWC} & WiSE-FT, fixed $\alpha\equiv0.5$ & 0.2273 / 0.3119 & 0.3566 / 0.3261 & 0.2330 / \textbf{0.3253} \\
 & WiSE-FT, learned $\alpha$ & 0.2334 / 0.3190 & 0.3618 / 0.3275 & \textbf{0.2338} / 0.3241 \\
 & AlphaWiSE, tensor-wise & \textbf{0.3034} / \textbf{0.3944} & \textbf{0.4224} / \textbf{0.3485} & 0.2305 / 0.3092 \\
\addlinespace[1.5pt]
\multirow{3}{*}{N + LwF} & WiSE-FT, fixed $\alpha\equiv0.5$ & 0.2295 / 0.3541 & 0.4096 / 0.3491 & 0.2392 / 0.3427 \\
 & WiSE-FT, learned $\alpha$ & 0.2298 / 0.3548 & 0.4096 / 0.3492 & 0.2411 / 0.3441 \\
 & AlphaWiSE, tensor-wise & \textbf{0.2397} / \textbf{0.3657} & \textbf{0.4109} / \textbf{0.3498} & \textbf{0.2521} / \textbf{0.3511} \\
\addlinespace[1.5pt]
\multirow{3}{*}{N + iCaRL} & WiSE-FT, fixed $\alpha\equiv0.5$ & 0.1871 / 0.3151 & 0.4052 / 0.3464 & 0.2737 / 0.3790 \\
 & WiSE-FT, learned $\alpha$ & 0.2015 / 0.3298 & 0.4071 / 0.3477 & \textbf{0.2756} / \textbf{0.3803} \\
 & AlphaWiSE, tensor-wise & \textbf{0.2191} / \textbf{0.3459} & \textbf{0.4077} / \textbf{0.3490} & 0.2638 / 0.3654 \\
\bottomrule
\end{tabular}
\end{table}

The learned global coefficient and the layer-wise row use the coefficient classes of the two-checkpoint learned soups of \citet{wortsman2022modelsoupsaveragingweights}, as Appendix~\ref{app:rules} explains. AlphaWiSE differs from those soups in three respects. Its objective is the InfoNCE retrieval loss on the exemplar memory and a class-balanced draw of the current phase, and its inputs are continual-learning checkpoints with a loss barrier on the segment between them. It also learns one coefficient per tensor rather than per model or per layer.

\subsection{\texorpdfstring{The layer-wise partition}{The layer-wise partition}}
\label{sec:groups}

The $40$ groups of the layer-wise class follow the architecture of the backbone. The image transformer contributes its $12$ residual blocks, with $12\times12=144$ tensors, and the text transformer its $12$ residual blocks, with $144$ tensors. The audio encoder contributes its four ResNeXt stages \texttt{layer1} to \texttt{layer4}, with $30$, $39$, $57$, and $30$ tensors, and its five attention modules \texttt{att1} to \texttt{att5}, with $6$ tensors each. Two further audio groups are the stem, whose input convolution, normalization, and learnable filterbank hold $7$ tensors, and the projection head, which holds $2$. Each transformer also contributes an embedding module and a head. The image embedding module holds the class embedding, patch convolution, positional embedding, and \texttt{ln\_pre}, $5$ tensors in all, and the image head holds \texttt{ln\_post} and the projection, $3$ tensors. The text embedding module holds the token and positional embeddings, $2$ tensors, and the text head holds \texttt{ln\_final} and the projection, $3$ tensors. The three temperature scalars form the last group, which brings the total to $499$ tensors.

\section{\texorpdfstring{Checkpoint compatibility and sensitivity to the endpoints}{Checkpoint compatibility and sensitivity to the endpoints}}
\label{app:compat}

\subsection{\texorpdfstring{Definitions}{Definitions}}
\label{app:defs}

{\begin{definition}[Structural compatibility]
Two checkpoints are structurally compatible if they share the same ordered key set and every corresponding tensor has the same shape. Structural compatibility makes Eq.~\ref{eq:alphawise_fusion} well defined, is checked at load time, and is the only requirement the implementation enforces.
\end{definition}}

{\begin{definition}[Trajectory compatibility]
For structurally compatible checkpoints and an objective $L$, the barrier along the global segment $\tilde\theta(a\mathbf{1})$ is
\begin{equation}
B(\theta^{\mathrm{un}},\theta^{\mathrm{reg}})=\max_{a\in[0,1]}\Bigl[L\bigl(\tilde\theta(a\mathbf{1})\bigr)-\bigl((1-a)L(\theta^{\mathrm{reg}})+aL(\theta^{\mathrm{un}})\bigr)\Bigr],
\label{eq:barrier}
\end{equation}
the linear mode connectivity criterion of \citet{frankle2020lmc} in the form of \citet{entezari2022permutation}, evaluated on the grid $a\in\{0,0.1,\ldots,1\}$ with $L$ the fitting objective on the fitting set, $11$ loss evaluations in all. A pairing is trajectory-compatible along the global segment if $B=0$, so that no grid point lies above the chord, the linear interpolation of the two endpoint losses.
\end{definition}}

Every pairing shares its phase-0 checkpoint and phase sequence, so its two networks correspond neuron to neuron by construction and need no permutation alignment \citep{entezari2022permutation,ainsworth2023rebasin}. The endpoints are also close, with relative parameter distances of $0.007$ to $0.076$ at phase~7, as Appendix~\ref{sec:props} measures. A shared phase-0 checkpoint does not by itself make the segment barrier-free \citep{juneja2023linearconnectivity,wen2023modeconnectivitycil}, so compatibility along the segment must be measured, and the barrier of Eq.~\ref{eq:barrier} measures it.

\subsection{\texorpdfstring{Measured barriers}{Measured barriers}}
\label{sec:barrier}

Table~\ref{tab:barrier} scans the fitting-set loss along the global segment of every pairing at phases 1, 4, and 7 and scores the learned tensor-wise blend on the same batches. Every loss is computed with batch size $128$ and seed $42$ under the joint objective of Eq.~\ref{eq:alphawise_objective}, which averages the two directed InfoNCE terms of Eq.~\ref{eq:directed_infonce} over the three modality pairs. Each checkpoint is evaluated with its own learned temperature, so iCaRL and DER, whose logit scale is larger, have $L$ above $\ln128$, and the loss $L$ and the barrier $B$ are therefore compared within a pairing rather than across pairings. The rows marked $\ast$ hold the second fit of rule~(iv) of Appendix~\ref{app:guidance}, started from the lower-loss endpoint at $\alpha_p=0.99$ for N+iCaRL and N+C-CLIP and at $0.01$ for N+EWC, LwF+EWC, and the reverse order. The loss of the second N+DER fit was scanned at phase 7 only, as Appendix~\ref{sec:sensitivity} reports.

\begin{table}[!htb]
\centering
\footnotesize
\setlength{\tabcolsep}{4pt}
\caption{Fitting-set loss in nats along the global segment $\tilde\theta(a\mathbf{1})$ and at the learned tensor-wise blend. $B$ is the barrier of Eq.~\ref{eq:barrier} with its grid maximizer in parentheses, $a^{\star}$ is the best global coefficient on the grid $\{0,0.1,\ldots,1\}$, where $a=0$ is $\theta^{\mathrm{reg}}$ and $a=1$ is $\theta^{\mathrm{un}}$, and the last column is the learned loss minus $L(a^{\star})$. Rows marked $\ast$ are second fits under rule~(iv). Bold marks a learned loss at or below $L(a^{\star})$.}
\label{tab:barrier}
\vspace{0.1cm}
\begin{tabular}{llcccccc}
\toprule
Pairing & Phase & $L(\theta^{\mathrm{reg}})$ & $L(a{=}0.5)$ & $B$ & $a^{\star}$ & learned tensor-wise & learned $-\,L(a^{\star})$ \\
\midrule
N+EWC & 1 & 2.73 & 5.14 & 2.62 (0.6) & 0.0 & 2.81 & 0.08 \\
N+EWC & 4 & 1.02 & 4.40 & 3.91 (0.7) & 0.0 & 1.32 & 0.31 \\
N+EWC & 7 & 0.69 & 3.66 & 4.02 (0.7) & 0.0 & 0.93 & 0.24 \\
N+EWC$^{\ast}$ & 1 & 2.73 & 5.14 & 2.62 (0.6) & 0.0 & \textbf{2.71} & $-0.02$ \\
N+EWC$^{\ast}$ & 4 & 1.02 & 4.40 & 3.91 (0.7) & 0.0 & \textbf{1.00} & $-0.02$ \\
N+EWC$^{\ast}$ & 7 & 0.69 & 3.66 & 4.02 (0.7) & 0.0 & \textbf{0.68} & $-0.01$ \\
N+LwF & 1 & 2.98 & 2.91 & 0.00 & 0.8 & \textbf{2.86} & $-0.03$ \\
N+LwF & 4 & 1.72 & 2.04 & 0.31 (0.5) & 0.1 & \textbf{1.70} & $-0.01$ \\
N+LwF & 7 & 1.16 & 1.72 & 0.60 (0.5) & 1.0 & \textbf{1.09} & 0.00 \\
N+iCaRL & 1 & 5.60 & 3.99 & 0.00 & 1.0 & 3.12 & 0.21 \\
N+iCaRL & 4 & 6.58 & 4.20 & 0.27 (0.3) & 1.0 & 2.61 & 0.87 \\
N+iCaRL & 7 & 6.94 & 4.08 & 0.30 (0.3) & 1.0 & 2.25 & 1.16 \\
N+iCaRL$^{\ast}$ & 1 & 5.60 & 3.99 & 0.00 & 1.0 & \textbf{2.90} & $-0.01$ \\
N+iCaRL$^{\ast}$ & 4 & 6.58 & 4.20 & 0.27 (0.3) & 1.0 & \textbf{1.73} & $-0.01$ \\
N+iCaRL$^{\ast}$ & 7 & 6.94 & 4.08 & 0.30 (0.3) & 1.0 & \textbf{1.09} & 0.00 \\
N+DER & 1 & 5.44 & 3.94 & 0.00 & 1.0 & 3.11 & 0.20 \\
N+DER & 4 & 8.03 & 5.27 & 2.45 (0.9) & 1.0 & 4.91 & 3.17 \\
N+DER & 7 & 8.65 & 5.14 & 2.53 (0.8) & 1.0 & 4.87 & 3.78 \\
LwF+EWC & 1 & 2.73 & 4.90 & 2.15 (0.6) & 0.0 & 3.00 & 0.27 \\
LwF+EWC & 4 & 1.02 & 4.26 & 3.43 (0.6) & 0.0 & 1.73 & 0.71 \\
LwF+EWC & 7 & 0.69 & 3.61 & 3.58 (0.7) & 0.0 & 1.27 & 0.58 \\
N+C-CLIP & 1 & 3.99 & 3.47 & 0.05 (0.3) & 1.0 & 2.99 & 0.08 \\
N+C-CLIP & 4 & 4.43 & 3.40 & 0.37 (0.4) & 1.0 & 2.32 & 0.58 \\
N+C-CLIP & 7 & 4.39 & 3.29 & 0.61 (0.4) & 1.0 & 2.01 & 0.92 \\
LwF+EWC$^{\ast}$ & 1 & 2.73 & 4.90 & 2.15 (0.6) & 0.0 & \textbf{2.71} & $-0.02$ \\
LwF+EWC$^{\ast}$ & 4 & 1.02 & 4.26 & 3.43 (0.6) & 0.0 & \textbf{1.00} & $-0.02$ \\
LwF+EWC$^{\ast}$ & 7 & 0.69 & 3.61 & 3.58 (0.7) & 0.0 & \textbf{0.68} & $-0.01$ \\
N+C-CLIP$^{\ast}$ & 1 & 3.99 & 3.47 & 0.05 (0.3) & 1.0 & \textbf{2.88} & $-0.03$ \\
N+C-CLIP$^{\ast}$ & 4 & 4.43 & 3.40 & 0.37 (0.4) & 1.0 & \textbf{1.71} & $-0.03$ \\
N+C-CLIP$^{\ast}$ & 7 & 4.39 & 3.29 & 0.61 (0.4) & 1.0 & \textbf{1.09} & 0.00 \\
N+EWC, reverse & 1 & 2.12 & 4.36 & 2.26 (0.7) & 0.0 & 2.39 & 0.27 \\
N+EWC, reverse & 4 & 1.11 & 4.03 & 3.37 (0.7) & 0.1 & 1.34 & 0.25 \\
N+EWC, reverse & 7 & 1.14 & 3.49 & 3.56 (0.7) & 0.1 & 1.32 & 0.18 \\
N+EWC, reverse$^{\ast}$ & 1 & 2.12 & 4.36 & 2.26 (0.7) & 0.0 & \textbf{2.12} & 0.00 \\
N+EWC, reverse$^{\ast}$ & 4 & 1.11 & 4.03 & 3.37 (0.7) & 0.1 & \textbf{1.09} & 0.00 \\
N+EWC, reverse$^{\ast}$ & 7 & 1.14 & 3.49 & 3.56 (0.7) & 0.1 & \textbf{1.12} & $-0.02$ \\
\bottomrule
\end{tabular}
\end{table}
\clearpage
\begin{table}[!htb]
\centering
\footnotesize
\setlength{\tabcolsep}{4.5pt}
\caption{Fitting-set loss in nats along the global segment for N+EWC at phase 7, the scan summarized by the phase-7 N+EWC row of Table~\ref{tab:barrier}. The chord interpolates the two endpoint losses linearly, and the maximum excess of the loss over the chord is the barrier $B$ of Eq.~\ref{eq:barrier}.}
\label{tab:loss}
\vspace{0.2cm}
\begin{tabular}{lccccccccccc}
\toprule
$a$ & 0 & 0.1 & 0.2 & 0.3 & 0.4 & 0.5 & 0.6 & 0.7 & 0.8 & 0.9 & 1 \\
\midrule
Loss $L(a)$ & 0.69 & 0.69 & 0.84 & 1.30 & 2.35 & 3.66 & 4.61 & 4.99 & 4.64 & 2.88 & 1.09 \\
Chord & 0.69 & 0.73 & 0.77 & 0.81 & 0.85 & 0.89 & 0.93 & 0.97 & 1.01 & 1.05 & 1.09 \\
Loss minus chord & 0.00 & $-0.04$ & 0.07 & 0.49 & 1.50 & 2.77 & 3.68 & 4.02 & 3.63 & 1.83 & 0.00 \\
\bottomrule
\end{tabular}
\end{table}

No pairing is trajectory-compatible at phases 4 and 7, since every pairing has a barrier there. For five of the seven pairings, the best global coefficient on the grid is an endpoint, so no grid point has a lower fitting loss than the lower-loss endpoint. That endpoint is EWC for N+EWC and LwF+EWC and N for N+iCaRL, N+DER, and N+C-CLIP. N+LwF, whose endpoints nearly coincide, has an interior optimum at phases 1 and 4. The reverse order has one at $a=0.1$ at phases 4 and 7, within $0.03$ nats of the EWC endpoint.

N+EWC has the largest barrier, and Table~\ref{tab:loss} lists its loss along the segment at phase 7. The loss is $0.69$ nats at the EWC endpoint but $3.66$ at $a=0.5$ and $4.99$ at the peak, $a=0.7$, above the uniform-prediction level $\ln128=4.85$. At all three phases, every grid point with $0.3\le a\le0.9$ lies above both endpoints, and these interior points form the high-loss ridge of the segment. A global coefficient fitted by gradient descent from $0.5$ stays on this ridge, as Appendix~\ref{app:ablation} shows. The tensor-wise fit started at $\alpha_p=0.5$ leaves the segment. It reaches $2.81$, $1.32$, and $0.93$ nats at phases 1, 4, and 7, in the learned tensor-wise column of Table~\ref{tab:barrier}. These losses lie below every grid point with $0.3\le a\le 0.9$ and $0.08$ to $0.31$ nats above the lower-loss EWC endpoint. The second fit of rule~(iv), started at the EWC endpoint, reaches $2.71$, $1.00$, and $0.68$ nats, below the best grid point at every phase. In the retrieval evaluation, this second fit keeps the performance of EWC in every direction, whereas the first fit, reported in Table~\ref{tab:continual_results_840}, exceeds EWC in every average cell of Table~\ref{tab:pairs}. For N+LwF, whose endpoints lie $0.007$ apart in relative parameter distance, the fit started at $0.5$ already lies at or below the best grid point at all three phases, at $2.86$, $1.70$, and $1.09$ nats. This pairing therefore needs no second fit.

Among the three main pairings, the initialization has the largest effect on the fitting loss for N+iCaRL. At phase~7, the segment descends monotonically from the iCaRL endpoint, at $6.94$ nats, toward N. Its $0.30$-nat barrier therefore measures the excess of that descent over the chord rather than a ridge. Started at $\alpha_p=0.5$, the fit ends at $2.25$ nats at phase~7, $1.16$ above the best grid point. The second fit of rule~(iv), started at the plastic endpoint N with $\alpha_p=0.99$, reaches $2.90$, $1.73$, and $1.09$ nats at phases 1, 4, and 7, at or below the best grid point at every phase, in the rows marked $\ast$ of Table~\ref{tab:barrier}. In the retrieval evaluation of Table~\ref{tab:pairs}, the fit from the plastic endpoint gives the higher A$\rightarrow$T retrieval and the fit from $0.5$ the larger I$\rightarrow$T gain, as Appendix~\ref{sec:sensitivity} reports.

The reverse class order exchanges the classes of the early and late phases, and its six rows in Table~\ref{tab:barrier} reproduce the N+EWC pattern. Its barriers are $2.3$ to $3.6$ nats, and its best global coefficient lies at the EWC endpoint or at the adjacent grid point. Started at $\alpha_p=0.5$, its fit ends $0.18$ to $0.27$ nats above the best grid point. The second fit of rule~(iv), from the EWC endpoint, reaches $2.12$, $1.09$, and $1.12$ nats in the rows marked $\ast$ and exceeds the EWC endpoint in all six average cells of Table~\ref{tab:pairs}. A shared phase-0 checkpoint and a small separation therefore do not make the global segment barrier-free. The tensor-wise fit is not confined to that segment, since its class contains tensor-wise assignments outside the global class.

\subsection{\texorpdfstring{Counts in Table 1}{Counts in Table 1}}
\label{app:wording}

In Table~\ref{tab:continual_results_840}, AlphaWiSE exceeds the better of its two endpoints in $13$ of the $18$ average cells, over three pairings, three directions, and two metrics. The $13$ cells are the positive entries in the six Average columns of the line labeled ``gain over the better endpoint'' under each AlphaWiSE row. AlphaWiSE exceeds the fixed blend of the same two checkpoints in $25$ of the $36$ average and last-phase cells, among them every A$\rightarrow$T and I$\rightarrow$A average cell. It exceeds its own stability endpoint, whether EWC, LwF, or iCaRL, in all $36$ average and last-phase cells, by at least $0.0063$. Its pairings hold the best average A$\rightarrow$T R@1 and mAP and the best average I$\rightarrow$A R@1 and mAP of the table.

\subsection{\texorpdfstring{Sensitivity to the endpoints}{Sensitivity to the endpoints}}
\label{sec:sensitivity}

\begin{table}[!htb]
\centering
\scriptsize
\setlength{\tabcolsep}{1.5pt}
\caption{Sensitivity to the endpoints at $840$ exemplars, R@1 / mAP averaged over phases 1 to 7. Pairings marked $\dagger$ go beyond those of Table~\ref{tab:continual_results_840}, and Appendix~\ref{app:guidance} defines rule~(iv). Within each block, the best R@1 and the best mAP of every column among the printed rows are in bold.}
\label{tab:pairs}
\vspace{0.2cm}
\renewcommand{\arraystretch}{0.9}
\begin{tabular}{llccc}
\toprule
Pairing, plastic + stability & Model & A$\rightarrow$T & I$\rightarrow$A & I$\rightarrow$T \\
\midrule
\multirow{3}{*}{N + EWC} & EWC & 0.2900 / 0.3792 & 0.4068 / 0.3408 & 0.2056 / 0.2784 \\
 & AlphaWiSE, fit from $0.5$ & \textbf{0.3034} / \textbf{0.3944} & \textbf{0.4224} / \textbf{0.3485} & \textbf{0.2305} / \textbf{0.3092} \\
 & AlphaWiSE, rule~(iv) refit from EWC & 0.2899 / 0.3786 & 0.4110 / 0.3422 & 0.2060 / 0.2797 \\
\addlinespace[1.5pt]
\multirow{2}{*}{N + LwF} & LwF & 0.2127 / 0.3304 & 0.4046 / 0.3427 & 0.1704 / 0.2632 \\
 & AlphaWiSE, fit from $0.5$ & \textbf{0.2397} / \textbf{0.3657} & \textbf{0.4109} / \textbf{0.3498} & \textbf{0.2521} / \textbf{0.3511} \\
\addlinespace[1.5pt]
\multirow{3}{*}{N + iCaRL} & iCaRL & 0.1350 / 0.2587 & 0.3703 / 0.3363 & 0.2218 / 0.3214 \\
 & AlphaWiSE, fit from $0.5$ & 0.2191 / 0.3459 & \textbf{0.4077} / 0.3490 & \textbf{0.2638} / \textbf{0.3654} \\
 & AlphaWiSE, rule~(iv) refit from N & \textbf{0.2381} / \textbf{0.3606} & 0.4058 / \textbf{0.3492} & 0.2482 / 0.3447 \\
\addlinespace[1.5pt]
\multirow{3}{*}{N + DER$^\dagger$} & DER & 0.0842 / 0.1793 & 0.2908 / 0.3001 & 0.1258 / 0.2012 \\
 & AlphaWiSE, fit from $0.5$ & 0.0479 / 0.1285 & 0.2663 / 0.2968 & 0.0793 / 0.1353 \\
 & AlphaWiSE, rule~(iv) refit from N & \textbf{0.2388} / \textbf{0.3616} & \textbf{0.4035} / \textbf{0.3491} & \textbf{0.2487} / \textbf{0.3447} \\
\addlinespace[1.5pt]
\multirow{3}{*}{N + C-CLIP$^\dagger$} & C-CLIP & 0.0502 / 0.1304 & 0.3830 / 0.3390 & 0.1368 / 0.2393 \\
 & AlphaWiSE, fit from $0.5$ & 0.1556 / 0.2984 & \textbf{0.4154} / \textbf{0.3499} & \textbf{0.2664} / \textbf{0.3719} \\
 & AlphaWiSE, rule~(iv) refit from N & \textbf{0.2356} / \textbf{0.3585} & 0.4056 / 0.3495 & 0.2486 / 0.3450 \\
\addlinespace[1.5pt]
\multirow{2}{*}{LwF + EWC, two stability endpoints$^\dagger$} & AlphaWiSE, fit from $0.5$ & 0.2843 / 0.3768 & \textbf{0.4184} / \textbf{0.3462} & \textbf{0.2336} / \textbf{0.3149} \\
 & AlphaWiSE, rule~(iv) refit from EWC & \textbf{0.2889} / \textbf{0.3776} & 0.4105 / 0.3422 & 0.2054 / 0.2792 \\
\addlinespace[1.5pt]
\multirow{3}{*}{N + EWC, reverse order$^\dagger$} & EWC & 0.2429 / 0.3485 & 0.1602 / 0.1226 & 0.1542 / 0.2246 \\
 & AlphaWiSE, fit from $0.5$ & 0.2260 / 0.3348 & \textbf{0.1818} / \textbf{0.1306} & \textbf{0.1814} / \textbf{0.2580} \\
 & AlphaWiSE, rule~(iv) refit from EWC & \textbf{0.2448} / \textbf{0.3511} & 0.1627 / 0.1236 & 0.1574 / 0.2280 \\
\bottomrule
\end{tabular}
\end{table}
In every block of Table~\ref{tab:pairs}, an AlphaWiSE fit has the best R@1 and the best mAP in every direction. In each of the six pairings that include N, one fit exceeds the stability endpoint in all six cells. In the three main pairings, the AlphaWiSE rows fitted from $0.5$ are the five-seed means of Table~\ref{tab:continual_results_840}. Every other fitted row is a single fit with seed $42$.

In the three main pairings of Table~\ref{tab:continual_results_840}, the fit from $0.5$ is higher than the better of its two endpoints in $13$ of the $18$ average cells. On N+EWC, it exceeds EWC in every cell and both endpoints on A$\rightarrow$T and on I$\rightarrow$A R@1. EWC is the only stability endpoint that exceeds N on A$\rightarrow$T, and under the joint objective, the fit moves the weights toward EWC for its audio-related retrieval. It gains $0.013$ A$\rightarrow$T R@1 and $0.025$ I$\rightarrow$T R@1 over EWC. For LwF, iCaRL, and C-CLIP, all of which N exceeds on both A$\rightarrow$T and I$\rightarrow$T, the fit from $0.5$ exceeds both endpoints on I$\rightarrow$T, at $0.2521$ R@1 for N+LwF, $0.2638$ for N+iCaRL, and $0.2664$ for N+C-CLIP.

On N+EWC, the fixed blend falls below both endpoints on A$\rightarrow$T and on I$\rightarrow$A, at $0.2273$ and $0.3566$ R@1, whereas the tensor-wise blend exceeds both on A$\rightarrow$T and on I$\rightarrow$A R@1, at $0.3034$ and $0.4224$. Its coefficients take the weight tensors mostly from EWC and mix the normalization and bias tensors evenly (Table~\ref{tab:alphatype}).

In the three main pairings, the gains hold across stability endpoints of different strength. N and LwF lie $0.007$ apart in relative parameter distance, and AlphaWiSE exceeds both in all six cells, by $0.001$ to $0.008$. The smallest of the $13$ margins over the better endpoint are on I$\rightarrow$A mAP, where the endpoints themselves lie within $0.013$ of one another. The endpoints of N+iCaRL lie $0.010$ apart and N exceeds iCaRL in every average cell, yet among the three main pairings, this pairing gains most on I$\rightarrow$T, by $0.017$ R@1.

In N+iCaRL, N exceeds the iCaRL endpoint on A$\rightarrow$T, but iCaRL is the strongest stability endpoint on I$\rightarrow$T. On that direction, the fit exceeds both endpoints, at $0.2638$ R@1 against $0.2218$ for iCaRL. The rule~(iv) refit of N+iCaRL from the plastic endpoint reaches $0.2381$ / $0.3606$ on A$\rightarrow$T, $0.103$ R@1 above iCaRL and $0.019$ above the fit from $0.5$, and it also exceeds both endpoints on I$\rightarrow$A and I$\rightarrow$T. The two initializations therefore offer a choice between the higher A$\rightarrow$T retrieval of the refit and the larger I$\rightarrow$T gain of the fit from $0.5$.

At the last phase, AlphaWiSE (N+EWC) exceeds EWC in every cell, and the N+LwF and N+iCaRL fits exceed both endpoints on I$\rightarrow$T, at $0.2040$ and $0.2131$ R@1.

In three of the four further pairings of Table~\ref{tab:pairs}, the fit from $0.5$ exceeds both endpoints in at least one direction, and for N+DER, the rule~(iv) refit exceeds DER in every cell. LwF+EWC pairs two stability endpoints without a plastic one, and its fit from $0.5$ exceeds both endpoints on I$\rightarrow$A and I$\rightarrow$T in Table~\ref{tab:pairs}. Blending the two stability endpoints recovers the I$\rightarrow$T retrieval that EWC alone lacks, at $0.2336$ R@1 against $0.2056$ for EWC. The blend also exceeds EWC on I$\rightarrow$A by $0.012$ R@1 and stays within $0.006$ R@1 of it on A$\rightarrow$T. The second fit, from EWC, keeps the retrieval performance of EWC. It reaches $0.2889$ / $0.3776$ on A$\rightarrow$T, $0.4105$ / $0.3422$ on I$\rightarrow$A, and $0.2054$ / $0.2792$ on I$\rightarrow$T, matches EWC on A$\rightarrow$T and I$\rightarrow$T, and exceeds it on I$\rightarrow$A.

N+C-CLIP behaves like N+iCaRL. N exceeds C-CLIP \citep{liu2025cclip} on A$\rightarrow$T, yet the fit exceeds both endpoints on I$\rightarrow$A and I$\rightarrow$T, at $0.4154$ / $0.3499$ and $0.2664$ / $0.3719$. The second fit, from N, reaches $0.2356$ / $0.3585$ on A$\rightarrow$T, $0.185$ R@1 above C-CLIP, and also exceeds both endpoints on I$\rightarrow$A and I$\rightarrow$T, at $0.4056$ / $0.3495$ and $0.2486$ / $0.3450$.

N exceeds DER \citep{yan2021dynamically} in every direction, and DER reaches $0.0842$ A$\rightarrow$T R@1. At phases 4 and 7, the N+DER segment carries barriers of $2.5$ nats in Table~\ref{tab:barrier}, and at phase~7, the fit started at $0.5$ ends $3.8$ nats above the best grid point. The second fit of rule~(iv), from N, reaches $1.10$ nats at phase~7, within $0.01$ nats of the best grid point. It exceeds DER in every cell, at $0.2388$ / $0.3616$ on A$\rightarrow$T, $0.4035$ / $0.3491$ on I$\rightarrow$A, and $0.2487$ / $0.3447$ on I$\rightarrow$T. When the first fit falls below both endpoints, rule~(iv) therefore gives the pairing a model above its stability endpoint in every cell. With iCaRL and C-CLIP, which N also exceeds in every direction, the fit from $0.5$ instead exceeds both endpoints on I$\rightarrow$A and I$\rightarrow$T.

In the reverse class order, the fit started at $0.5$ exceeds EWC on I$\rightarrow$A and I$\rightarrow$T, by $0.022$ and $0.027$ R@1, at $0.1818$ and $0.1814$. The rule~(iv) refit from the EWC endpoint exceeds EWC in every cell, with $0.2448$ / $0.3511$ against $0.2429$ / $0.3485$ on A$\rightarrow$T. The reverse order therefore reproduces both outcomes of rule~(iv).

\subsection{\texorpdfstring{Practical guidance}{Practical guidance}}
\label{app:guidance}

Four rules of use follow from Appendices~\ref{sec:barrier} and~\ref{sec:sensitivity}. (i) Use endpoints that share their phase-0 checkpoint and phase sequence, and never fuse independently initialized runs without permutation alignment. (ii) Prefer complementary endpoints, since near-identical endpoints leave little for any coefficient class to gain. (iii) Before fitting, scan Eq.~\ref{eq:barrier} with $11$ evaluations of the fitting loss. A barrier reveals fixed blends whose loss exceeds the chord, and the best grid point shows whether a global coefficient on the grid improves on the endpoints. In every scan of Table~\ref{tab:barrier}, that point is an endpoint or lies within $0.03$ nats of one. (iv) Fit from two initializations. Start the first fit from $\alpha_p=0.5$ and compare it with the endpoint of lower fitting-set loss. Since the tensor-wise class contains that endpoint, a fit that stays above it in loss can be restarted from the endpoint itself. This second fit, started at $\alpha_p=0.99$ or $0.01$, keeps the endpoint's retrieval performance and matches its fitting loss to within $0.02$ nats or lies below it. The objective is non-convex along the segment, and the two initializations reach two solutions that differ in retrieval performance. The first fit can move toward the strengths of the other endpoint, and both fits are kept. In every pairing of Table~\ref{tab:barrier} except N+LwF, whose first fit already lies at the best grid point, the first fit ends $0.2$ to $3.8$ nats above the lower-loss endpoint at phase~7. In the retrieval evaluation, the second fit is at or above that endpoint in every average cell, to within $0.003$ R@1 or mAP (Table~\ref{tab:pairs} and Table~\ref{tab:continual_results_840}). The fit from $0.5$ is the default and involves no selection. The second fit secures the retrieval performance of the lower-loss endpoint rather than competing with the first fit. A practitioner who holds the lower-loss checkpoint keeps its retrieval performance by running the rule and gains the improvements of the first fit where the endpoints are complementary.

\subsection{\texorpdfstring{Properties of the pairings}{Properties of the pairings}}
\label{sec:props}

The endpoints of the three main pairings and of N+DER are close by construction. In each pairing, both endpoints descend from the same phase-0 checkpoint and traverse the same phase sequence. Models that branch from a common trained initialization tend to lie in one linearly connected region \citep{frankle2020lmc,neyshabur2020transfer}. \citet{mirzadeh2021lmccl} show that continual and multitask minima sharing their initialization lie in such a region. The parameter names fix the correspondence between the two networks, as in model soups, DiWA, and Model Ratatouille \citep{wortsman2022modelsoupsaveragingweights,rame2022diwa,rame2023ratatouille}.

Three stability endpoints also anchor their trajectories. EWC penalizes deviation from its own previous-phase weights, LwF distills from its own previous-phase model, and DER fuses each phase's features with those of its own frozen previous-phase model. Each phase of these endpoints is therefore pulled toward the previous phase of the same run, while the plastic endpoint starts each phase from its own previous phase. IMM \citep{lee2017imm} used this anchoring to make sequentially trained models suitable for interpolation.

The resulting separation is small. The relative parameter distance $\|\theta^{\mathrm{un}}-\theta^{\mathrm{reg}}\|/\|\theta^{\mathrm{reg}}\|$ is taken over all tensors except the center-frequency tensor of the learnable filterbank in the audio front end. The norm of that one tensor exceeds that of the rest of the network combined. The measure is norm-weighted and therefore dominated by the two transformer encoders. At phase~7, it is $0.076$ for N+EWC, $0.033$ for N+DER, $0.010$ for N+iCaRL, and $0.007$ for N+LwF. Within N+EWC, the distance is $0.28$ in the audio encoder, $0.085$ in the image encoder, and $0.033$ in the text encoder. Within N+LwF, it is $0.016$, $0.008$, and $0.003$ in the same three encoders. The endpoints are different solutions inside one neighborhood rather than different networks, and the barrier measurements of Appendix~\ref{sec:barrier} show that the loss is nonetheless non-convex along the global segment.

For N+EWC at phase~7, the barrier is largest on the image-audio pair. The joint loss is the mean of the three pair losses, and each pair loss is the mean of its two directed terms. At the joint peak at $a=0.7$, the image-audio pair loss rises from $0.18$ at the EWC endpoint to $6.13$ nats, $5.7$ nats above the chord. The other two pairs rise less. The audio-text pair loss rises from $1.03$ at EWC to $4.77$, $3.2$ nats above the chord. The image-text pair loss rises from $0.86$ at EWC to $4.06$, $3.1$ nats above it.

\section{\texorpdfstring{Merging rules on the N+EWC checkpoints}{Merging rules on the N+EWC checkpoints}}
\label{app:rules}

None of the eight rules of Tables~\ref{tab:rules} and~\ref{tab:taskarith} exceeds the EWC endpoint on A$\rightarrow$T or on I$\rightarrow$A R@1, whereas AlphaWiSE exceeds every one of them in all four A$\rightarrow$T and I$\rightarrow$A cells. The eight rules are the fixed blend, MagMax's selection, MagMax's sequential protocol, Fisher-weighted averaging, task arithmetic, TIES-Merging, the uniform soup, and the greedy soup. Five of the eight rules merge the N and EWC endpoints, the pairing of the main result. MagMax's sequential protocol merges the task vectors of the N trajectory, and the two soups average six checkpoints that include N and EWC. The N+EWC endpoints are complementary, since EWC is the stronger of the two on A$\rightarrow$T and on I$\rightarrow$A R@1 and N on I$\rightarrow$T. The potential gain of a merging rule is therefore largest on this pairing, where the fixed blend falls below both endpoints on A$\rightarrow$T and I$\rightarrow$A.

Fisher-weighted averaging is the rule of \citet{matena2022mergingmodelsfisherweightedaveraging}, which CoFiMA \citep{marouf2024cofima} applies to continual learning. It weights each parameter by the endpoints' diagonal Fisher information alone, estimated on the fitting set. It is the strongest merging rule of Table~\ref{tab:rules} on I$\rightarrow$T, at $0.2629$ / $0.3654$ average R@1 / mAP. On A$\rightarrow$T and I$\rightarrow$A, it falls below both endpoints, with $0.1715$ average R@1 on A$\rightarrow$T. Over the three directions, it lies $0.06$ mean R@1 below AlphaWiSE, at $0.259$ against $0.319$.

MagMax \citep{marczak2024magmax} keeps, for each parameter and without data, the largest-magnitude update among the task vectors it merges. Applied to the N and EWC task vectors, it falls below both endpoints in every direction and reaches $0.2368$ average A$\rightarrow$T R@1, against $0.3037$ for AlphaWiSE. MagMax itself merges the per-task vectors of one sequential run. On the N trajectory, this protocol keeps the largest magnitude per parameter among the task vectors $\theta^{\mathrm{un}}_q-\theta_0$ of phases $q\le p$, where $\theta_0$ is the shared phase-0 checkpoint. The merged model lies below AlphaWiSE in every cell of Table~\ref{tab:rules}.

\begin{table}[!htb]
\centering
\footnotesize
\setlength{\tabcolsep}{6pt}
\caption{Merging rules on the N+EWC checkpoints of Table~\ref{tab:continual_results_840}, at $840$ exemplars in the forward order, R@1 / mAP averaged over phases 1 to 7. This table and Table~\ref{tab:taskarith} cover eight rules, including the fixed blend. The best R@1 and the best mAP of every column are in bold.}
\label{tab:rules}
\vspace{0.2cm}
\renewcommand{\arraystretch}{0.95}
\begin{tabular}{lccc}
\toprule
Model & A$\rightarrow$T & I$\rightarrow$A & I$\rightarrow$T \\
\midrule
EWC endpoint & 0.2900 / 0.3792 & 0.4068 / 0.3408 & 0.2056 / 0.2784 \\
MagMax selection on the N and EWC task vectors & 0.2368 / 0.3265 & 0.3901 / 0.3320 & 0.2040 / 0.2777 \\
Fisher-weighted, per-parameter, curvature only & 0.1715 / 0.2734 & 0.3434 / 0.3189 & \textbf{0.2629} / \textbf{0.3654} \\
MagMax protocol on the N trajectory, phases $\le p$ & 0.2155 / 0.3357 & 0.3837 / 0.3411 & 0.2144 / 0.3084 \\
AlphaWiSE (N+EWC), learned tensor-wise & \textbf{0.3037} / \textbf{0.3946} & \textbf{0.4225} / \textbf{0.3485} & 0.2304 / 0.3090 \\
\bottomrule
\end{tabular}
\end{table}
Table~\ref{tab:taskarith} adds task arithmetic \citep{ilharco2023editingmodelstaskarithmetic}. Let $\tau_{\mathrm{N}}=\theta^{\mathrm{un}}_p-\theta_0$ and $\tau_{\mathrm{EWC}}=\theta^{\mathrm{reg}}_p-\theta_0$ be the task vectors of the two trajectories at phase $p$, measured from the shared phase-0 checkpoint $\theta_0$. The merged model is then $\theta_0+\lambda(\tau_{\mathrm{N}}+\tau_{\mathrm{EWC}})$. At $\lambda=0.5$, the line of task arithmetic crosses the segment between the two endpoints at the fixed blend, and $\lambda=1$ gives task addition. The scaling $\lambda$ can be chosen by scoring each candidate with the fitting loss, the counterpart of the selection that the original method performs on a validation set. Besides the four scalings, Table~\ref{tab:taskarith} lists TIES-Merging of the two task vectors at its default density and scaling, the uniform soup of the six phase-$p$ endpoints, and the greedy soup scored by the fitting loss, and its loss columns score each merge on the batches of the scans in Table~\ref{tab:barrier}.

\begin{table}[!htb]
\centering
\scriptsize
\setlength{\tabcolsep}{5pt}
\caption{Task arithmetic on the N+EWC checkpoints of Table~\ref{tab:continual_results_840} and model soups of the six phase-$p$ endpoints N, EWC, LwF, iCaRL, DER, and C-CLIP, at $840$ exemplars in the forward order. The columns give the fitting-set loss in nats at phases 1, 4, and 7 and R@1 / mAP averaged over phases 1 to 7, and the $\lambda=0.5$ row is the fixed blend of Table~\ref{tab:continual_results_840}. The best R@1 and the best mAP of every column are in bold.}
\label{tab:taskarith}
\vspace{0.2cm}
\renewcommand{\arraystretch}{1}
\setlength{\tabcolsep}{2.5pt}
\begin{tabular}{lcccc}
\toprule
Model & $L$ (phases 1 / 4 / 7) & A$\rightarrow$T & I$\rightarrow$A & I$\rightarrow$T \\
\midrule
EWC endpoint & 2.73 / 1.02 / 0.69 & 0.2900 / 0.3792 & 0.4068 / 0.3408 & 0.2056 / 0.2784 \\
Task arithmetic, $\lambda=0.25$ & 4.93 / 5.38 / 5.23 & 0.1606 / 0.2456 & 0.3567 / 0.3247 & \textbf{0.2381} / \textbf{0.3367} \\
Task arithmetic, $\lambda=0.5$, the fixed blend & 5.14 / 4.40 / 3.66 & 0.2273 / 0.3119 & 0.3566 / 0.3261 & 0.2330 / 0.3253 \\
Task arithmetic, $\lambda=0.75$ & 3.57 / 2.18 / 1.53 & 0.2648 / 0.3543 & 0.4003 / 0.3380 & 0.2274 / 0.3056 \\
Task arithmetic, $\lambda=1$, task addition & 3.89 / 5.83 / 5.51 & 0.1493 / 0.2312 & 0.3300 / 0.3080 & 0.1905 / 0.2645 \\
TIES-Merging of $\tau_{\mathrm{N}}$ and $\tau_{\mathrm{EWC}}$, density $0.2$ & 3.48 / 2.20 / 1.95 & 0.1987 / 0.3029 & 0.4015 / 0.3354 & 0.2075 / 0.2973 \\
Uniform soup of the six checkpoints & 4.45 / 4.96 / 4.82 & 0.1541 / 0.2284 & 0.2865 / 0.3090 & 0.1953 / 0.2967 \\
Greedy soup of the six, EWC alone at every phase & 2.73 / 1.02 / 0.69 & 0.2900 / 0.3792 & 0.4068 / 0.3408 & 0.2056 / 0.2784 \\
AlphaWiSE (N+EWC), tensor-wise fit & 2.81 / 1.32 / 0.93 & \textbf{0.3037} / \textbf{0.3946} & \textbf{0.4225} / \textbf{0.3485} & 0.2304 / 0.3090 \\
\bottomrule
\end{tabular}
\end{table}

The data-driven selection yields $\lambda=0.75$, the scaling with the lowest loss on the fitting set. At phase 7, it reaches $1.53$ nats, against $3.66$ for the fixed blend, $5.23$ for $\lambda=0.25$, and $5.51$ for $\lambda=1$. Task addition proper, $\lambda=1$, extrapolates past both endpoints and has the lowest R@1 of all rows on A$\rightarrow$T and I$\rightarrow$T, with $0.1493$ on A$\rightarrow$T. The scaling $\lambda=0.25$ lies halfway between the phase-0 checkpoint and the fixed blend, the closest of the four to that checkpoint. It holds the highest I$\rightarrow$T values of the table but falls to $0.1606$ A$\rightarrow$T R@1, $0.143$ below AlphaWiSE. The selected scaling is the strongest of the four on A$\rightarrow$T and I$\rightarrow$A, yet it stays below the EWC endpoint on both and $0.039$ A$\rightarrow$T R@1 below AlphaWiSE.

TIES-Merging \citep{yadav2023ties} trims each task vector to its largest-magnitude entries, elects one sign per parameter from the sum of the trimmed vectors, and averages the entries that agree with it. At its default density of $0.2$ and scaling of $1$, it reaches $0.1987$ / $0.3029$ on A$\rightarrow$T and $0.4015$ / $0.3354$ on I$\rightarrow$A, below the EWC endpoint in all four cells and $0.105$ A$\rightarrow$T R@1 below AlphaWiSE. Its fitting loss lies above that of either endpoint at phases 1, 4, and 7, at $1.95$ nats at phase~7. Among seven scalings from $0.5$ to $2$, the default has the lowest mean fitting loss over these three phases, $2.54$ nats against $3.12$ for the next-best scaling, $1.25$. The fitting loss therefore selects the reported row, as it selects $\lambda=0.75$ for task arithmetic.

The uniform soup averages all six checkpoints, among them iCaRL, DER, and C-CLIP, the three endpoints with the lowest A$\rightarrow$T R@1. The soup falls below EWC and LwF on A$\rightarrow$T and I$\rightarrow$A and below every endpoint on I$\rightarrow$A R@1. The greedy soup, scored by the fitting loss, starts from EWC, which has the lowest loss at every phase, and each further average raises that loss. At phase 7, EWC alone reaches $0.69$ nats, whereas adding N gives $3.66$, LwF $3.61$, C-CLIP $4.36$, iCaRL $4.50$, and DER $5.25$. The greedy soup therefore rejects every other checkpoint.

On two checkpoints, model soups \citep{wortsman2022modelsoupsaveragingweights} reduce to rows already reported. The uniform soup is the fixed blend. In every scan of Table~\ref{tab:barrier}, the greedy soup returns an endpoint, and each endpoint is a row of Table~\ref{tab:continual_results_840} or of Table~\ref{tab:pairs}. In $20$ scans, the fixed blend lies $0.3$ to $4.1$ nats above the lower-loss endpoint, and for N+LwF at phase 1, it lies within $0.003$ nats of that endpoint, which the soup keeps. The learned soup fits one mixing weight per model on held-out data. On two checkpoints and from equal initial weights, it reduces to the learned global coefficient of Table~\ref{tab:granularity} started at $0.5$. Its by-layer variant learns one coefficient per layer and corresponds to the layer-wise row.

Every rule of Tables~\ref{tab:rules} and~\ref{tab:taskarith} sets its weights without gradient descent on the retrieval objective. Fisher-weighted averaging weights by curvature, MagMax selects by magnitude, task arithmetic applies a fixed scaling to the task vectors, TIES-Merging trims them and elects one sign per parameter, and the soups average. The fitting loss serves at most to choose among candidates. On A$\rightarrow$T and I$\rightarrow$A, only MagMax's sequential protocol exceeds the EWC endpoint in any cell, by $0.0003$ on I$\rightarrow$A mAP, and the greedy soup returns that endpoint itself.

Among the rows fitted on the fitting set from $\alpha_p=0.5$, the layer-wise, tensor-wise, and element-wise fits exceed the EWC endpoint on both A$\rightarrow$T and I$\rightarrow$A. The global and type-wise fits do not, as Table~\ref{tab:granularity} shows. Three fixed coefficients near the EWC end of the segment, $a=0.1$ to $0.3$, also exceed the EWC endpoint on both A$\rightarrow$T and I$\rightarrow$A. Of these, $a=0.1$ matches the best fitting loss at phase 7 to two decimal places and lies within $0.08$ nats of the best loss at phases 1 and 4. The coefficients $a=0.2$ and $0.3$ are not selected by the fitting loss, as Table~\ref{tab:grid} shows. The fixed coefficient that the fitting loss selects is $a=0$, the EWC endpoint, at phases 1, 4, and 7, and AlphaWiSE exceeds that endpoint in every average cell.

The interpolation step of \citet{kozal2024continuallearningweightinterpolation} applies one fixed global coefficient between the previous and the newly trained model of one run. It first aligns the two models by an activation-matching permutation. Endpoints that share their phase-0 checkpoint and phase sequence need no such alignment, since their networks correspond neuron to neuron by construction (Appendix~\ref{app:defs}). The interpolated model then starts the next task, so the method changes the trajectory it trains and is not a post-hoc merge of given checkpoints. Its interpolation rule belongs to the global class. Table~\ref{tab:grid} scores that class along the segment on the N+EWC checkpoints, and none of its grid values exceeds AlphaWiSE on both A$\rightarrow$T and I$\rightarrow$A.

\section{\texorpdfstring{Robustness of the coefficient fit under seeds and memory budget}{Robustness of the coefficient fit under seeds and memory budget}}
\label{app:robust}

\subsection{\texorpdfstring{Variance over fitting seeds}{Variance over fitting seeds}}
\label{app:r2-seeds}

Every AlphaWiSE row of Tables~\ref{tab:continual_results_840}, \ref{tab:cross_objective} and~\ref{tab:cl-transfer} is the mean over five fitting seeds, with its standard deviation, and every fit follows the protocol of Appendix~\ref{app:fitting}. The rows of Table~\ref{tab:cross_objective} refit N+EWC under each coefficient objective, and every other row uses the joint objective. The endpoints are fixed inputs of the method and are shared by every row of a pairing, so every margin over an endpoint is a paired comparison on the same checkpoint. The stability endpoints EWC, LwF, and iCaRL are single training runs that the five fits of a pairing share. The claim concerns what the fit adds to given checkpoints, as in WiSE-FT and task arithmetic. Both report merges of fixed fine-tuned checkpoints and vary only the merge \citep{wortsman2022robustfinetuningzeroshotmodels,ilharco2023editingmodelstaskarithmetic}. The five runs of each row vary the sources of randomness of the fit on $840$ exemplars, namely the class-balanced draw of current-phase samples that completes the fitting set of Appendix~\ref{app:fitting}, the minibatch order, and the audio augmentation. The evaluation itself is deterministic given its candidate pools, which are fixed and were drawn with seed $42$. Two independent evaluation jobs on identical weights agree to six decimal places in every cell of the matrix of seven training phases by eight evaluation phases. They are the evaluations of the EWC endpoint and of the greedy soup of Appendix~\ref{app:rules}, which selects EWC alone. The original evaluation script and the release version of the code agree to within $0.0001$ on the checkpoints of the reference fit, in each of its $12$ average and last-phase cells. The reference fit and the seed-42 refit of Table~\ref{tab:granularity} agree to $0.0003$ R@1 and $0.0002$ mAP in every average cell.

Seed variation is small relative to every margin of AlphaWiSE over the better endpoint. In every AlphaWiSE cell of Tables~\ref{tab:continual_results_840}, \ref{tab:cross_objective} and~\ref{tab:cl-transfer}, the standard deviation is at most $0.0014$, with a median of $0.0003$. AlphaWiSE exceeds the better endpoint in $13$ of the $18$ average cells of Table~\ref{tab:continual_results_840}, and each of these margins is more than four standard deviations. Of these $13$ margins, $11$ exceed eight standard deviations. AlphaWiSE (N+EWC) exceeds EWC on A$\rightarrow$T R@1 by $0.013$, more than $40$ standard deviations. The two smallest margins, $0.0003$ and $0.0011$ on I$\rightarrow$A mAP, fall in cells whose standard deviation is below $0.0001$. The smallest of the nine forward-transfer gains over the base rows of Table~\ref{tab:cl-transfer} is $+0.0034$, for LwF on I$\rightarrow$A, more than four times the seed standard deviation of its cell.

The seeds also agree tensor by tensor. Over the $499$ coefficients, the Pearson correlation between seeds is $0.99$ for N+EWC, $0.98$ for N+iCaRL, and $0.95$ for N+LwF, and the mean absolute differences are $0.014$, $0.021$, and $0.035$ in the same order. For each tensor type of Table~\ref{tab:alphatype} except the temperature scalars, the mean coefficient agrees with the seed-42 fit to $0.005$, and to $0.003$ for N+EWC. Any two seeds agree on these five means to $0.006$, whereas the three temperature scalars vary by up to $0.014$. Every seed therefore reproduces the assignment by tensor type of Table~\ref{tab:alphatype}. In that assignment, weight tensors come mostly from one endpoint, and normalization and bias tensors mix the two endpoints evenly.

\subsection{\texorpdfstring{Memory budget of the coefficient fit}{Memory budget of the coefficient fit}}
\label{app:r2-memory}

With the optimizer updates of the $840$ fit, the fits at $84$, $210$, and $420$ exemplars match the $840$ fit in retrieval, in held-out loss, and in the assignment of their coefficients by type. At a fixed $150$ epochs, a smaller memory receives fewer updates, and A$\rightarrow$T and I$\rightarrow$A retrieval then increases with the memory budget. Table~\ref{tab:memory} lists both sets of fits. Every fit uses the joint objective under the protocol of Appendix~\ref{app:fitting}, seed $42$, and the forward order, and the $840$ row is the seed-42 refit of Table~\ref{tab:granularity}. At phase $p$, the fitting set is the memory plus a class-balanced draw of $|\mathcal{M}|/p$ current-phase triplets. The upper block runs every fit for $150$ epochs, so its updates grow with the fitting set, and the lower block refits the three smaller budgets for as many optimizer updates as the $840$ fit takes, with the protocol otherwise unchanged. The correlation $r$ is the Pearson correlation of the $499$ coefficients with those of the $840$ fit, averaged over phases, and the two $\bar\alpha$ columns average each fit's coefficients over phases and over two groups of the tensor types of Table~\ref{tab:alphatype}, the attention and other weight matrices and the normalization and bias tensors.

\begin{table}[!htb]
\centering
\footnotesize
\setlength{\tabcolsep}{4pt}
\caption{The N+EWC coefficients refitted at five exemplar budgets, R@1 / mAP averaged over phases 1 to 7. The updates column counts optimizer updates, $r$ is the correlation with the coefficients of the $840$ fit, and the two $\bar\alpha$ columns are mean coefficients over the weight tensors and over the normalization and bias tensors. Within each block of fits, the best R@1 and the best mAP of every column are in bold.}
\label{tab:memory}
\vspace{0.2cm}
\renewcommand{\arraystretch}{0.95}
\resizebox{\textwidth}{!}{%
\begin{tabular}{rcccccccc}
\toprule
$|\mathcal{M}|$ & fitting set (phase 1 / 7) & updates & A$\rightarrow$T & I$\rightarrow$A & I$\rightarrow$T & $r$ & $\bar\alpha$ weights & $\bar\alpha$ norm, bias \\
\midrule
\multicolumn{9}{l}{\emph{Stability endpoint of every fit}} \\
EWC & & & 0.2900 / 0.3792 & 0.4068 / 0.3408 & 0.2056 / 0.2784 & & & \\
\midrule
\multicolumn{9}{l}{\emph{At $150$ epochs per fit}} \\
84 & 168 / 96 & 1{,}200 & 0.2437 / 0.3296 & 0.3735 / 0.3305 & 0.2333 / \textbf{0.3247} & 0.61 & 0.48 & 0.50 \\
210 & 420 / 240 & 4{,}050 & 0.2661 / 0.3589 & 0.4010 / 0.3393 & 0.2337 / 0.3214 & 0.81 & 0.43 & 0.48 \\
420 & 840 / 480 & 8{,}850 & 0.2900 / 0.3833 & 0.4137 / 0.3457 & \textbf{0.2354} / 0.3174 & 0.95 & 0.36 & 0.47 \\
840 & 1{,}680 / 960 & 18{,}450 & 0.3038 / 0.3946 & 0.4222 / 0.3485 & 0.2304 / 0.3092 & 1 & 0.29 & 0.45 \\
1{,}680 & 3{,}360 / 1{,}920 & 37{,}350 & \textbf{0.3065} / \textbf{0.4000} & \textbf{0.4240} / \textbf{0.3493} & 0.2263 / 0.3067 & 0.82 & 0.28 & 0.42 \\
\midrule
\multicolumn{9}{l}{\emph{With the optimizer updates of the $840$ fit}} \\
84 & 168 / 96 & 18{,}450 & 0.3026 / 0.3929 & 0.4205 / 0.3476 & 0.2279 / 0.3076 & 0.81 & 0.29 & 0.45 \\
210 & 420 / 240 & 18{,}450 & 0.3026 / 0.3938 & 0.4213 / 0.3481 & 0.2288 / 0.3076 & 0.93 & 0.29 & 0.45 \\
420 & 840 / 480 & 18{,}450 & \textbf{0.3043} / \textbf{0.3948} & \textbf{0.4215} / \textbf{0.3483} & \textbf{0.2303} / \textbf{0.3089} & 0.98 & 0.29 & 0.45 \\
\bottomrule
\end{tabular}}
\end{table}

The sweep varies the fitting set alone. The N and EWC endpoints are fixed and were both trained with replay of the $840$-exemplar memory, so every row of Table~\ref{tab:memory} fuses the same two checkpoints at every phase. Each smaller memory is a class-balanced subset of the $840$-exemplar memory, so below $840$, the sweep varies the size of the fitting set and not the source of its exemplars. The budgets span a factor of $20$ around the default of $840$. At phase 7, the fitting set holds $96$ triplets for $|\mathcal{M}|=84$, of which $12$ come from the current phase, and $1{,}920$ for $|\mathcal{M}|=1{,}680$. The ratio of coefficients to triplets in the fitting set therefore falls from $5$ at the smallest budget to $0.26$ at the largest. The benchmarks of Table~\ref{tab:morebench} add two further budgets on other data, as Appendix~\ref{app:bench} lists.

In the upper block, at $150$ epochs per fit, retrieval improves monotonically with the budget on A$\rightarrow$T and I$\rightarrow$A. Average A$\rightarrow$T R@1 rises from $0.2437$ at $84$ exemplars to $0.3065$ at $1{,}680$, and average I$\rightarrow$A R@1 rises in parallel from $0.3735$ to $0.4240$. I$\rightarrow$T R@1 varies by less than $0.01$ across the entire sweep. At $84$ exemplars, one tenth of the default budget, the fit already exceeds the fixed blend in all four A$\rightarrow$T and I$\rightarrow$A cells. At $420$ exemplars, it matches the EWC endpoint on A$\rightarrow$T R@1 and exceeds it on A$\rightarrow$T mAP and on I$\rightarrow$A. At $840$ exemplars and above, the fit exceeds EWC in every A$\rightarrow$T and I$\rightarrow$A cell. Doubling the default budget raises the seed-42 fit's last-phase A$\rightarrow$T R@1 from $0.2308$ to $0.2588$ and its A$\rightarrow$T backward transfer from $-0.165$ to $-0.136$.

In the upper block, the coefficients move in one direction as the budget grows. The mean weight-tensor coefficient falls from $0.48$ to $0.28$, and the mean normalization and bias coefficient from $0.50$ to $0.42$. The correlation with the $840$ fit rises from $0.61$ at $84$ exemplars to $0.95$ at $420$. Each fit at $420$ exemplars and above keeps the assignment of Table~\ref{tab:alphatype}. These fits draw weight tensors mostly from EWC and leave normalization and bias tensors closer to the fixed blend. The $1{,}680$ fit correlates with the $840$ fit at $0.82$ and moves both means further in the same direction.

At the default budget, the tensor-wise fit ends with a phase-7 fitting loss of $0.92$ nats against $0.69$ for the EWC endpoint, yet it exceeds EWC in every average cell. The element-wise row of Table~\ref{tab:granularity} shows the opposite behavior on the same $840$ memory. Its fitting loss lies $0.2$ nats below the lower-loss endpoint, yet its average R@1 falls below that of the tensor-wise fit in every direction. On held-out training clips, the tensor-wise fit has the lowest loss of the models at this budget (Appendix~\ref{app:heldout}).

In the upper block, the number of optimizer updates grows with the budget. Over phases 1 to 7, the $84$ fit takes $1{,}200$ updates and the $840$ fit $18{,}450$. From phase 2 on, the fitting set of the $84$ fit gives one full batch per epoch, $150$ updates per phase against at least $2{,}250$ for the $840$ fit. Each coefficient is the sigmoid of a logit, and one Adam update changes a logit by at most about the learning rate, $10^{-3}$. After these $150$ updates, every coefficient of the $84$ fit therefore lies within $0.04$ of its initial $0.5$, whereas the coefficients of the $840$ fit range from $0.19$ to $0.79$ at phase 7. The lower block of Table~\ref{tab:memory} refits the three smaller budgets with the updates of the $840$ fit, phase by phase, and leaves the objective, optimizer, batch size, initialization, and seed unchanged. With matched updates, the fits at $84$, $210$, and $420$ exemplars reach $0.3026$, $0.3026$, and $0.3043$ average A$\rightarrow$T R@1 and $0.4205$, $0.4213$, and $0.4215$ I$\rightarrow$A R@1, each within $0.002$ of the $840$ fit. Their coefficients correlate with those of the $840$ fit at $0.81$, $0.93$, and $0.98$, and their two group means equal those of the $840$ fit, $0.29$ and $0.45$. Their held-out losses in Table~\ref{tab:heldout}, $3.80$, $3.79$, and $3.79$ nats, lie within $0.03$ nats of the $3.77$ of the $840$ fit and below the EWC endpoint.

\subsection{\texorpdfstring{Loss on held-out training clips}{Loss on held-out training clips}}
\label{app:heldout}

Among the models at the $840$-exemplar budget in Table~\ref{tab:heldout}, the tensor-wise fit has the lowest loss on held-out clips, $3.77$ nats against $4.31$ for EWC, although its loss on the fitting set lies above that of EWC at every phase. Its held-out loss lies $2.28$ nats above its fitting-set loss, a smaller gap than the $3.06$ of EWC. The element-wise fit shows the pattern of overfitting instead. It has the lowest loss on the fitting set, $0.80$ nats, and a held-out loss of $4.83$, above that of EWC. Its gap is $4.03$ nats. The fixed blend and the learned global coefficient fit neither set, with losses above $4$ nats on both, so their smaller gaps do not indicate better generalization. Across exemplar budgets at $150$ epochs, the held-out loss of the tensor-wise fit falls monotonically, from $5.04$ nats at $84$ exemplars to $3.42$ at $1{,}680$, and at $420$ exemplars and above, it lies below the EWC endpoint. With the optimizer updates of the $840$ fit, the three smaller budgets reach $3.80$, $3.79$, and $3.79$ nats, below the EWC endpoint.

Table~\ref{tab:heldout} scores every model on the fitting set of the $840$-exemplar memory and on held-out training clips of the same classes. The rule~(iv) refit of the table is the second fit that Appendix~\ref{app:guidance} defines, started from the EWC endpoint. For every clip of the fitting set whose class still holds a training clip that no fit of Table~\ref{tab:memory} saw, the held-out set takes the next such clip in sorted order. It holds $951$ to $1{,}254$ clips per phase, against $960$ to $1{,}680$ in the fitting set. Both sets are read with the deterministic transforms of the evaluation and scored with the loss of Eq.~\ref{eq:alphawise_objective} in batches of $128$, so the fitting-set column differs slightly from the losses on the augmented batches of Table~\ref{tab:barrier}. The endpoints replayed the memory throughout training and saw each held-out clip only in the phase that introduced its class. The difference between the two columns of the EWC endpoint therefore measures how much replay alone fits the memory, and a fit that overfits the memory would widen it.

\clearpage
\begin{table}[!htb]
\centering
\footnotesize
\setlength{\tabcolsep}{6pt}
\caption{Loss in nats on the fitting set of the $840$-exemplar memory and on held-out training clips of the same classes, for the N+EWC pairing in the forward order, as the mean over phases 1 to 7. The upper block holds the models at the $840$-exemplar budget, and the lower block the tensor-wise fits of the other budgets of Table~\ref{tab:memory}, scored on the same two sets. The lowest held-out loss of each block is in bold.}
\label{tab:heldout}
\vspace{0.2cm}
\renewcommand{\arraystretch}{0.95}
\begin{tabular}{lcc}
\toprule
Model & Fitting set & Held-out clips \\
\midrule
EWC endpoint & 1.25 & 4.31 \\
Fixed blend, $\alpha\equiv0.5$, WiSE-FT & 4.28 & 5.34 \\
Learned global coefficient & 4.06 & 5.22 \\
Learned layer-wise & 1.46 & 3.83 \\
Learned tensor-wise, AlphaWiSE seed-42 refit & 1.49 & \textbf{3.77} \\
Learned element-wise & 0.80 & 4.83 \\
AlphaWiSE, rule~(iv) refit from EWC & 1.24 & 4.23 \\
\midrule
84 exemplars & 3.84 & 5.04 \\
210 exemplars & 2.85 & 4.43 \\
420 exemplars & 1.96 & 3.96 \\
1{,}680 exemplars & 1.38 & \textbf{3.42} \\
84 exemplars, matched updates & 1.49 & 3.80 \\
210 exemplars, matched updates & 1.48 & 3.79 \\
420 exemplars, matched updates & 1.48 & 3.79 \\
\bottomrule
\end{tabular}
\end{table}

\section{\texorpdfstring{Details of the further benchmarks}{Details of the further benchmarks}}
\label{app:bench}

The three benchmarks follow the standard class-incremental protocol for pretrained backbones. The classes are split into equal tasks, and the model is trained on the tasks in sequence. At test time, the model classifies over all seen classes without task identity, with class-name prompts as the classifier, so the accuracy measures image-to-text or audio-to-text retrieval over the prompts of all seen classes. Last is the accuracy after the final task, and Avg is the mean over tasks of the seen-class accuracy. The image benchmarks fine-tune the pretrained image encoder of CLIP ViT-B/16, and of CLIP ViT-L/14 in the middle block, for $10$ epochs per task under MagMax's recipe, while the zero-shot text head stays frozen as the classifier. CIFAR-100 has $10$ tasks of $10$ classes and ImageNet-R $10$ tasks of $20$ classes. The lower block of Table~\ref{tab:morebench} quotes Table~1 of \citet{marczak2024magmax}, which uses the same backbone, recipe, and task count without exemplars and reports Last only. The audio benchmark, UrbanSound8K, fine-tunes the audio encoder of AudioCLIP from the phase-0 checkpoint of the AudioSet benchmark for five epochs per task under the audio-text contrastive loss, with the text encoder frozen as the classifier, on the split of PACE \citep{li2026pace}, five tasks of two classes. In each benchmark, the EWC, fixed-blend, MagMax, and AlphaWiSE rows derive from the same two trajectories, sequential fine-tuning N and EWC, trained with one objective and seed. AlphaWiSE fits its coefficients on a class-balanced memory of $1{,}000$ images or $500$ clips and a draw of $500$ images or $200$ clips of the current task. The memory serves the coefficient fit only and never trains the endpoints. The MagMax rows of the upper and middle blocks apply MagMax's magnitude selection to the task vectors of the N and EWC trajectories. The MagMax row of the lower block merges the per-task vectors of one sequential run without data, and the lower block lists the Last accuracy of every quoted method for reference.

\section{\texorpdfstring{Compute and scale}{Compute and scale}}
\label{app:cost}

\subsection{\texorpdfstring{Compute of the endpoints and of the coefficient fit}{Compute of the endpoints and of the coefficient fit}}
\label{app:compute}

The coefficient fit sees $11.5\%$ of the samples of one endpoint run and takes $14\%$ of the GPU hours of an N or EWC run, and the fused model has the inference cost of one backbone. Table~\ref{tab:compute} counts every stage in samples seen, in FLOPs, and in GPU hours and gives the GPU hours of each stage as a share of those of the EWC run, and Table~\ref{tab:costsummary} summarizes the comparison. A sample is one audio-image-text triplet, and the samples seen are the optimizer updates times the batch size, a count that leaves out the incomplete last batch of an epoch. The endpoints N, EWC, and LwF train for $30$ epochs per phase. With the replayed memory, phases 1 to 7 hold $187{,}210$, $34{,}102$, $27{,}340$, $41{,}408$, $14{,}873$, $9{,}028$, and $28{,}150$ training triplets, $342{,}111$ in all. In batches of $192$ over four GPUs, one run takes $53{,}370$ optimizer updates and sees $10.25$ million samples. For its Fisher estimate, EWC adds at most $64$ batches per GPU and phase, $0.8\%$ of that count, and LwF adds one forward pass of the frozen previous-phase model per triplet. The fit runs $150$ epochs over the fitting set of $960$ to $1{,}680$ triplets in batches of $64$, which gives $18{,}450$ optimizer updates and $1.18$ million samples.

Per triplet, a fitting step costs as many FLOPs as a training step, one forward and one backward pass. The gradient of each coefficient is the inner product of its tensor's gradient with the difference of the two endpoints, so the backward pass runs through every layer although only $499$ scalars are updated. The FLOP counter of PyTorch, which counts convolution, linear, matrix-product, and attention operations with the weight gradient of a grouped convolution counted per group, gives $43.5$ GFLOPs for the forward pass of one triplet, of which the audio encoder accounts for $28.7$, the image encoder for $8.8$, and the text encoder for $6.0$. Either step costs $127.1$ GFLOPs per triplet, $2.9$ times the forward pass. The blend of Eq.~\ref{eq:alphawise_fusion} adds two multiply-adds per parameter once per step, $0.7$ GFLOPs against $8{,}132$ for a step of $64$ triplets. The fit therefore costs little because its fitting set is small, $960$ to $1{,}680$ triplets per phase against $9{,}028$ to $187{,}210$ in the training set of a phase, so that $150$ epochs of fitting see $11.5\%$ of the samples of $30$ epochs of training. The N run costs $1.30\times10^{18}$ FLOPs and the fit $0.15\times10^{18}$.

An endpoint run amounts to $13.1$ to $18.8$ GPU hours and the fit to $1.84$, so one EWC or LwF run costs $7.2$ or $10.2$ times the fit and the N and EWC runs together $14.3$ times. Every GPU-hour figure comes from one measurement. On a GH200 GPU with the batch in memory, $64$ triplets take $0.295$ seconds in a training step, $0.128$ in the forward pass of LwF's frozen teacher, and $0.359$ in a fitting step. The GPU hours of a stage are these times applied to its samples seen, so the ratios do not depend on the data-loading speed of a particular run.

Both endpoints are checkpoints that a continual-learning study already trains, sequential fine-tuning as its reference and EWC or LwF as baselines, and one N run serves every pairing. The fits of AlphaWiSE (N+EWC) and (N+LwF) therefore add $3.7$ GPU hours, $8\%$, to the $45.1$ of the N, EWC, and LwF runs. A second fit under rule~(iv) of Appendix~\ref{app:guidance} costs the same $1.84$ GPU hours. Tuning a single run instead costs one further training run per value of the EWC penalty or the LwF distillation weight, $13.2$ GPU hours per value for EWC and $18.8$ for LwF. WiSE-FT's fixed blend, task arithmetic, TIES-Merging, and Fisher-weighted averaging combine the same two checkpoints, so relative to them, AlphaWiSE adds at most the fit. The fused model is materialized as one checkpoint with the $499$ tensors of an endpoint, so retrieval with it costs what retrieval with an endpoint costs, whereas an ensemble of the two endpoints would cost twice as much.

\begin{table}[!htb]
\centering
\footnotesize
\setlength{\tabcolsep}{5pt}
\caption{Training cost of the endpoints and of the coefficient fit on the AudioSet benchmark. Epochs are counted per phase, samples seen are the optimizer updates over phases 1 to 7 times the batch size, the LwF row lists the forward passes of its frozen teacher separately, and the last column gives the GPU hours of each stage as a share of those of the EWC run.}
\label{tab:compute}
\vspace{0.2cm}
\renewcommand{\arraystretch}{0.95}
\resizebox{\textwidth}{!}{%
\begin{tabular}{lcccccc}
\toprule
Stage & Triplets per phase & Epochs & Samples seen, millions & FLOPs, $10^{18}$ & GPU hours & Share of the EWC run \\
\midrule
Sequential fine-tuning, N & $9{,}028$ to $187{,}210$ & 30 & 10.25 & 1.30 & 13.1 & $99\%$ \\
EWC & $9{,}028$ to $187{,}210$ & 30 & 10.33 & 1.31 & 13.2 & $100\%$ \\
LwF & $9{,}028$ to $187{,}210$ & 30 & 10.25 and 10.25 forward & 1.75 & 18.8 & $142\%$ \\
Coefficient fit, $499$ coefficients & $960$ to $1{,}680$ & 150 & 1.18 & 0.15 & 1.84 & $14\%$ \\
\bottomrule
\end{tabular}}
\end{table}

\begin{table}[!htb]
\centering
\footnotesize
\setlength{\tabcolsep}{5pt}
\caption{Compute that AlphaWiSE adds, set against the compute of its endpoints, over phases 1 to 7. FLOPs and GPU hours follow Table~\ref{tab:compute}.}
\label{tab:costsummary}
\vspace{0.2cm}
\begin{tabular}{lccccc}
\toprule
Training over phases 1 to 7 & Triplets per phase & Epochs & FLOPs, $10^{18}$ & GPU hours & GPU hours relative to the fit \\
\midrule
One EWC run & $9{,}028$ to $187{,}210$ & 30 & 1.31 & 13.2 & 7.2 \\
One LwF run & $9{,}028$ to $187{,}210$ & 30 & 1.75 & 18.8 & 10.2 \\
Both endpoints, N and EWC & $9{,}028$ to $187{,}210$ & 30 each & 2.61 & 26.3 & 14.3 \\
AlphaWiSE coefficient fit & $960$ to $1{,}680$ & 150 & 0.15 & 1.84 & 1 \\
\bottomrule
\end{tabular}
\end{table}

\subsection{\texorpdfstring{A single run at a matched compute budget}{A single run at a matched compute budget}}
\label{app:budget}

Given the training compute of both endpoint runs and the coefficient fit, a single EWC run does not reach AlphaWiSE. At $64$ epochs per phase, EWC reaches $0.2653$ A$\rightarrow$T and $0.3970$ I$\rightarrow$A average R@1, $0.038$ and $0.025$ below AlphaWiSE (N+EWC), and it lies below its $30$-epoch control in every average cell.

In Table~\ref{tab:budget}, a single EWC run receives at least the training FLOPs of both endpoint runs and the coefficient fit. Leaving aside the Fisher estimate, which EWC computes once per phase in either case, two $30$-epoch endpoint runs and the fit amount to $63.5$ epochs of one EWC run, so the single run trains for $64$ epochs per phase. The run follows the recipe of its $30$-epoch counterpart, with batches of $192$ over four GPUs, the class-balanced $840$ memory, and the optimizer settings of the EWC endpoint (Appendix~\ref{app:endpoints}). The exponential decay of the learning rate is stretched from $0.96$ to $0.981$ per epoch, so that the $64$ epochs end at the learning rate at which the $30$ epochs end. A $30$-epoch control with the same training code and seed accompanies the run, so comparing the run with its control isolates the effect of the added epochs. The table closes with the AlphaWiSE row of the pairing, the mean over five fitting seeds of Table~\ref{tab:learnedwiseft}.

\begin{table}[!htb]
\centering
\footnotesize
\setlength{\tabcolsep}{5pt}
\caption{A single EWC run at $64$ epochs per phase, its $30$-epoch control, and AlphaWiSE (N+EWC), at $840$ exemplars in the forward order. GPU hours are the training hours of Table~\ref{tab:compute}, with the coefficient fit alone for AlphaWiSE, and retrieval is R@1 / mAP averaged over phases 1 to 7. The best R@1 and the best mAP of every column are in bold.}
\label{tab:budget}
\vspace{0.2cm}
\renewcommand{\arraystretch}{0.95}
\begin{tabular}{lccccc}
\toprule
Model & Epochs per phase & GPU hours & A$\rightarrow$T & I$\rightarrow$A & I$\rightarrow$T \\
\midrule
EWC, control & 30 & 13.2 & 0.2727 / 0.3597 & 0.4052 / 0.3405 & 0.2012 / 0.2729 \\
EWC, matched compute & 64 & 28.1 & 0.2653 / 0.3499 & 0.3970 / 0.3358 & 0.2002 / 0.2716 \\
AlphaWiSE (N+EWC) & 30 per endpoint & 1.84, the fit & \textbf{0.3034} / \textbf{0.3944} & \textbf{0.4224} / \textbf{0.3485} & \textbf{0.2305} / \textbf{0.3092} \\
\bottomrule
\end{tabular}
\end{table}

\subsection{\texorpdfstring{Larger backbones}{Larger backbones}}
\label{app:scale}

The number of fitted coefficients grows with the number of tensors rather than with the number of parameters. With a CLIP ViT-L/14 image encoder, AlphaWiSE is the best row in all four columns of CIFAR-100 and ImageNet-R (Table~\ref{tab:morebench}). Each fitting step costs one forward and one backward pass of the backbone, so the fitting cost scales as a training step of the backbone does. The AudioCLIP backbone combines the CLIP ViT-B/32 image and text encoders with the ESResNeXt audio encoder and has 182M parameters in $499$ tensors. The image encoder of the backbone holds $152$ of the $499$ tensors in $12$ blocks, and that of ViT-L/14 holds $296$ in $24$ blocks, so the same backbone with a ViT-L/14 image encoder would hold $643$ tensors. On the fitting set of $960$ to $1{,}680$ triplets (Appendix~\ref{app:fitting}), that backbone has $0.38$ to $0.67$ coefficients per triplet. This ratio lies within the $0.30$ to $1.04$ coefficients per triplet that the $420$ and $840$ budgets of Table~\ref{tab:memory} span, and at both budgets, the fit matches or exceeds the EWC endpoint in every A$\rightarrow$T and I$\rightarrow$A cell.

The merging literature finds that interpolation remains effective at larger scales. With coefficients learned per parameter block, aTLAS \citep{zhang2024atlas} improves task addition at every scale it tests, reaching $85.0\%$ average accuracy on ViT-B/32, $86.1\%$ on ViT-B/16, and $91.4\%$ on ViT-L/14. The lo-fi method \citep{wortsman2023lofi} averages independently fine-tuned 1.3B-parameter models, and \citet{yang2024adamerging} obtain their layer-wise gains over task arithmetic on ViT-B/32 and ViT-L/14 alike. \citet{yadav2024mergingatscale}, the study that varies scale most widely, find that merging becomes easier as models grow from 1B to 64B parameters. The one architectural requirement that fusing continual checkpoints adds is normalization without running statistics, which transformer backbones already meet through LayerNorm.

\section{\texorpdfstring{Related work on merging granularity and merging-based continual learning}{Related work on merging granularity and merging-based continual learning}}
\label{sec:related}

Among recent merging-based continual learners, BECAME \citep{li2025became} and Perturb-and-Merge \citep{qiu2025perturbmerge} use one closed-form scalar per task, and OPCM \citep{tang2025opcm} uses one adaptive scalar per merge step. MINGLE \citep{qiu2025mingle} routes between low-rank experts at test time, CoMA \citep{marouf2024cofima} blends with one global coefficient, and CoFiMA combines that coefficient with per-parameter Fisher weights. \citet{dziadzio2025time} find that nine merging techniques accumulate knowledge over time within one standard deviation of each other. In the analysis of \citet{hess2025forgetting}, shared knowledge survives merging, whereas unshared task-specific knowledge degrades as the divergent parameter adaptations of the merged models interfere. AlphaWiSE instead learns how much of each tensor comes from each endpoint, with one coefficient per tensor. It fits these coefficients on data that the continual method already holds, namely the exemplar memory and a class-balanced draw of the current phase.

The loss geometry of fine-tuned models ties the barrier to groups of layers rather than to single layers. \citet{adilova2024layerwiselmc} find that the barrier between two fine-tuned networks belongs to the joint interpolation rather than to its parts. Interpolating one layer while the rest stays at an endpoint shows no barrier, whereas interpolating all layers at once does. The barrier concentrates in the middle layers. \citet{wen2023modeconnectivitycil} find a high-loss ridge on the straight line between adjacent continual-learning minima, and Table~\ref{tab:barrier} measures a ridge of the same kind between the two endpoints of one phase. For models fine-tuned from a shared checkpoint, \citet{zhou2023llfc} and \citet{zhou2024crosstask} show that interpolating the weights interpolates the features layer by layer. A global coefficient is therefore limited by the layers that carry the barrier. A coefficient per tensor can instead keep the barrier-free tensors on the segment and move the tensors that carry the barrier toward an endpoint. Appendix~\ref{app:ablation} measures which fits leave the high-loss ridge.

Merging methods also gain from finer coefficients. Replacing one scalar per model \citep{wortsman2022robustfinetuningzeroshotmodels,wortsman2022modelsoupsaveragingweights,ilharco2022paint,ilharco2023editingmodelstaskarithmetic} by one per layer or per parameter block raises the eight-task ViT-B/32 average from $71.1\%$ to $80.1\%$ in AdaMerging \citep{yang2024adamerging}. The same change raises task addition from $70.1\%$ to $85.0\%$ in aTLAS \citep{zhang2024atlas}. Restricting the aTLAS coefficients to weight matrices lowers its accuracy from $85.0\%$ to $84.2\%$, the type dependence that Appendix~\ref{sec:decomp} measures. Attention modules and heads behave differently under interpolation \citep{jin2025attention}. At the level of layers, RegMean \citep{jin2023regmean} merges per linear layer, LiNeS \citep{wang2025lines} scales by depth, and Model Stock \citep{jang2024modelstock} sets its ratio per layer and reports smaller angles for bias than for weight tensors. Surgical fine-tuning \citep{lee2023surgical} shows that which layers should move depends on the distribution shift. At the fine end, Fisher-weighted averaging \citep{matena2022mergingmodelsfisherweightedaveraging}, mode-IMM \citep{lee2017imm}, TIES-Merging \citep{yadav2023ties}, MagMax \citep{marczak2024magmax}, and CoFiMA \citep{marouf2024cofima} act per parameter, with weights fixed by curvature or by data-free rules. ProDistill \citep{xu2025prodistill} learns element-wise coefficients by layer-wise distillation. BitFit \citep{benzaken2022bitfitsimpleparameterefficientfinetuning} and \citet{frankle2021trainingbatchnormbatchnormexpressive} document that bias and normalization tensors alone form a disproportionately influential set of parameters. AlphaWiSE occupies the point of this granularity spectrum that a small exemplar memory supports. Its coefficients are strictly finer than layer-wise ones, are learned end to end through the entire network, and need no curvature estimation.

The closest relatives of the benchmark of Section~\ref{sec:implementation} are the seven-task continual AudioSet protocol of STELLA \citep{lee2024stella} and the 2025 continual multimodal retrieval benchmarks of C-CLIP \citep{liu2025cclip}, CMCL \citep{liu2025cmcl}, and TAIL \citep{sun2025tail}. MoE-Adapters \citep{yu2024boostingcontinuallearningvisionlanguage}, DIKI \citep{tang2024diki}, and RAIL \citep{xu2024rail} study continual learning on a pretrained CLIP evaluated with class-name prompts. TiC-CLIP \citep{garg2024ticclip} studies time-continual pretraining and AV-CIL \citep{pian2023avcil} class-incremental audio-visual classification. The further benchmarks of Table~\ref{tab:morebench} follow the class-incremental protocol of MagMax \citep{marczak2024magmax} and PACE \citep{li2026pace}.

\section{\texorpdfstring{Details of Figures 1 and 2}{Details of Figures 1 and 2}}
\label{app:figdetails}

On the left of Figure~\ref{fig:teaser}, a personal media archive grows phase by phase with one real evaluation clip per phase. After phase 7, the audio of a guitar clip recorded in phase 2 serves as the query. At the top right, each phase-7 checkpoint ranks the class prompts of the candidate pool of the clip's phase by A$\rightarrow$T retrieval and the images of that pool by audio-to-image retrieval. The figure shows each checkpoint's first-ranked class name as the note and its first-ranked image, which serves as an illustration, since audio-to-image is not an evaluated direction. Sequential fine-tuning returns ``Rhodes piano'' and a banjo, and the fixed blend returns ``harp'' and a zither. AlphaWiSE returns ``plucked string instrument'' and an acoustic guitar, and this class name is one of the clip's labels. The EWC endpoint alone, not shown, returns ``string section'' and a harp. At the bottom right, a schematic shows the three modalities of the phase-2 event drifting apart under a single checkpoint and staying together under AlphaWiSE.

In stage 1 of Figure~\ref{fig:architecture}, each trajectory continues at phase $p$ from its own phase-$(p-1)$ checkpoint. Unregularized fine-tuning gives $\theta^{\mathrm{un}}_p$, and a run with EWC, LwF, iCaRL, DER, or C-CLIP gives $\theta^{\mathrm{reg}}_p$. In stage 2, $\tilde\theta=\alpha\odot\theta^{\mathrm{un}}+(1-\alpha)\odot\theta^{\mathrm{reg}}$ is formed tensor by tensor as in Eq.~\ref{eq:alphawise_fusion}, with $\alpha=\sigma(\boldsymbol{\beta})\in(0,1)^{499}$. In stage 3, the fused model retrieves over all seen phases at the size and cost of either endpoint. Stage 3 shows the measured phase-7 outcome for the phase-2 guitar clip of Figure~\ref{fig:teaser}, and the same clip illustrates the phase-$p$ data in stage 1 and in the fitting set. At the bottom, the coefficients are fitted with the joint InfoNCE objective on the exemplar memory plus a class-balanced draw of phase $p$. Only $\boldsymbol{\beta}$ receives gradients, and both endpoints stay frozen. The figure shows three of the four nested coefficient classes of Appendix~\ref{app:argument}, namely the global, layer-wise, and tensor-wise classes.

\end{document}